\documentclass[11pt]{article}
\usepackage[final]{acl}
\usepackage{times}
\usepackage{latexsym}
\usepackage[T1]{fontenc}
\usepackage[utf8]{inputenc}
\usepackage{inconsolata}
\usepackage{microtype,graphicx,xcolor,eso-pic,amsmath,caption,fvextra,placeins}
\IfFileExists{build-mode.tex}{
}{\errmessage{Use scripts/build_paper.py; verified build mode is required}}
\newcommand{\JLNumber}[1]{\ifcsname jl@#1\endcsname\csname jl@#1\endcsname\else\errmessage{Unknown report number}\fi}
\expandafter\def\csname jl@/agreement/0/alpha\endcsname{0.653}
\expandafter\def\csname jl@/agreement/0/complete_items\endcsname{320.000}
\expandafter\def\csname jl@/agreement/0/fleiss_kappa\endcsname{0.242}
\expandafter\def\csname jl@/agreement/0/icc_2k\endcsname{0.915}
\expandafter\def\csname jl@/agreement/0/n_items\endcsname{320.000}
\expandafter\def\csname jl@/agreement/1/alpha\endcsname{0.741}
\expandafter\def\csname jl@/agreement/1/complete_items\endcsname{320.000}
\expandafter\def\csname jl@/agreement/1/fleiss_kappa\endcsname{0.499}
\expandafter\def\csname jl@/agreement/1/icc_2k\endcsname{0.966}
\expandafter\def\csname jl@/agreement/1/n_items\endcsname{320.000}
\expandafter\def\csname jl@/agreement/10/alpha\endcsname{0.394}
\expandafter\def\csname jl@/agreement/10/complete_items\endcsname{320.000}
\expandafter\def\csname jl@/agreement/10/fleiss_kappa\endcsname{0.126}
\expandafter\def\csname jl@/agreement/10/icc_2k\endcsname{0.766}
\expandafter\def\csname jl@/agreement/10/n_items\endcsname{320.000}
\expandafter\def\csname jl@/agreement/11/alpha\endcsname{0.540}
\expandafter\def\csname jl@/agreement/11/complete_items\endcsname{320.000}
\expandafter\def\csname jl@/agreement/11/fleiss_kappa\endcsname{0.173}
\expandafter\def\csname jl@/agreement/11/icc_2k\endcsname{0.867}
\expandafter\def\csname jl@/agreement/11/n_items\endcsname{320.000}
\expandafter\def\csname jl@/agreement/12/alpha\endcsname{0.614}
\expandafter\def\csname jl@/agreement/12/complete_items\endcsname{320.000}
\expandafter\def\csname jl@/agreement/12/fleiss_kappa\endcsname{0.236}
\expandafter\def\csname jl@/agreement/12/icc_2k\endcsname{0.897}
\expandafter\def\csname jl@/agreement/12/n_items\endcsname{320.000}
\expandafter\def\csname jl@/agreement/13/alpha\endcsname{0.750}
\expandafter\def\csname jl@/agreement/13/complete_items\endcsname{320.000}
\expandafter\def\csname jl@/agreement/13/fleiss_kappa\endcsname{0.495}
\expandafter\def\csname jl@/agreement/13/icc_2k\endcsname{0.968}
\expandafter\def\csname jl@/agreement/13/n_items\endcsname{320.000}
\expandafter\def\csname jl@/agreement/14/alpha\endcsname{0.278}
\expandafter\def\csname jl@/agreement/14/complete_items\endcsname{320.000}
\expandafter\def\csname jl@/agreement/14/fleiss_kappa\endcsname{0.100}
\expandafter\def\csname jl@/agreement/14/icc_2k\endcsname{0.770}
\expandafter\def\csname jl@/agreement/14/n_items\endcsname{320.000}
\expandafter\def\csname jl@/agreement/15/alpha\endcsname{0.581}
\expandafter\def\csname jl@/agreement/15/complete_items\endcsname{320.000}
\expandafter\def\csname jl@/agreement/15/fleiss_kappa\endcsname{0.208}
\expandafter\def\csname jl@/agreement/15/icc_2k\endcsname{0.884}
\expandafter\def\csname jl@/agreement/15/n_items\endcsname{320.000}
\expandafter\def\csname jl@/agreement/16/alpha\endcsname{0.705}
\expandafter\def\csname jl@/agreement/16/complete_items\endcsname{320.000}
\expandafter\def\csname jl@/agreement/16/fleiss_kappa\endcsname{0.338}
\expandafter\def\csname jl@/agreement/16/icc_2k\endcsname{0.935}
\expandafter\def\csname jl@/agreement/16/n_items\endcsname{320.000}
\expandafter\def\csname jl@/agreement/17/alpha\endcsname{0.770}
\expandafter\def\csname jl@/agreement/17/complete_items\endcsname{320.000}
\expandafter\def\csname jl@/agreement/17/fleiss_kappa\endcsname{0.506}
\expandafter\def\csname jl@/agreement/17/icc_2k\endcsname{0.970}
\expandafter\def\csname jl@/agreement/17/n_items\endcsname{320.000}
\expandafter\def\csname jl@/agreement/18/alpha\endcsname{0.440}
\expandafter\def\csname jl@/agreement/18/complete_items\endcsname{320.000}
\expandafter\def\csname jl@/agreement/18/fleiss_kappa\endcsname{0.158}
\expandafter\def\csname jl@/agreement/18/icc_2k\endcsname{0.834}
\expandafter\def\csname jl@/agreement/18/n_items\endcsname{320.000}
\expandafter\def\csname jl@/agreement/19/alpha\endcsname{0.666}
\expandafter\def\csname jl@/agreement/19/complete_items\endcsname{320.000}
\expandafter\def\csname jl@/agreement/19/fleiss_kappa\endcsname{0.279}
\expandafter\def\csname jl@/agreement/19/icc_2k\endcsname{0.916}
\expandafter\def\csname jl@/agreement/19/n_items\endcsname{320.000}
\expandafter\def\csname jl@/agreement/2/alpha\endcsname{0.366}
\expandafter\def\csname jl@/agreement/2/complete_items\endcsname{320.000}
\expandafter\def\csname jl@/agreement/2/fleiss_kappa\endcsname{0.097}
\expandafter\def\csname jl@/agreement/2/icc_2k\endcsname{0.752}
\expandafter\def\csname jl@/agreement/2/n_items\endcsname{320.000}
\expandafter\def\csname jl@/agreement/20/alpha\endcsname{0.682}
\expandafter\def\csname jl@/agreement/20/complete_items\endcsname{320.000}
\expandafter\def\csname jl@/agreement/20/fleiss_kappa\endcsname{0.339}
\expandafter\def\csname jl@/agreement/20/icc_2k\endcsname{0.932}
\expandafter\def\csname jl@/agreement/20/n_items\endcsname{320.000}
\expandafter\def\csname jl@/agreement/21/alpha\endcsname{0.759}
\expandafter\def\csname jl@/agreement/21/complete_items\endcsname{320.000}
\expandafter\def\csname jl@/agreement/21/fleiss_kappa\endcsname{0.499}
\expandafter\def\csname jl@/agreement/21/icc_2k\endcsname{0.963}
\expandafter\def\csname jl@/agreement/21/n_items\endcsname{320.000}
\expandafter\def\csname jl@/agreement/22/alpha\endcsname{0.240}
\expandafter\def\csname jl@/agreement/22/complete_items\endcsname{320.000}
\expandafter\def\csname jl@/agreement/22/fleiss_kappa\endcsname{0.122}
\expandafter\def\csname jl@/agreement/22/icc_2k\endcsname{0.639}
\expandafter\def\csname jl@/agreement/22/n_items\endcsname{320.000}
\expandafter\def\csname jl@/agreement/23/alpha\endcsname{0.607}
\expandafter\def\csname jl@/agreement/23/complete_items\endcsname{320.000}
\expandafter\def\csname jl@/agreement/23/fleiss_kappa\endcsname{0.249}
\expandafter\def\csname jl@/agreement/23/icc_2k\endcsname{0.889}
\expandafter\def\csname jl@/agreement/23/n_items\endcsname{320.000}
\expandafter\def\csname jl@/agreement/24/alpha\endcsname{0.375}
\expandafter\def\csname jl@/agreement/24/complete_items\endcsname{797.000}
\expandafter\def\csname jl@/agreement/24/fleiss_kappa\endcsname{0.376}
\expandafter\def\csname jl@/agreement/24/icc_2k\endcsname{0.763}
\expandafter\def\csname jl@/agreement/24/n_items\endcsname{799.000}
\expandafter\def\csname jl@/agreement/25/alpha\endcsname{0.369}
\expandafter\def\csname jl@/agreement/25/complete_items\endcsname{798.000}
\expandafter\def\csname jl@/agreement/25/fleiss_kappa\endcsname{0.370}
\expandafter\def\csname jl@/agreement/25/icc_2k\endcsname{0.757}
\expandafter\def\csname jl@/agreement/25/n_items\endcsname{799.000}
\expandafter\def\csname jl@/agreement/26/alpha\endcsname{0.389}
\expandafter\def\csname jl@/agreement/26/complete_items\endcsname{792.000}
\expandafter\def\csname jl@/agreement/26/fleiss_kappa\endcsname{0.393}
\expandafter\def\csname jl@/agreement/26/icc_2k\endcsname{0.775}
\expandafter\def\csname jl@/agreement/26/n_items\endcsname{799.000}
\expandafter\def\csname jl@/agreement/27/alpha\endcsname{0.358}
\expandafter\def\csname jl@/agreement/27/complete_items\endcsname{795.000}
\expandafter\def\csname jl@/agreement/27/fleiss_kappa\endcsname{0.360}
\expandafter\def\csname jl@/agreement/27/icc_2k\endcsname{0.754}
\expandafter\def\csname jl@/agreement/27/n_items\endcsname{799.000}
\expandafter\def\csname jl@/agreement/28/alpha\endcsname{0.362}
\expandafter\def\csname jl@/agreement/28/complete_items\endcsname{792.000}
\expandafter\def\csname jl@/agreement/28/fleiss_kappa\endcsname{0.366}
\expandafter\def\csname jl@/agreement/28/icc_2k\endcsname{0.757}
\expandafter\def\csname jl@/agreement/28/n_items\endcsname{799.000}
\expandafter\def\csname jl@/agreement/29/alpha\endcsname{0.359}
\expandafter\def\csname jl@/agreement/29/complete_items\endcsname{796.000}
\expandafter\def\csname jl@/agreement/29/fleiss_kappa\endcsname{0.360}
\expandafter\def\csname jl@/agreement/29/icc_2k\endcsname{0.753}
\expandafter\def\csname jl@/agreement/29/n_items\endcsname{799.000}
\expandafter\def\csname jl@/agreement/3/alpha\endcsname{0.547}
\expandafter\def\csname jl@/agreement/3/complete_items\endcsname{320.000}
\expandafter\def\csname jl@/agreement/3/fleiss_kappa\endcsname{0.184}
\expandafter\def\csname jl@/agreement/3/icc_2k\endcsname{0.874}
\expandafter\def\csname jl@/agreement/3/n_items\endcsname{320.000}
\expandafter\def\csname jl@/agreement/30/alpha\endcsname{0.730}
\expandafter\def\csname jl@/agreement/30/complete_items\endcsname{288.000}
\expandafter\def\csname jl@/agreement/30/fleiss_kappa\endcsname{0.526}
\expandafter\def\csname jl@/agreement/30/icc_2k\endcsname{0.934}
\expandafter\def\csname jl@/agreement/30/n_items\endcsname{288.000}
\expandafter\def\csname jl@/agreement/31/alpha\endcsname{0.535}
\expandafter\def\csname jl@/agreement/31/complete_items\endcsname{288.000}
\expandafter\def\csname jl@/agreement/31/fleiss_kappa\endcsname{0.365}
\expandafter\def\csname jl@/agreement/31/icc_2k\endcsname{0.856}
\expandafter\def\csname jl@/agreement/31/n_items\endcsname{288.000}
\expandafter\def\csname jl@/agreement/32/alpha\endcsname{0.707}
\expandafter\def\csname jl@/agreement/32/complete_items\endcsname{288.000}
\expandafter\def\csname jl@/agreement/32/fleiss_kappa\endcsname{0.471}
\expandafter\def\csname jl@/agreement/32/icc_2k\endcsname{0.919}
\expandafter\def\csname jl@/agreement/32/n_items\endcsname{288.000}
\expandafter\def\csname jl@/agreement/33/alpha\endcsname{0.701}
\expandafter\def\csname jl@/agreement/33/complete_items\endcsname{288.000}
\expandafter\def\csname jl@/agreement/33/fleiss_kappa\endcsname{0.307}
\expandafter\def\csname jl@/agreement/33/icc_2k\endcsname{0.920}
\expandafter\def\csname jl@/agreement/33/n_items\endcsname{288.000}
\expandafter\def\csname jl@/agreement/34/alpha\endcsname{0.483}
\expandafter\def\csname jl@/agreement/34/complete_items\endcsname{288.000}
\expandafter\def\csname jl@/agreement/34/fleiss_kappa\endcsname{0.483}
\expandafter\def\csname jl@/agreement/34/icc_2k\endcsname{0.827}
\expandafter\def\csname jl@/agreement/34/n_items\endcsname{288.000}
\expandafter\def\csname jl@/agreement/35/alpha\endcsname{0.796}
\expandafter\def\csname jl@/agreement/35/complete_items\endcsname{288.000}
\expandafter\def\csname jl@/agreement/35/fleiss_kappa\endcsname{0.796}
\expandafter\def\csname jl@/agreement/35/icc_2k\endcsname{0.951}
\expandafter\def\csname jl@/agreement/35/n_items\endcsname{288.000}
\expandafter\def\csname jl@/agreement/36/alpha\endcsname{0.728}
\expandafter\def\csname jl@/agreement/36/complete_items\endcsname{288.000}
\expandafter\def\csname jl@/agreement/36/fleiss_kappa\endcsname{0.527}
\expandafter\def\csname jl@/agreement/36/icc_2k\endcsname{0.932}
\expandafter\def\csname jl@/agreement/36/n_items\endcsname{288.000}
\expandafter\def\csname jl@/agreement/37/alpha\endcsname{0.528}
\expandafter\def\csname jl@/agreement/37/complete_items\endcsname{288.000}
\expandafter\def\csname jl@/agreement/37/fleiss_kappa\endcsname{0.387}
\expandafter\def\csname jl@/agreement/37/icc_2k\endcsname{0.866}
\expandafter\def\csname jl@/agreement/37/n_items\endcsname{288.000}
\expandafter\def\csname jl@/agreement/38/alpha\endcsname{0.742}
\expandafter\def\csname jl@/agreement/38/complete_items\endcsname{288.000}
\expandafter\def\csname jl@/agreement/38/fleiss_kappa\endcsname{0.550}
\expandafter\def\csname jl@/agreement/38/icc_2k\endcsname{0.940}
\expandafter\def\csname jl@/agreement/38/n_items\endcsname{288.000}
\expandafter\def\csname jl@/agreement/39/alpha\endcsname{0.718}
\expandafter\def\csname jl@/agreement/39/complete_items\endcsname{288.000}
\expandafter\def\csname jl@/agreement/39/fleiss_kappa\endcsname{0.337}
\expandafter\def\csname jl@/agreement/39/icc_2k\endcsname{0.921}
\expandafter\def\csname jl@/agreement/39/n_items\endcsname{288.000}
\expandafter\def\csname jl@/agreement/4/alpha\endcsname{0.634}
\expandafter\def\csname jl@/agreement/4/complete_items\endcsname{320.000}
\expandafter\def\csname jl@/agreement/4/fleiss_kappa\endcsname{0.251}
\expandafter\def\csname jl@/agreement/4/icc_2k\endcsname{0.912}
\expandafter\def\csname jl@/agreement/4/n_items\endcsname{320.000}
\expandafter\def\csname jl@/agreement/40/alpha\endcsname{0.464}
\expandafter\def\csname jl@/agreement/40/complete_items\endcsname{288.000}
\expandafter\def\csname jl@/agreement/40/fleiss_kappa\endcsname{0.464}
\expandafter\def\csname jl@/agreement/40/icc_2k\endcsname{0.816}
\expandafter\def\csname jl@/agreement/40/n_items\endcsname{288.000}
\expandafter\def\csname jl@/agreement/41/alpha\endcsname{0.842}
\expandafter\def\csname jl@/agreement/41/complete_items\endcsname{288.000}
\expandafter\def\csname jl@/agreement/41/fleiss_kappa\endcsname{0.842}
\expandafter\def\csname jl@/agreement/41/icc_2k\endcsname{0.964}
\expandafter\def\csname jl@/agreement/41/n_items\endcsname{288.000}
\expandafter\def\csname jl@/agreement/42/alpha\endcsname{0.758}
\expandafter\def\csname jl@/agreement/42/complete_items\endcsname{288.000}
\expandafter\def\csname jl@/agreement/42/fleiss_kappa\endcsname{0.570}
\expandafter\def\csname jl@/agreement/42/icc_2k\endcsname{0.942}
\expandafter\def\csname jl@/agreement/42/n_items\endcsname{288.000}
\expandafter\def\csname jl@/agreement/43/alpha\endcsname{0.510}
\expandafter\def\csname jl@/agreement/43/complete_items\endcsname{288.000}
\expandafter\def\csname jl@/agreement/43/fleiss_kappa\endcsname{0.364}
\expandafter\def\csname jl@/agreement/43/icc_2k\endcsname{0.847}
\expandafter\def\csname jl@/agreement/43/n_items\endcsname{288.000}
\expandafter\def\csname jl@/agreement/44/alpha\endcsname{0.717}
\expandafter\def\csname jl@/agreement/44/complete_items\endcsname{288.000}
\expandafter\def\csname jl@/agreement/44/fleiss_kappa\endcsname{0.488}
\expandafter\def\csname jl@/agreement/44/icc_2k\endcsname{0.928}
\expandafter\def\csname jl@/agreement/44/n_items\endcsname{288.000}
\expandafter\def\csname jl@/agreement/45/alpha\endcsname{0.650}
\expandafter\def\csname jl@/agreement/45/complete_items\endcsname{288.000}
\expandafter\def\csname jl@/agreement/45/fleiss_kappa\endcsname{0.288}
\expandafter\def\csname jl@/agreement/45/icc_2k\endcsname{0.899}
\expandafter\def\csname jl@/agreement/45/n_items\endcsname{288.000}
\expandafter\def\csname jl@/agreement/46/alpha\endcsname{0.450}
\expandafter\def\csname jl@/agreement/46/complete_items\endcsname{288.000}
\expandafter\def\csname jl@/agreement/46/fleiss_kappa\endcsname{0.450}
\expandafter\def\csname jl@/agreement/46/icc_2k\endcsname{0.808}
\expandafter\def\csname jl@/agreement/46/n_items\endcsname{288.000}
\expandafter\def\csname jl@/agreement/47/alpha\endcsname{0.795}
\expandafter\def\csname jl@/agreement/47/complete_items\endcsname{288.000}
\expandafter\def\csname jl@/agreement/47/fleiss_kappa\endcsname{0.794}
\expandafter\def\csname jl@/agreement/47/icc_2k\endcsname{0.951}
\expandafter\def\csname jl@/agreement/47/n_items\endcsname{288.000}
\expandafter\def\csname jl@/agreement/48/alpha\endcsname{0.718}
\expandafter\def\csname jl@/agreement/48/complete_items\endcsname{288.000}
\expandafter\def\csname jl@/agreement/48/fleiss_kappa\endcsname{0.512}
\expandafter\def\csname jl@/agreement/48/icc_2k\endcsname{0.930}
\expandafter\def\csname jl@/agreement/48/n_items\endcsname{288.000}
\expandafter\def\csname jl@/agreement/49/alpha\endcsname{0.564}
\expandafter\def\csname jl@/agreement/49/complete_items\endcsname{288.000}
\expandafter\def\csname jl@/agreement/49/fleiss_kappa\endcsname{0.406}
\expandafter\def\csname jl@/agreement/49/icc_2k\endcsname{0.878}
\expandafter\def\csname jl@/agreement/49/n_items\endcsname{288.000}
\expandafter\def\csname jl@/agreement/5/alpha\endcsname{0.721}
\expandafter\def\csname jl@/agreement/5/complete_items\endcsname{320.000}
\expandafter\def\csname jl@/agreement/5/fleiss_kappa\endcsname{0.472}
\expandafter\def\csname jl@/agreement/5/icc_2k\endcsname{0.958}
\expandafter\def\csname jl@/agreement/5/n_items\endcsname{320.000}
\expandafter\def\csname jl@/agreement/50/alpha\endcsname{0.705}
\expandafter\def\csname jl@/agreement/50/complete_items\endcsname{288.000}
\expandafter\def\csname jl@/agreement/50/fleiss_kappa\endcsname{0.487}
\expandafter\def\csname jl@/agreement/50/icc_2k\endcsname{0.926}
\expandafter\def\csname jl@/agreement/50/n_items\endcsname{288.000}
\expandafter\def\csname jl@/agreement/51/alpha\endcsname{0.684}
\expandafter\def\csname jl@/agreement/51/complete_items\endcsname{288.000}
\expandafter\def\csname jl@/agreement/51/fleiss_kappa\endcsname{0.291}
\expandafter\def\csname jl@/agreement/51/icc_2k\endcsname{0.912}
\expandafter\def\csname jl@/agreement/51/n_items\endcsname{288.000}
\expandafter\def\csname jl@/agreement/52/alpha\endcsname{0.518}
\expandafter\def\csname jl@/agreement/52/complete_items\endcsname{288.000}
\expandafter\def\csname jl@/agreement/52/fleiss_kappa\endcsname{0.518}
\expandafter\def\csname jl@/agreement/52/icc_2k\endcsname{0.846}
\expandafter\def\csname jl@/agreement/52/n_items\endcsname{288.000}
\expandafter\def\csname jl@/agreement/53/alpha\endcsname{0.787}
\expandafter\def\csname jl@/agreement/53/complete_items\endcsname{288.000}
\expandafter\def\csname jl@/agreement/53/fleiss_kappa\endcsname{0.787}
\expandafter\def\csname jl@/agreement/53/icc_2k\endcsname{0.949}
\expandafter\def\csname jl@/agreement/53/n_items\endcsname{288.000}
\expandafter\def\csname jl@/agreement/54/alpha\endcsname{0.730}
\expandafter\def\csname jl@/agreement/54/complete_items\endcsname{288.000}
\expandafter\def\csname jl@/agreement/54/fleiss_kappa\endcsname{0.539}
\expandafter\def\csname jl@/agreement/54/icc_2k\endcsname{0.934}
\expandafter\def\csname jl@/agreement/54/n_items\endcsname{288.000}
\expandafter\def\csname jl@/agreement/55/alpha\endcsname{0.449}
\expandafter\def\csname jl@/agreement/55/complete_items\endcsname{288.000}
\expandafter\def\csname jl@/agreement/55/fleiss_kappa\endcsname{0.311}
\expandafter\def\csname jl@/agreement/55/icc_2k\endcsname{0.822}
\expandafter\def\csname jl@/agreement/55/n_items\endcsname{288.000}
\expandafter\def\csname jl@/agreement/56/alpha\endcsname{0.687}
\expandafter\def\csname jl@/agreement/56/complete_items\endcsname{288.000}
\expandafter\def\csname jl@/agreement/56/fleiss_kappa\endcsname{0.456}
\expandafter\def\csname jl@/agreement/56/icc_2k\endcsname{0.916}
\expandafter\def\csname jl@/agreement/56/n_items\endcsname{288.000}
\expandafter\def\csname jl@/agreement/57/alpha\endcsname{0.722}
\expandafter\def\csname jl@/agreement/57/complete_items\endcsname{288.000}
\expandafter\def\csname jl@/agreement/57/fleiss_kappa\endcsname{0.406}
\expandafter\def\csname jl@/agreement/57/icc_2k\endcsname{0.920}
\expandafter\def\csname jl@/agreement/57/n_items\endcsname{288.000}
\expandafter\def\csname jl@/agreement/58/alpha\endcsname{0.613}
\expandafter\def\csname jl@/agreement/58/complete_items\endcsname{288.000}
\expandafter\def\csname jl@/agreement/58/fleiss_kappa\endcsname{0.613}
\expandafter\def\csname jl@/agreement/58/icc_2k\endcsname{0.889}
\expandafter\def\csname jl@/agreement/58/n_items\endcsname{288.000}
\expandafter\def\csname jl@/agreement/59/alpha\endcsname{0.910}
\expandafter\def\csname jl@/agreement/59/complete_items\endcsname{288.000}
\expandafter\def\csname jl@/agreement/59/fleiss_kappa\endcsname{0.910}
\expandafter\def\csname jl@/agreement/59/icc_2k\endcsname{0.981}
\expandafter\def\csname jl@/agreement/59/n_items\endcsname{288.000}
\expandafter\def\csname jl@/agreement/6/alpha\endcsname{0.283}
\expandafter\def\csname jl@/agreement/6/complete_items\endcsname{320.000}
\expandafter\def\csname jl@/agreement/6/fleiss_kappa\endcsname{0.173}
\expandafter\def\csname jl@/agreement/6/icc_2k\endcsname{0.730}
\expandafter\def\csname jl@/agreement/6/n_items\endcsname{320.000}
\expandafter\def\csname jl@/agreement/60/alpha\endcsname{0.689}
\expandafter\def\csname jl@/agreement/60/complete_items\endcsname{288.000}
\expandafter\def\csname jl@/agreement/60/fleiss_kappa\endcsname{0.492}
\expandafter\def\csname jl@/agreement/60/icc_2k\endcsname{0.919}
\expandafter\def\csname jl@/agreement/60/n_items\endcsname{288.000}
\expandafter\def\csname jl@/agreement/61/alpha\endcsname{0.547}
\expandafter\def\csname jl@/agreement/61/complete_items\endcsname{288.000}
\expandafter\def\csname jl@/agreement/61/fleiss_kappa\endcsname{0.383}
\expandafter\def\csname jl@/agreement/61/icc_2k\endcsname{0.859}
\expandafter\def\csname jl@/agreement/61/n_items\endcsname{288.000}
\expandafter\def\csname jl@/agreement/62/alpha\endcsname{0.722}
\expandafter\def\csname jl@/agreement/62/complete_items\endcsname{288.000}
\expandafter\def\csname jl@/agreement/62/fleiss_kappa\endcsname{0.505}
\expandafter\def\csname jl@/agreement/62/icc_2k\endcsname{0.930}
\expandafter\def\csname jl@/agreement/62/n_items\endcsname{288.000}
\expandafter\def\csname jl@/agreement/63/alpha\endcsname{0.707}
\expandafter\def\csname jl@/agreement/63/complete_items\endcsname{288.000}
\expandafter\def\csname jl@/agreement/63/fleiss_kappa\endcsname{0.370}
\expandafter\def\csname jl@/agreement/63/icc_2k\endcsname{0.917}
\expandafter\def\csname jl@/agreement/63/n_items\endcsname{288.000}
\expandafter\def\csname jl@/agreement/64/alpha\endcsname{0.565}
\expandafter\def\csname jl@/agreement/64/complete_items\endcsname{288.000}
\expandafter\def\csname jl@/agreement/64/fleiss_kappa\endcsname{0.565}
\expandafter\def\csname jl@/agreement/64/icc_2k\endcsname{0.869}
\expandafter\def\csname jl@/agreement/64/n_items\endcsname{288.000}
\expandafter\def\csname jl@/agreement/65/alpha\endcsname{0.837}
\expandafter\def\csname jl@/agreement/65/complete_items\endcsname{288.000}
\expandafter\def\csname jl@/agreement/65/fleiss_kappa\endcsname{0.837}
\expandafter\def\csname jl@/agreement/65/icc_2k\endcsname{0.963}
\expandafter\def\csname jl@/agreement/65/n_items\endcsname{288.000}
\expandafter\def\csname jl@/agreement/7/alpha\endcsname{0.527}
\expandafter\def\csname jl@/agreement/7/complete_items\endcsname{320.000}
\expandafter\def\csname jl@/agreement/7/fleiss_kappa\endcsname{0.189}
\expandafter\def\csname jl@/agreement/7/icc_2k\endcsname{0.869}
\expandafter\def\csname jl@/agreement/7/n_items\endcsname{320.000}
\expandafter\def\csname jl@/agreement/8/alpha\endcsname{0.609}
\expandafter\def\csname jl@/agreement/8/complete_items\endcsname{320.000}
\expandafter\def\csname jl@/agreement/8/fleiss_kappa\endcsname{0.222}
\expandafter\def\csname jl@/agreement/8/icc_2k\endcsname{0.900}
\expandafter\def\csname jl@/agreement/8/n_items\endcsname{320.000}
\expandafter\def\csname jl@/agreement/9/alpha\endcsname{0.737}
\expandafter\def\csname jl@/agreement/9/complete_items\endcsname{320.000}
\expandafter\def\csname jl@/agreement/9/fleiss_kappa\endcsname{0.500}
\expandafter\def\csname jl@/agreement/9/icc_2k\endcsname{0.962}
\expandafter\def\csname jl@/agreement/9/n_items\endcsname{320.000}
\expandafter\def\csname jl@/analysis_plan/alpha_threshold\endcsname{0.050}
\expandafter\def\csname jl@/analysis_plan/holm_family\endcsname{9.000}
\expandafter\def\csname jl@/analysis_plan/iterations\endcsname{10000.000}
\expandafter\def\csname jl@/analysis_plan/seed\endcsname{20261005.000}
\expandafter\def\csname jl@/comparisons/0/ci_high\endcsname{0.092}
\expandafter\def\csname jl@/comparisons/0/ci_low\endcsname{0.046}
\expandafter\def\csname jl@/comparisons/0/clusters\endcsname{20.000}
\expandafter\def\csname jl@/comparisons/0/delta\endcsname{0.069}
\expandafter\def\csname jl@/comparisons/0/dropped_items\endcsname{0.000}
\expandafter\def\csname jl@/comparisons/0/iterations\endcsname{10000.000}
\expandafter\def\csname jl@/comparisons/0/p_holm\endcsname{0.001}
\expandafter\def\csname jl@/comparisons/0/p_value\endcsname{0.000}
\expandafter\def\csname jl@/comparisons/0/paired_items\endcsname{1280.000}
\expandafter\def\csname jl@/comparisons/0/valid_replicates\endcsname{10000.000}
\expandafter\def\csname jl@/comparisons/1/ci_high\endcsname{0.021}
\expandafter\def\csname jl@/comparisons/1/ci_low\endcsname{-0.037}
\expandafter\def\csname jl@/comparisons/1/clusters\endcsname{20.000}
\expandafter\def\csname jl@/comparisons/1/delta\endcsname{-0.007}
\expandafter\def\csname jl@/comparisons/1/dropped_items\endcsname{0.000}
\expandafter\def\csname jl@/comparisons/1/iterations\endcsname{10000.000}
\expandafter\def\csname jl@/comparisons/1/p_holm\endcsname{1.000}
\expandafter\def\csname jl@/comparisons/1/p_value\endcsname{0.636}
\expandafter\def\csname jl@/comparisons/1/paired_items\endcsname{1280.000}
\expandafter\def\csname jl@/comparisons/1/valid_replicates\endcsname{10000.000}
\expandafter\def\csname jl@/comparisons/2/ci_high\endcsname{0.097}
\expandafter\def\csname jl@/comparisons/2/ci_low\endcsname{0.047}
\expandafter\def\csname jl@/comparisons/2/clusters\endcsname{20.000}
\expandafter\def\csname jl@/comparisons/2/delta\endcsname{0.074}
\expandafter\def\csname jl@/comparisons/2/dropped_items\endcsname{0.000}
\expandafter\def\csname jl@/comparisons/2/iterations\endcsname{10000.000}
\expandafter\def\csname jl@/comparisons/2/p_holm\endcsname{0.001}
\expandafter\def\csname jl@/comparisons/2/p_value\endcsname{0.000}
\expandafter\def\csname jl@/comparisons/2/paired_items\endcsname{1280.000}
\expandafter\def\csname jl@/comparisons/2/valid_replicates\endcsname{10000.000}
\expandafter\def\csname jl@/comparisons/3/ci_high\endcsname{0.007}
\expandafter\def\csname jl@/comparisons/3/ci_low\endcsname{-0.031}
\expandafter\def\csname jl@/comparisons/3/clusters\endcsname{705.000}
\expandafter\def\csname jl@/comparisons/3/delta\endcsname{-0.012}
\expandafter\def\csname jl@/comparisons/3/dropped_items\endcsname{7.000}
\expandafter\def\csname jl@/comparisons/3/iterations\endcsname{10000.000}
\expandafter\def\csname jl@/comparisons/3/p_holm\endcsname{1.000}
\expandafter\def\csname jl@/comparisons/3/p_value\endcsname{0.215}
\expandafter\def\csname jl@/comparisons/3/paired_items\endcsname{792.000}
\expandafter\def\csname jl@/comparisons/3/valid_replicates\endcsname{10000.000}
\expandafter\def\csname jl@/comparisons/4/ci_high\endcsname{-0.003}
\expandafter\def\csname jl@/comparisons/4/ci_low\endcsname{-0.043}
\expandafter\def\csname jl@/comparisons/4/clusters\endcsname{591.000}
\expandafter\def\csname jl@/comparisons/4/delta\endcsname{-0.023}
\expandafter\def\csname jl@/comparisons/4/dropped_items\endcsname{148.000}
\expandafter\def\csname jl@/comparisons/4/iterations\endcsname{10000.000}
\expandafter\def\csname jl@/comparisons/4/p_holm\endcsname{0.155}
\expandafter\def\csname jl@/comparisons/4/p_value\endcsname{0.026}
\expandafter\def\csname jl@/comparisons/4/paired_items\endcsname{651.000}
\expandafter\def\csname jl@/comparisons/4/valid_replicates\endcsname{10000.000}
\expandafter\def\csname jl@/comparisons/5/ci_high\endcsname{0.023}
\expandafter\def\csname jl@/comparisons/5/ci_low\endcsname{-0.017}
\expandafter\def\csname jl@/comparisons/5/clusters\endcsname{705.000}
\expandafter\def\csname jl@/comparisons/5/delta\endcsname{0.003}
\expandafter\def\csname jl@/comparisons/5/dropped_items\endcsname{7.000}
\expandafter\def\csname jl@/comparisons/5/iterations\endcsname{10000.000}
\expandafter\def\csname jl@/comparisons/5/p_holm\endcsname{1.000}
\expandafter\def\csname jl@/comparisons/5/p_value\endcsname{0.797}
\expandafter\def\csname jl@/comparisons/5/paired_items\endcsname{792.000}
\expandafter\def\csname jl@/comparisons/5/valid_replicates\endcsname{10000.000}
\expandafter\def\csname jl@/comparisons/6/ci_high\endcsname{0.047}
\expandafter\def\csname jl@/comparisons/6/ci_low\endcsname{0.007}
\expandafter\def\csname jl@/comparisons/6/clusters\endcsname{48.000}
\expandafter\def\csname jl@/comparisons/6/delta\endcsname{0.027}
\expandafter\def\csname jl@/comparisons/6/dropped_items\endcsname{0.000}
\expandafter\def\csname jl@/comparisons/6/iterations\endcsname{10000.000}
\expandafter\def\csname jl@/comparisons/6/p_holm\endcsname{0.070}
\expandafter\def\csname jl@/comparisons/6/p_value\endcsname{0.010}
\expandafter\def\csname jl@/comparisons/6/paired_items\endcsname{1728.000}
\expandafter\def\csname jl@/comparisons/6/valid_replicates\endcsname{10000.000}
\expandafter\def\csname jl@/comparisons/7/ci_high\endcsname{0.023}
\expandafter\def\csname jl@/comparisons/7/ci_low\endcsname{-0.011}
\expandafter\def\csname jl@/comparisons/7/clusters\endcsname{48.000}
\expandafter\def\csname jl@/comparisons/7/delta\endcsname{0.005}
\expandafter\def\csname jl@/comparisons/7/dropped_items\endcsname{0.000}
\expandafter\def\csname jl@/comparisons/7/iterations\endcsname{10000.000}
\expandafter\def\csname jl@/comparisons/7/p_holm\endcsname{1.000}
\expandafter\def\csname jl@/comparisons/7/p_value\endcsname{0.568}
\expandafter\def\csname jl@/comparisons/7/paired_items\endcsname{1728.000}
\expandafter\def\csname jl@/comparisons/7/valid_replicates\endcsname{10000.000}
\expandafter\def\csname jl@/comparisons/8/ci_high\endcsname{0.023}
\expandafter\def\csname jl@/comparisons/8/ci_low\endcsname{-0.009}
\expandafter\def\csname jl@/comparisons/8/clusters\endcsname{48.000}
\expandafter\def\csname jl@/comparisons/8/delta\endcsname{0.007}
\expandafter\def\csname jl@/comparisons/8/dropped_items\endcsname{0.000}
\expandafter\def\csname jl@/comparisons/8/iterations\endcsname{10000.000}
\expandafter\def\csname jl@/comparisons/8/p_holm\endcsname{1.000}
\expandafter\def\csname jl@/comparisons/8/p_value\endcsname{0.397}
\expandafter\def\csname jl@/comparisons/8/paired_items\endcsname{1728.000}
\expandafter\def\csname jl@/comparisons/8/valid_replicates\endcsname{10000.000}
\expandafter\def\csname jl@/cost_usd/anthropic\endcsname{243.798}
\expandafter\def\csname jl@/cost_usd/others\endcsname{9.535}
\expandafter\def\csname jl@/dataset_summary/0/alpha_macro\endcsname{0.577}
\expandafter\def\csname jl@/dataset_summary/0/defined_dimensions\endcsname{4.000}
\expandafter\def\csname jl@/dataset_summary/0/total_dimensions\endcsname{4.000}
\expandafter\def\csname jl@/dataset_summary/1/alpha_macro\endcsname{0.541}
\expandafter\def\csname jl@/dataset_summary/1/defined_dimensions\endcsname{4.000}
\expandafter\def\csname jl@/dataset_summary/1/total_dimensions\endcsname{4.000}
\expandafter\def\csname jl@/dataset_summary/10/alpha_macro\endcsname{0.362}
\expandafter\def\csname jl@/dataset_summary/10/defined_dimensions\endcsname{1.000}
\expandafter\def\csname jl@/dataset_summary/10/total_dimensions\endcsname{1.000}
\expandafter\def\csname jl@/dataset_summary/11/alpha_macro\endcsname{0.359}
\expandafter\def\csname jl@/dataset_summary/11/defined_dimensions\endcsname{1.000}
\expandafter\def\csname jl@/dataset_summary/11/total_dimensions\endcsname{1.000}
\expandafter\def\csname jl@/dataset_summary/12/alpha_macro\endcsname{0.659}
\expandafter\def\csname jl@/dataset_summary/12/defined_dimensions\endcsname{6.000}
\expandafter\def\csname jl@/dataset_summary/12/total_dimensions\endcsname{6.000}
\expandafter\def\csname jl@/dataset_summary/13/alpha_macro\endcsname{0.670}
\expandafter\def\csname jl@/dataset_summary/13/defined_dimensions\endcsname{6.000}
\expandafter\def\csname jl@/dataset_summary/13/total_dimensions\endcsname{6.000}
\expandafter\def\csname jl@/dataset_summary/14/alpha_macro\endcsname{0.647}
\expandafter\def\csname jl@/dataset_summary/14/defined_dimensions\endcsname{6.000}
\expandafter\def\csname jl@/dataset_summary/14/total_dimensions\endcsname{6.000}
\expandafter\def\csname jl@/dataset_summary/15/alpha_macro\endcsname{0.663}
\expandafter\def\csname jl@/dataset_summary/15/defined_dimensions\endcsname{6.000}
\expandafter\def\csname jl@/dataset_summary/15/total_dimensions\endcsname{6.000}
\expandafter\def\csname jl@/dataset_summary/16/alpha_macro\endcsname{0.685}
\expandafter\def\csname jl@/dataset_summary/16/defined_dimensions\endcsname{6.000}
\expandafter\def\csname jl@/dataset_summary/16/total_dimensions\endcsname{6.000}
\expandafter\def\csname jl@/dataset_summary/17/alpha_macro\endcsname{0.678}
\expandafter\def\csname jl@/dataset_summary/17/defined_dimensions\endcsname{6.000}
\expandafter\def\csname jl@/dataset_summary/17/total_dimensions\endcsname{6.000}
\expandafter\def\csname jl@/dataset_summary/2/alpha_macro\endcsname{0.570}
\expandafter\def\csname jl@/dataset_summary/2/defined_dimensions\endcsname{4.000}
\expandafter\def\csname jl@/dataset_summary/2/total_dimensions\endcsname{4.000}
\expandafter\def\csname jl@/dataset_summary/3/alpha_macro\endcsname{0.556}
\expandafter\def\csname jl@/dataset_summary/3/defined_dimensions\endcsname{4.000}
\expandafter\def\csname jl@/dataset_summary/3/total_dimensions\endcsname{4.000}
\expandafter\def\csname jl@/dataset_summary/4/alpha_macro\endcsname{0.646}
\expandafter\def\csname jl@/dataset_summary/4/defined_dimensions\endcsname{4.000}
\expandafter\def\csname jl@/dataset_summary/4/total_dimensions\endcsname{4.000}
\expandafter\def\csname jl@/dataset_summary/5/alpha_macro\endcsname{0.572}
\expandafter\def\csname jl@/dataset_summary/5/defined_dimensions\endcsname{4.000}
\expandafter\def\csname jl@/dataset_summary/5/total_dimensions\endcsname{4.000}
\expandafter\def\csname jl@/dataset_summary/6/alpha_macro\endcsname{0.375}
\expandafter\def\csname jl@/dataset_summary/6/defined_dimensions\endcsname{1.000}
\expandafter\def\csname jl@/dataset_summary/6/total_dimensions\endcsname{1.000}
\expandafter\def\csname jl@/dataset_summary/7/alpha_macro\endcsname{0.369}
\expandafter\def\csname jl@/dataset_summary/7/defined_dimensions\endcsname{1.000}
\expandafter\def\csname jl@/dataset_summary/7/total_dimensions\endcsname{1.000}
\expandafter\def\csname jl@/dataset_summary/8/alpha_macro\endcsname{0.389}
\expandafter\def\csname jl@/dataset_summary/8/defined_dimensions\endcsname{1.000}
\expandafter\def\csname jl@/dataset_summary/8/total_dimensions\endcsname{1.000}
\expandafter\def\csname jl@/dataset_summary/9/alpha_macro\endcsname{0.358}
\expandafter\def\csname jl@/dataset_summary/9/defined_dimensions\endcsname{1.000}
\expandafter\def\csname jl@/dataset_summary/9/total_dimensions\endcsname{1.000}
\expandafter\def\csname jl@/failures/external/failure_rate\endcsname{0.000}
\expandafter\def\csname jl@/failures/external/parse_error\endcsname{1.000}
\expandafter\def\csname jl@/failures/external/parse_error_rate\endcsname{0.000}
\expandafter\def\csname jl@/failures/external/provider_error\endcsname{0.000}
\expandafter\def\csname jl@/failures/external/requests\endcsname{19035.000}
\expandafter\def\csname jl@/failures/full/failure_rate\endcsname{0.000}
\expandafter\def\csname jl@/failures/full/parse_error\endcsname{9.000}
\expandafter\def\csname jl@/failures/full/parse_error_rate\endcsname{0.000}
\expandafter\def\csname jl@/failures/full/provider_error\endcsname{0.000}
\expandafter\def\csname jl@/failures/full/requests\endcsname{26815.000}
\expandafter\def\csname jl@/failures/original/failure_rate\endcsname{0.000}
\expandafter\def\csname jl@/failures/original/parse_error\endcsname{3.000}
\expandafter\def\csname jl@/failures/original/parse_error_rate\endcsname{0.000}
\expandafter\def\csname jl@/failures/original/provider_error\endcsname{0.000}
\expandafter\def\csname jl@/failures/original/requests\endcsname{26815.000}
\expandafter\def\csname jl@/failures/random_clause/failure_rate\endcsname{0.000}
\expandafter\def\csname jl@/failures/random_clause/parse_error\endcsname{3.000}
\expandafter\def\csname jl@/failures/random_clause/parse_error_rate\endcsname{0.000}
\expandafter\def\csname jl@/failures/random_clause/provider_error\endcsname{0.000}
\expandafter\def\csname jl@/failures/random_clause/requests\endcsname{19035.000}
\expandafter\def\csname jl@/failures/rewrite/failure_rate\endcsname{0.000}
\expandafter\def\csname jl@/failures/rewrite/parse_error\endcsname{7.000}
\expandafter\def\csname jl@/failures/rewrite/parse_error_rate\endcsname{0.000}
\expandafter\def\csname jl@/failures/rewrite/provider_error\endcsname{0.000}
\expandafter\def\csname jl@/failures/rewrite/requests\endcsname{19035.000}
\expandafter\def\csname jl@/failures/stage1/failure_rate\endcsname{0.000}
\expandafter\def\csname jl@/failures/stage1/parse_error\endcsname{4.000}
\expandafter\def\csname jl@/failures/stage1/parse_error_rate\endcsname{0.000}
\expandafter\def\csname jl@/failures/stage1/provider_error\endcsname{0.000}
\expandafter\def\csname jl@/failures/stage1/requests\endcsname{19035.000}
\expandafter\def\csname jl@/human/0/kendall_tau\endcsname{0.566}
\expandafter\def\csname jl@/human/0/n\endcsname{320.000}
\expandafter\def\csname jl@/human/0/spearman_rho\endcsname{0.663}
\expandafter\def\csname jl@/human/1/kendall_tau\endcsname{0.521}
\expandafter\def\csname jl@/human/1/n\endcsname{320.000}
\expandafter\def\csname jl@/human/1/spearman_rho\endcsname{0.623}
\expandafter\def\csname jl@/human/10/kendall_tau\endcsname{0.660}
\expandafter\def\csname jl@/human/10/n\endcsname{320.000}
\expandafter\def\csname jl@/human/10/spearman_rho\endcsname{0.694}
\expandafter\def\csname jl@/human/100/kendall_tau\endcsname{0.578}
\expandafter\def\csname jl@/human/100/n\endcsname{320.000}
\expandafter\def\csname jl@/human/100/spearman_rho\endcsname{0.687}
\expandafter\def\csname jl@/human/101/kendall_tau\endcsname{0.550}
\expandafter\def\csname jl@/human/101/n\endcsname{320.000}
\expandafter\def\csname jl@/human/101/spearman_rho\endcsname{0.659}
\expandafter\def\csname jl@/human/102/kendall_tau\endcsname{0.501}
\expandafter\def\csname jl@/human/102/n\endcsname{320.000}
\expandafter\def\csname jl@/human/102/spearman_rho\endcsname{0.604}
\expandafter\def\csname jl@/human/103/kendall_tau\endcsname{0.516}
\expandafter\def\csname jl@/human/103/n\endcsname{320.000}
\expandafter\def\csname jl@/human/103/spearman_rho\endcsname{0.624}
\expandafter\def\csname jl@/human/104/kendall_tau\endcsname{0.529}
\expandafter\def\csname jl@/human/104/n\endcsname{320.000}
\expandafter\def\csname jl@/human/104/spearman_rho\endcsname{0.639}
\expandafter\def\csname jl@/human/105/kendall_tau\endcsname{0.546}
\expandafter\def\csname jl@/human/105/n\endcsname{320.000}
\expandafter\def\csname jl@/human/105/spearman_rho\endcsname{0.698}
\expandafter\def\csname jl@/human/106/kendall_tau\endcsname{0.659}
\expandafter\def\csname jl@/human/106/n\endcsname{320.000}
\expandafter\def\csname jl@/human/106/spearman_rho\endcsname{0.693}
\expandafter\def\csname jl@/human/107/kendall_tau\endcsname{0.645}
\expandafter\def\csname jl@/human/107/n\endcsname{320.000}
\expandafter\def\csname jl@/human/107/spearman_rho\endcsname{0.683}
\expandafter\def\csname jl@/human/108/kendall_tau\endcsname{0.544}
\expandafter\def\csname jl@/human/108/n\endcsname{320.000}
\expandafter\def\csname jl@/human/108/spearman_rho\endcsname{0.578}
\expandafter\def\csname jl@/human/109/kendall_tau\endcsname{0.653}
\expandafter\def\csname jl@/human/109/n\endcsname{320.000}
\expandafter\def\csname jl@/human/109/spearman_rho\endcsname{0.690}
\expandafter\def\csname jl@/human/11/kendall_tau\endcsname{0.622}
\expandafter\def\csname jl@/human/11/n\endcsname{320.000}
\expandafter\def\csname jl@/human/11/spearman_rho\endcsname{0.652}
\expandafter\def\csname jl@/human/110/kendall_tau\endcsname{0.578}
\expandafter\def\csname jl@/human/110/n\endcsname{320.000}
\expandafter\def\csname jl@/human/110/spearman_rho\endcsname{0.607}
\expandafter\def\csname jl@/human/111/kendall_tau\endcsname{0.556}
\expandafter\def\csname jl@/human/111/n\endcsname{320.000}
\expandafter\def\csname jl@/human/111/spearman_rho\endcsname{0.607}
\expandafter\def\csname jl@/human/112/kendall_tau\endcsname{0.547}
\expandafter\def\csname jl@/human/112/n\endcsname{320.000}
\expandafter\def\csname jl@/human/112/spearman_rho\endcsname{0.618}
\expandafter\def\csname jl@/human/113/kendall_tau\endcsname{0.419}
\expandafter\def\csname jl@/human/113/n\endcsname{320.000}
\expandafter\def\csname jl@/human/113/spearman_rho\endcsname{0.486}
\expandafter\def\csname jl@/human/114/kendall_tau\endcsname{0.338}
\expandafter\def\csname jl@/human/114/n\endcsname{320.000}
\expandafter\def\csname jl@/human/114/spearman_rho\endcsname{0.371}
\expandafter\def\csname jl@/human/115/kendall_tau\endcsname{0.489}
\expandafter\def\csname jl@/human/115/n\endcsname{320.000}
\expandafter\def\csname jl@/human/115/spearman_rho\endcsname{0.547}
\expandafter\def\csname jl@/human/116/kendall_tau\endcsname{0.230}
\expandafter\def\csname jl@/human/116/n\endcsname{320.000}
\expandafter\def\csname jl@/human/116/spearman_rho\endcsname{0.246}
\expandafter\def\csname jl@/human/117/kendall_tau\endcsname{0.437}
\expandafter\def\csname jl@/human/117/n\endcsname{320.000}
\expandafter\def\csname jl@/human/117/spearman_rho\endcsname{0.528}
\expandafter\def\csname jl@/human/118/kendall_tau\endcsname{0.496}
\expandafter\def\csname jl@/human/118/n\endcsname{320.000}
\expandafter\def\csname jl@/human/118/spearman_rho\endcsname{0.580}
\expandafter\def\csname jl@/human/119/kendall_tau\endcsname{0.468}
\expandafter\def\csname jl@/human/119/n\endcsname{320.000}
\expandafter\def\csname jl@/human/119/spearman_rho\endcsname{0.552}
\expandafter\def\csname jl@/human/12/kendall_tau\endcsname{0.571}
\expandafter\def\csname jl@/human/12/n\endcsname{320.000}
\expandafter\def\csname jl@/human/12/spearman_rho\endcsname{0.627}
\expandafter\def\csname jl@/human/120/kendall_tau\endcsname{0.394}
\expandafter\def\csname jl@/human/120/n\endcsname{320.000}
\expandafter\def\csname jl@/human/120/spearman_rho\endcsname{0.480}
\expandafter\def\csname jl@/human/121/kendall_tau\endcsname{0.338}
\expandafter\def\csname jl@/human/121/n\endcsname{320.000}
\expandafter\def\csname jl@/human/121/spearman_rho\endcsname{0.418}
\expandafter\def\csname jl@/human/122/kendall_tau\endcsname{0.467}
\expandafter\def\csname jl@/human/122/n\endcsname{320.000}
\expandafter\def\csname jl@/human/122/spearman_rho\endcsname{0.566}
\expandafter\def\csname jl@/human/123/kendall_tau\endcsname{0.437}
\expandafter\def\csname jl@/human/123/n\endcsname{320.000}
\expandafter\def\csname jl@/human/123/spearman_rho\endcsname{0.563}
\expandafter\def\csname jl@/human/124/kendall_tau\endcsname{0.563}
\expandafter\def\csname jl@/human/124/n\endcsname{320.000}
\expandafter\def\csname jl@/human/124/spearman_rho\endcsname{0.673}
\expandafter\def\csname jl@/human/125/kendall_tau\endcsname{0.553}
\expandafter\def\csname jl@/human/125/n\endcsname{320.000}
\expandafter\def\csname jl@/human/125/spearman_rho\endcsname{0.671}
\expandafter\def\csname jl@/human/126/kendall_tau\endcsname{0.493}
\expandafter\def\csname jl@/human/126/n\endcsname{320.000}
\expandafter\def\csname jl@/human/126/spearman_rho\endcsname{0.596}
\expandafter\def\csname jl@/human/127/kendall_tau\endcsname{0.520}
\expandafter\def\csname jl@/human/127/n\endcsname{320.000}
\expandafter\def\csname jl@/human/127/spearman_rho\endcsname{0.627}
\expandafter\def\csname jl@/human/128/kendall_tau\endcsname{0.529}
\expandafter\def\csname jl@/human/128/n\endcsname{320.000}
\expandafter\def\csname jl@/human/128/spearman_rho\endcsname{0.630}
\expandafter\def\csname jl@/human/129/kendall_tau\endcsname{0.540}
\expandafter\def\csname jl@/human/129/n\endcsname{320.000}
\expandafter\def\csname jl@/human/129/spearman_rho\endcsname{0.692}
\expandafter\def\csname jl@/human/13/kendall_tau\endcsname{0.665}
\expandafter\def\csname jl@/human/13/n\endcsname{304.000}
\expandafter\def\csname jl@/human/13/spearman_rho\endcsname{0.696}
\expandafter\def\csname jl@/human/130/kendall_tau\endcsname{0.631}
\expandafter\def\csname jl@/human/130/n\endcsname{320.000}
\expandafter\def\csname jl@/human/130/spearman_rho\endcsname{0.669}
\expandafter\def\csname jl@/human/131/kendall_tau\endcsname{0.627}
\expandafter\def\csname jl@/human/131/n\endcsname{320.000}
\expandafter\def\csname jl@/human/131/spearman_rho\endcsname{0.667}
\expandafter\def\csname jl@/human/132/kendall_tau\endcsname{0.547}
\expandafter\def\csname jl@/human/132/n\endcsname{320.000}
\expandafter\def\csname jl@/human/132/spearman_rho\endcsname{0.586}
\expandafter\def\csname jl@/human/133/kendall_tau\endcsname{0.636}
\expandafter\def\csname jl@/human/133/n\endcsname{320.000}
\expandafter\def\csname jl@/human/133/spearman_rho\endcsname{0.672}
\expandafter\def\csname jl@/human/134/kendall_tau\endcsname{0.589}
\expandafter\def\csname jl@/human/134/n\endcsname{320.000}
\expandafter\def\csname jl@/human/134/spearman_rho\endcsname{0.621}
\expandafter\def\csname jl@/human/135/kendall_tau\endcsname{0.559}
\expandafter\def\csname jl@/human/135/n\endcsname{320.000}
\expandafter\def\csname jl@/human/135/spearman_rho\endcsname{0.618}
\expandafter\def\csname jl@/human/136/kendall_tau\endcsname{0.542}
\expandafter\def\csname jl@/human/136/n\endcsname{320.000}
\expandafter\def\csname jl@/human/136/spearman_rho\endcsname{0.593}
\expandafter\def\csname jl@/human/137/kendall_tau\endcsname{0.627}
\expandafter\def\csname jl@/human/137/n\endcsname{320.000}
\expandafter\def\csname jl@/human/137/spearman_rho\endcsname{0.677}
\expandafter\def\csname jl@/human/138/kendall_tau\endcsname{0.135}
\expandafter\def\csname jl@/human/138/n\endcsname{320.000}
\expandafter\def\csname jl@/human/138/spearman_rho\endcsname{0.152}
\expandafter\def\csname jl@/human/139/kendall_tau\endcsname{0.567}
\expandafter\def\csname jl@/human/139/n\endcsname{320.000}
\expandafter\def\csname jl@/human/139/spearman_rho\endcsname{0.612}
\expandafter\def\csname jl@/human/14/kendall_tau\endcsname{0.582}
\expandafter\def\csname jl@/human/14/n\endcsname{320.000}
\expandafter\def\csname jl@/human/14/spearman_rho\endcsname{0.642}
\expandafter\def\csname jl@/human/140/kendall_tau\endcsname{0.433}
\expandafter\def\csname jl@/human/140/n\endcsname{320.000}
\expandafter\def\csname jl@/human/140/spearman_rho\endcsname{0.469}
\expandafter\def\csname jl@/human/141/kendall_tau\endcsname{0.510}
\expandafter\def\csname jl@/human/141/n\endcsname{320.000}
\expandafter\def\csname jl@/human/141/spearman_rho\endcsname{0.601}
\expandafter\def\csname jl@/human/142/kendall_tau\endcsname{0.456}
\expandafter\def\csname jl@/human/142/n\endcsname{320.000}
\expandafter\def\csname jl@/human/142/spearman_rho\endcsname{0.538}
\expandafter\def\csname jl@/human/143/kendall_tau\endcsname{0.495}
\expandafter\def\csname jl@/human/143/n\endcsname{320.000}
\expandafter\def\csname jl@/human/143/spearman_rho\endcsname{0.581}
\expandafter\def\csname jl@/human/144/kendall_tau\endcsname{0.421}
\expandafter\def\csname jl@/human/144/n\endcsname{320.000}
\expandafter\def\csname jl@/human/144/spearman_rho\endcsname{0.503}
\expandafter\def\csname jl@/human/145/kendall_tau\endcsname{0.380}
\expandafter\def\csname jl@/human/145/n\endcsname{320.000}
\expandafter\def\csname jl@/human/145/spearman_rho\endcsname{0.468}
\expandafter\def\csname jl@/human/146/kendall_tau\endcsname{0.444}
\expandafter\def\csname jl@/human/146/n\endcsname{320.000}
\expandafter\def\csname jl@/human/146/spearman_rho\endcsname{0.535}
\expandafter\def\csname jl@/human/147/kendall_tau\endcsname{0.454}
\expandafter\def\csname jl@/human/147/n\endcsname{320.000}
\expandafter\def\csname jl@/human/147/spearman_rho\endcsname{0.583}
\expandafter\def\csname jl@/human/148/cohen_kappa\endcsname{0.325}
\expandafter\def\csname jl@/human/148/macro_f1\endcsname{0.634}
\expandafter\def\csname jl@/human/148/n\endcsname{654.000}
\expandafter\def\csname jl@/human/149/cohen_kappa\endcsname{0.285}
\expandafter\def\csname jl@/human/149/macro_f1\endcsname{0.605}
\expandafter\def\csname jl@/human/149/n\endcsname{656.000}
\expandafter\def\csname jl@/human/15/kendall_tau\endcsname{0.419}
\expandafter\def\csname jl@/human/15/n\endcsname{320.000}
\expandafter\def\csname jl@/human/15/spearman_rho\endcsname{0.468}
\expandafter\def\csname jl@/human/150/cohen_kappa\endcsname{0.460}
\expandafter\def\csname jl@/human/150/macro_f1\endcsname{0.730}
\expandafter\def\csname jl@/human/150/n\endcsname{656.000}
\expandafter\def\csname jl@/human/151/cohen_kappa\endcsname{0.274}
\expandafter\def\csname jl@/human/151/macro_f1\endcsname{0.606}
\expandafter\def\csname jl@/human/151/n\endcsname{655.000}
\expandafter\def\csname jl@/human/152/cohen_kappa\endcsname{0.491}
\expandafter\def\csname jl@/human/152/macro_f1\endcsname{0.745}
\expandafter\def\csname jl@/human/152/n\endcsname{656.000}
\expandafter\def\csname jl@/human/153/cohen_kappa\endcsname{0.388}
\expandafter\def\csname jl@/human/153/macro_f1\endcsname{0.681}
\expandafter\def\csname jl@/human/153/n\endcsname{654.000}
\expandafter\def\csname jl@/human/154/cohen_kappa\endcsname{0.388}
\expandafter\def\csname jl@/human/154/macro_f1\endcsname{0.681}
\expandafter\def\csname jl@/human/154/n\endcsname{654.000}
\expandafter\def\csname jl@/human/155/cohen_kappa\endcsname{0.254}
\expandafter\def\csname jl@/human/155/macro_f1\endcsname{0.583}
\expandafter\def\csname jl@/human/155/n\endcsname{655.000}
\expandafter\def\csname jl@/human/156/cohen_kappa\endcsname{0.180}
\expandafter\def\csname jl@/human/156/macro_f1\endcsname{0.506}
\expandafter\def\csname jl@/human/156/n\endcsname{656.000}
\expandafter\def\csname jl@/human/157/cohen_kappa\endcsname{0.385}
\expandafter\def\csname jl@/human/157/macro_f1\endcsname{0.683}
\expandafter\def\csname jl@/human/157/n\endcsname{656.000}
\expandafter\def\csname jl@/human/158/cohen_kappa\endcsname{0.204}
\expandafter\def\csname jl@/human/158/macro_f1\endcsname{0.544}
\expandafter\def\csname jl@/human/158/n\endcsname{656.000}
\expandafter\def\csname jl@/human/159/cohen_kappa\endcsname{0.571}
\expandafter\def\csname jl@/human/159/macro_f1\endcsname{0.785}
\expandafter\def\csname jl@/human/159/n\endcsname{656.000}
\expandafter\def\csname jl@/human/16/kendall_tau\endcsname{0.346}
\expandafter\def\csname jl@/human/16/n\endcsname{320.000}
\expandafter\def\csname jl@/human/16/spearman_rho\endcsname{0.386}
\expandafter\def\csname jl@/human/160/cohen_kappa\endcsname{0.305}
\expandafter\def\csname jl@/human/160/macro_f1\endcsname{0.619}
\expandafter\def\csname jl@/human/160/n\endcsname{655.000}
\expandafter\def\csname jl@/human/161/cohen_kappa\endcsname{0.345}
\expandafter\def\csname jl@/human/161/macro_f1\endcsname{0.647}
\expandafter\def\csname jl@/human/161/n\endcsname{654.000}
\expandafter\def\csname jl@/human/162/cohen_kappa\endcsname{0.273}
\expandafter\def\csname jl@/human/162/macro_f1\endcsname{0.593}
\expandafter\def\csname jl@/human/162/n\endcsname{656.000}
\expandafter\def\csname jl@/human/163/cohen_kappa\endcsname{0.500}
\expandafter\def\csname jl@/human/163/macro_f1\endcsname{0.750}
\expandafter\def\csname jl@/human/163/n\endcsname{656.000}
\expandafter\def\csname jl@/human/164/cohen_kappa\endcsname{0.266}
\expandafter\def\csname jl@/human/164/macro_f1\endcsname{0.602}
\expandafter\def\csname jl@/human/164/n\endcsname{655.000}
\expandafter\def\csname jl@/human/165/cohen_kappa\endcsname{0.487}
\expandafter\def\csname jl@/human/165/macro_f1\endcsname{0.743}
\expandafter\def\csname jl@/human/165/n\endcsname{656.000}
\expandafter\def\csname jl@/human/166/cohen_kappa\endcsname{0.403}
\expandafter\def\csname jl@/human/166/macro_f1\endcsname{0.689}
\expandafter\def\csname jl@/human/166/n\endcsname{653.000}
\expandafter\def\csname jl@/human/167/cohen_kappa\endcsname{0.303}
\expandafter\def\csname jl@/human/167/macro_f1\endcsname{0.617}
\expandafter\def\csname jl@/human/167/n\endcsname{654.000}
\expandafter\def\csname jl@/human/168/cohen_kappa\endcsname{0.266}
\expandafter\def\csname jl@/human/168/macro_f1\endcsname{0.587}
\expandafter\def\csname jl@/human/168/n\endcsname{656.000}
\expandafter\def\csname jl@/human/169/cohen_kappa\endcsname{0.463}
\expandafter\def\csname jl@/human/169/macro_f1\endcsname{0.731}
\expandafter\def\csname jl@/human/169/n\endcsname{656.000}
\expandafter\def\csname jl@/human/17/kendall_tau\endcsname{0.392}
\expandafter\def\csname jl@/human/17/n\endcsname{320.000}
\expandafter\def\csname jl@/human/17/spearman_rho\endcsname{0.427}
\expandafter\def\csname jl@/human/170/cohen_kappa\endcsname{0.278}
\expandafter\def\csname jl@/human/170/macro_f1\endcsname{0.608}
\expandafter\def\csname jl@/human/170/n\endcsname{655.000}
\expandafter\def\csname jl@/human/171/cohen_kappa\endcsname{0.495}
\expandafter\def\csname jl@/human/171/macro_f1\endcsname{0.746}
\expandafter\def\csname jl@/human/171/n\endcsname{656.000}
\expandafter\def\csname jl@/human/172/cohen_kappa\endcsname{0.389}
\expandafter\def\csname jl@/human/172/macro_f1\endcsname{0.681}
\expandafter\def\csname jl@/human/172/n\endcsname{653.000}
\expandafter\def\csname jl@/human/173/cohen_kappa\endcsname{0.303}
\expandafter\def\csname jl@/human/173/macro_f1\endcsname{0.617}
\expandafter\def\csname jl@/human/173/n\endcsname{654.000}
\expandafter\def\csname jl@/human/174/cohen_kappa\endcsname{0.276}
\expandafter\def\csname jl@/human/174/macro_f1\endcsname{0.590}
\expandafter\def\csname jl@/human/174/n\endcsname{656.000}
\expandafter\def\csname jl@/human/175/cohen_kappa\endcsname{0.438}
\expandafter\def\csname jl@/human/175/macro_f1\endcsname{0.718}
\expandafter\def\csname jl@/human/175/n\endcsname{656.000}
\expandafter\def\csname jl@/human/176/cohen_kappa\endcsname{0.260}
\expandafter\def\csname jl@/human/176/macro_f1\endcsname{0.597}
\expandafter\def\csname jl@/human/176/n\endcsname{653.000}
\expandafter\def\csname jl@/human/177/cohen_kappa\endcsname{0.434}
\expandafter\def\csname jl@/human/177/macro_f1\endcsname{0.716}
\expandafter\def\csname jl@/human/177/n\endcsname{656.000}
\expandafter\def\csname jl@/human/178/cohen_kappa\endcsname{0.352}
\expandafter\def\csname jl@/human/178/macro_f1\endcsname{0.659}
\expandafter\def\csname jl@/human/178/n\endcsname{651.000}
\expandafter\def\csname jl@/human/179/cohen_kappa\endcsname{0.309}
\expandafter\def\csname jl@/human/179/macro_f1\endcsname{0.619}
\expandafter\def\csname jl@/human/179/n\endcsname{655.000}
\expandafter\def\csname jl@/human/18/kendall_tau\endcsname{0.382}
\expandafter\def\csname jl@/human/18/n\endcsname{320.000}
\expandafter\def\csname jl@/human/18/spearman_rho\endcsname{0.416}
\expandafter\def\csname jl@/human/180/cohen_kappa\endcsname{0.267}
\expandafter\def\csname jl@/human/180/macro_f1\endcsname{0.588}
\expandafter\def\csname jl@/human/180/n\endcsname{656.000}
\expandafter\def\csname jl@/human/181/cohen_kappa\endcsname{0.477}
\expandafter\def\csname jl@/human/181/macro_f1\endcsname{0.738}
\expandafter\def\csname jl@/human/181/n\endcsname{656.000}
\expandafter\def\csname jl@/human/182/cohen_kappa\endcsname{0.252}
\expandafter\def\csname jl@/human/182/macro_f1\endcsname{0.588}
\expandafter\def\csname jl@/human/182/n\endcsname{655.000}
\expandafter\def\csname jl@/human/183/cohen_kappa\endcsname{0.506}
\expandafter\def\csname jl@/human/183/macro_f1\endcsname{0.752}
\expandafter\def\csname jl@/human/183/n\endcsname{656.000}
\expandafter\def\csname jl@/human/184/cohen_kappa\endcsname{0.367}
\expandafter\def\csname jl@/human/184/macro_f1\endcsname{0.665}
\expandafter\def\csname jl@/human/184/n\endcsname{654.000}
\expandafter\def\csname jl@/human/185/n\endcsname{288.000}
\expandafter\def\csname jl@/human/185/spearman_rho\endcsname{0.744}
\expandafter\def\csname jl@/human/186/n\endcsname{288.000}
\expandafter\def\csname jl@/human/186/spearman_rho\endcsname{0.764}
\expandafter\def\csname jl@/human/187/n\endcsname{288.000}
\expandafter\def\csname jl@/human/187/spearman_rho\endcsname{0.709}
\expandafter\def\csname jl@/human/188/n\endcsname{288.000}
\expandafter\def\csname jl@/human/188/spearman_rho\endcsname{0.588}
\expandafter\def\csname jl@/human/189/n\endcsname{288.000}
\expandafter\def\csname jl@/human/189/spearman_rho\endcsname{0.578}
\expandafter\def\csname jl@/human/19/kendall_tau\endcsname{0.451}
\expandafter\def\csname jl@/human/19/n\endcsname{320.000}
\expandafter\def\csname jl@/human/19/spearman_rho\endcsname{0.546}
\expandafter\def\csname jl@/human/190/n\endcsname{288.000}
\expandafter\def\csname jl@/human/190/spearman_rho\endcsname{0.748}
\expandafter\def\csname jl@/human/191/n\endcsname{283.000}
\expandafter\def\csname jl@/human/191/spearman_rho\endcsname{0.750}
\expandafter\def\csname jl@/human/192/n\endcsname{288.000}
\expandafter\def\csname jl@/human/192/spearman_rho\endcsname{0.730}
\expandafter\def\csname jl@/human/193/n\endcsname{288.000}
\expandafter\def\csname jl@/human/193/spearman_rho\endcsname{0.685}
\expandafter\def\csname jl@/human/194/n\endcsname{288.000}
\expandafter\def\csname jl@/human/194/spearman_rho\endcsname{0.573}
\expandafter\def\csname jl@/human/195/n\endcsname{288.000}
\expandafter\def\csname jl@/human/195/spearman_rho\endcsname{0.484}
\expandafter\def\csname jl@/human/196/n\endcsname{288.000}
\expandafter\def\csname jl@/human/196/spearman_rho\endcsname{0.489}
\expandafter\def\csname jl@/human/197/n\endcsname{288.000}
\expandafter\def\csname jl@/human/197/spearman_rho\endcsname{0.741}
\expandafter\def\csname jl@/human/198/n\endcsname{270.000}
\expandafter\def\csname jl@/human/198/spearman_rho\endcsname{0.701}
\expandafter\def\csname jl@/human/199/n\endcsname{288.000}
\expandafter\def\csname jl@/human/199/spearman_rho\endcsname{0.711}
\expandafter\def\csname jl@/human/2/kendall_tau\endcsname{0.486}
\expandafter\def\csname jl@/human/2/n\endcsname{320.000}
\expandafter\def\csname jl@/human/2/spearman_rho\endcsname{0.575}
\expandafter\def\csname jl@/human/20/kendall_tau\endcsname{0.499}
\expandafter\def\csname jl@/human/20/n\endcsname{156.000}
\expandafter\def\csname jl@/human/20/spearman_rho\endcsname{0.569}
\expandafter\def\csname jl@/human/200/n\endcsname{288.000}
\expandafter\def\csname jl@/human/200/spearman_rho\endcsname{0.704}
\expandafter\def\csname jl@/human/201/n\endcsname{288.000}
\expandafter\def\csname jl@/human/201/spearman_rho\endcsname{0.616}
\expandafter\def\csname jl@/human/202/n\endcsname{288.000}
\expandafter\def\csname jl@/human/202/spearman_rho\endcsname{0.612}
\expandafter\def\csname jl@/human/203/n\endcsname{288.000}
\expandafter\def\csname jl@/human/203/spearman_rho\endcsname{0.601}
\expandafter\def\csname jl@/human/204/n\endcsname{288.000}
\expandafter\def\csname jl@/human/204/spearman_rho\endcsname{0.735}
\expandafter\def\csname jl@/human/205/n\endcsname{281.000}
\expandafter\def\csname jl@/human/205/spearman_rho\endcsname{0.687}
\expandafter\def\csname jl@/human/206/n\endcsname{288.000}
\expandafter\def\csname jl@/human/206/spearman_rho\endcsname{0.820}
\expandafter\def\csname jl@/human/207/n\endcsname{288.000}
\expandafter\def\csname jl@/human/207/spearman_rho\endcsname{0.794}
\expandafter\def\csname jl@/human/208/n\endcsname{288.000}
\expandafter\def\csname jl@/human/208/spearman_rho\endcsname{0.651}
\expandafter\def\csname jl@/human/209/n\endcsname{288.000}
\expandafter\def\csname jl@/human/209/spearman_rho\endcsname{0.694}
\expandafter\def\csname jl@/human/21/kendall_tau\endcsname{0.452}
\expandafter\def\csname jl@/human/21/n\endcsname{320.000}
\expandafter\def\csname jl@/human/21/spearman_rho\endcsname{0.533}
\expandafter\def\csname jl@/human/210/n\endcsname{288.000}
\expandafter\def\csname jl@/human/210/spearman_rho\endcsname{0.682}
\expandafter\def\csname jl@/human/211/n\endcsname{288.000}
\expandafter\def\csname jl@/human/211/spearman_rho\endcsname{0.801}
\expandafter\def\csname jl@/human/212/n\endcsname{236.000}
\expandafter\def\csname jl@/human/212/spearman_rho\endcsname{0.817}
\expandafter\def\csname jl@/human/213/n\endcsname{288.000}
\expandafter\def\csname jl@/human/213/spearman_rho\endcsname{0.700}
\expandafter\def\csname jl@/human/214/n\endcsname{288.000}
\expandafter\def\csname jl@/human/214/spearman_rho\endcsname{0.568}
\expandafter\def\csname jl@/human/215/n\endcsname{288.000}
\expandafter\def\csname jl@/human/215/spearman_rho\endcsname{0.482}
\expandafter\def\csname jl@/human/216/n\endcsname{288.000}
\expandafter\def\csname jl@/human/216/spearman_rho\endcsname{0.334}
\expandafter\def\csname jl@/human/217/n\endcsname{288.000}
\expandafter\def\csname jl@/human/217/spearman_rho\endcsname{0.359}
\expandafter\def\csname jl@/human/218/n\endcsname{288.000}
\expandafter\def\csname jl@/human/218/spearman_rho\endcsname{0.704}
\expandafter\def\csname jl@/human/219/n\endcsname{288.000}
\expandafter\def\csname jl@/human/219/spearman_rho\endcsname{0.483}
\expandafter\def\csname jl@/human/22/kendall_tau\endcsname{0.501}
\expandafter\def\csname jl@/human/22/n\endcsname{320.000}
\expandafter\def\csname jl@/human/22/spearman_rho\endcsname{0.585}
\expandafter\def\csname jl@/human/220/n\endcsname{288.000}
\expandafter\def\csname jl@/human/220/spearman_rho\endcsname{0.573}
\expandafter\def\csname jl@/human/221/n\endcsname{288.000}
\expandafter\def\csname jl@/human/221/spearman_rho\endcsname{0.687}
\expandafter\def\csname jl@/human/222/n\endcsname{288.000}
\expandafter\def\csname jl@/human/222/spearman_rho\endcsname{0.610}
\expandafter\def\csname jl@/human/223/n\endcsname{288.000}
\expandafter\def\csname jl@/human/223/spearman_rho\endcsname{0.561}
\expandafter\def\csname jl@/human/224/n\endcsname{288.000}
\expandafter\def\csname jl@/human/224/spearman_rho\endcsname{0.566}
\expandafter\def\csname jl@/human/225/n\endcsname{288.000}
\expandafter\def\csname jl@/human/225/spearman_rho\endcsname{0.652}
\expandafter\def\csname jl@/human/226/n\endcsname{288.000}
\expandafter\def\csname jl@/human/226/spearman_rho\endcsname{0.633}
\expandafter\def\csname jl@/human/227/n\endcsname{288.000}
\expandafter\def\csname jl@/human/227/spearman_rho\endcsname{0.732}
\expandafter\def\csname jl@/human/228/n\endcsname{288.000}
\expandafter\def\csname jl@/human/228/spearman_rho\endcsname{0.731}
\expandafter\def\csname jl@/human/229/n\endcsname{288.000}
\expandafter\def\csname jl@/human/229/spearman_rho\endcsname{0.684}
\expandafter\def\csname jl@/human/23/kendall_tau\endcsname{0.379}
\expandafter\def\csname jl@/human/23/n\endcsname{320.000}
\expandafter\def\csname jl@/human/23/spearman_rho\endcsname{0.445}
\expandafter\def\csname jl@/human/230/n\endcsname{288.000}
\expandafter\def\csname jl@/human/230/spearman_rho\endcsname{0.596}
\expandafter\def\csname jl@/human/231/n\endcsname{288.000}
\expandafter\def\csname jl@/human/231/spearman_rho\endcsname{0.636}
\expandafter\def\csname jl@/human/232/n\endcsname{288.000}
\expandafter\def\csname jl@/human/232/spearman_rho\endcsname{0.745}
\expandafter\def\csname jl@/human/233/n\endcsname{288.000}
\expandafter\def\csname jl@/human/233/spearman_rho\endcsname{0.707}
\expandafter\def\csname jl@/human/234/n\endcsname{288.000}
\expandafter\def\csname jl@/human/234/spearman_rho\endcsname{0.678}
\expandafter\def\csname jl@/human/235/n\endcsname{288.000}
\expandafter\def\csname jl@/human/235/spearman_rho\endcsname{0.601}
\expandafter\def\csname jl@/human/236/n\endcsname{288.000}
\expandafter\def\csname jl@/human/236/spearman_rho\endcsname{0.472}
\expandafter\def\csname jl@/human/237/n\endcsname{288.000}
\expandafter\def\csname jl@/human/237/spearman_rho\endcsname{0.463}
\expandafter\def\csname jl@/human/238/n\endcsname{288.000}
\expandafter\def\csname jl@/human/238/spearman_rho\endcsname{0.730}
\expandafter\def\csname jl@/human/239/n\endcsname{288.000}
\expandafter\def\csname jl@/human/239/spearman_rho\endcsname{0.720}
\expandafter\def\csname jl@/human/24/kendall_tau\endcsname{0.364}
\expandafter\def\csname jl@/human/24/n\endcsname{320.000}
\expandafter\def\csname jl@/human/24/spearman_rho\endcsname{0.440}
\expandafter\def\csname jl@/human/240/n\endcsname{288.000}
\expandafter\def\csname jl@/human/240/spearman_rho\endcsname{0.672}
\expandafter\def\csname jl@/human/241/n\endcsname{288.000}
\expandafter\def\csname jl@/human/241/spearman_rho\endcsname{0.614}
\expandafter\def\csname jl@/human/242/n\endcsname{288.000}
\expandafter\def\csname jl@/human/242/spearman_rho\endcsname{0.618}
\expandafter\def\csname jl@/human/243/n\endcsname{288.000}
\expandafter\def\csname jl@/human/243/spearman_rho\endcsname{0.582}
\expandafter\def\csname jl@/human/244/n\endcsname{288.000}
\expandafter\def\csname jl@/human/244/spearman_rho\endcsname{0.715}
\expandafter\def\csname jl@/human/245/n\endcsname{288.000}
\expandafter\def\csname jl@/human/245/spearman_rho\endcsname{0.803}
\expandafter\def\csname jl@/human/246/n\endcsname{288.000}
\expandafter\def\csname jl@/human/246/spearman_rho\endcsname{0.818}
\expandafter\def\csname jl@/human/247/n\endcsname{288.000}
\expandafter\def\csname jl@/human/247/spearman_rho\endcsname{0.738}
\expandafter\def\csname jl@/human/248/n\endcsname{288.000}
\expandafter\def\csname jl@/human/248/spearman_rho\endcsname{0.687}
\expandafter\def\csname jl@/human/249/n\endcsname{288.000}
\expandafter\def\csname jl@/human/249/spearman_rho\endcsname{0.664}
\expandafter\def\csname jl@/human/25/kendall_tau\endcsname{0.407}
\expandafter\def\csname jl@/human/25/n\endcsname{320.000}
\expandafter\def\csname jl@/human/25/spearman_rho\endcsname{0.495}
\expandafter\def\csname jl@/human/250/n\endcsname{288.000}
\expandafter\def\csname jl@/human/250/spearman_rho\endcsname{0.803}
\expandafter\def\csname jl@/human/251/n\endcsname{288.000}
\expandafter\def\csname jl@/human/251/spearman_rho\endcsname{0.678}
\expandafter\def\csname jl@/human/252/n\endcsname{288.000}
\expandafter\def\csname jl@/human/252/spearman_rho\endcsname{0.543}
\expandafter\def\csname jl@/human/253/n\endcsname{288.000}
\expandafter\def\csname jl@/human/253/spearman_rho\endcsname{0.502}
\expandafter\def\csname jl@/human/254/n\endcsname{288.000}
\expandafter\def\csname jl@/human/254/spearman_rho\endcsname{0.370}
\expandafter\def\csname jl@/human/255/n\endcsname{288.000}
\expandafter\def\csname jl@/human/255/spearman_rho\endcsname{0.347}
\expandafter\def\csname jl@/human/256/n\endcsname{288.000}
\expandafter\def\csname jl@/human/256/spearman_rho\endcsname{0.728}
\expandafter\def\csname jl@/human/257/n\endcsname{288.000}
\expandafter\def\csname jl@/human/257/spearman_rho\endcsname{0.596}
\expandafter\def\csname jl@/human/258/n\endcsname{288.000}
\expandafter\def\csname jl@/human/258/spearman_rho\endcsname{0.665}
\expandafter\def\csname jl@/human/259/n\endcsname{288.000}
\expandafter\def\csname jl@/human/259/spearman_rho\endcsname{0.641}
\expandafter\def\csname jl@/human/26/kendall_tau\endcsname{0.437}
\expandafter\def\csname jl@/human/26/n\endcsname{320.000}
\expandafter\def\csname jl@/human/26/spearman_rho\endcsname{0.566}
\expandafter\def\csname jl@/human/260/n\endcsname{288.000}
\expandafter\def\csname jl@/human/260/spearman_rho\endcsname{0.535}
\expandafter\def\csname jl@/human/261/n\endcsname{288.000}
\expandafter\def\csname jl@/human/261/spearman_rho\endcsname{0.597}
\expandafter\def\csname jl@/human/262/n\endcsname{288.000}
\expandafter\def\csname jl@/human/262/spearman_rho\endcsname{0.648}
\expandafter\def\csname jl@/human/263/n\endcsname{288.000}
\expandafter\def\csname jl@/human/263/spearman_rho\endcsname{0.751}
\expandafter\def\csname jl@/human/264/n\endcsname{288.000}
\expandafter\def\csname jl@/human/264/spearman_rho\endcsname{0.751}
\expandafter\def\csname jl@/human/265/n\endcsname{288.000}
\expandafter\def\csname jl@/human/265/spearman_rho\endcsname{0.664}
\expandafter\def\csname jl@/human/266/n\endcsname{288.000}
\expandafter\def\csname jl@/human/266/spearman_rho\endcsname{0.652}
\expandafter\def\csname jl@/human/267/n\endcsname{288.000}
\expandafter\def\csname jl@/human/267/spearman_rho\endcsname{0.596}
\expandafter\def\csname jl@/human/268/n\endcsname{288.000}
\expandafter\def\csname jl@/human/268/spearman_rho\endcsname{0.756}
\expandafter\def\csname jl@/human/269/n\endcsname{288.000}
\expandafter\def\csname jl@/human/269/spearman_rho\endcsname{0.691}
\expandafter\def\csname jl@/human/27/kendall_tau\endcsname{0.484}
\expandafter\def\csname jl@/human/27/n\endcsname{248.000}
\expandafter\def\csname jl@/human/27/spearman_rho\endcsname{0.576}
\expandafter\def\csname jl@/human/270/n\endcsname{288.000}
\expandafter\def\csname jl@/human/270/spearman_rho\endcsname{0.647}
\expandafter\def\csname jl@/human/271/n\endcsname{288.000}
\expandafter\def\csname jl@/human/271/spearman_rho\endcsname{0.487}
\expandafter\def\csname jl@/human/272/n\endcsname{288.000}
\expandafter\def\csname jl@/human/272/spearman_rho\endcsname{0.478}
\expandafter\def\csname jl@/human/273/n\endcsname{288.000}
\expandafter\def\csname jl@/human/273/spearman_rho\endcsname{0.440}
\expandafter\def\csname jl@/human/274/n\endcsname{288.000}
\expandafter\def\csname jl@/human/274/spearman_rho\endcsname{0.709}
\expandafter\def\csname jl@/human/275/n\endcsname{288.000}
\expandafter\def\csname jl@/human/275/spearman_rho\endcsname{0.705}
\expandafter\def\csname jl@/human/276/n\endcsname{288.000}
\expandafter\def\csname jl@/human/276/spearman_rho\endcsname{0.709}
\expandafter\def\csname jl@/human/277/n\endcsname{288.000}
\expandafter\def\csname jl@/human/277/spearman_rho\endcsname{0.592}
\expandafter\def\csname jl@/human/278/n\endcsname{288.000}
\expandafter\def\csname jl@/human/278/spearman_rho\endcsname{0.592}
\expandafter\def\csname jl@/human/279/n\endcsname{288.000}
\expandafter\def\csname jl@/human/279/spearman_rho\endcsname{0.594}
\expandafter\def\csname jl@/human/28/kendall_tau\endcsname{0.551}
\expandafter\def\csname jl@/human/28/n\endcsname{320.000}
\expandafter\def\csname jl@/human/28/spearman_rho\endcsname{0.647}
\expandafter\def\csname jl@/human/280/n\endcsname{288.000}
\expandafter\def\csname jl@/human/280/spearman_rho\endcsname{0.720}
\expandafter\def\csname jl@/human/281/n\endcsname{288.000}
\expandafter\def\csname jl@/human/281/spearman_rho\endcsname{0.822}
\expandafter\def\csname jl@/human/282/n\endcsname{288.000}
\expandafter\def\csname jl@/human/282/spearman_rho\endcsname{0.807}
\expandafter\def\csname jl@/human/283/n\endcsname{288.000}
\expandafter\def\csname jl@/human/283/spearman_rho\endcsname{0.599}
\expandafter\def\csname jl@/human/284/n\endcsname{288.000}
\expandafter\def\csname jl@/human/284/spearman_rho\endcsname{0.677}
\expandafter\def\csname jl@/human/285/n\endcsname{288.000}
\expandafter\def\csname jl@/human/285/spearman_rho\endcsname{0.623}
\expandafter\def\csname jl@/human/286/n\endcsname{288.000}
\expandafter\def\csname jl@/human/286/spearman_rho\endcsname{0.786}
\expandafter\def\csname jl@/human/287/n\endcsname{288.000}
\expandafter\def\csname jl@/human/287/spearman_rho\endcsname{0.704}
\expandafter\def\csname jl@/human/288/n\endcsname{288.000}
\expandafter\def\csname jl@/human/288/spearman_rho\endcsname{0.689}
\expandafter\def\csname jl@/human/289/n\endcsname{288.000}
\expandafter\def\csname jl@/human/289/spearman_rho\endcsname{0.503}
\expandafter\def\csname jl@/human/29/kendall_tau\endcsname{0.550}
\expandafter\def\csname jl@/human/29/n\endcsname{320.000}
\expandafter\def\csname jl@/human/29/spearman_rho\endcsname{0.646}
\expandafter\def\csname jl@/human/290/n\endcsname{288.000}
\expandafter\def\csname jl@/human/290/spearman_rho\endcsname{0.387}
\expandafter\def\csname jl@/human/291/n\endcsname{288.000}
\expandafter\def\csname jl@/human/291/spearman_rho\endcsname{0.348}
\expandafter\def\csname jl@/human/292/n\endcsname{288.000}
\expandafter\def\csname jl@/human/292/spearman_rho\endcsname{0.762}
\expandafter\def\csname jl@/human/293/n\endcsname{288.000}
\expandafter\def\csname jl@/human/293/spearman_rho\endcsname{0.589}
\expandafter\def\csname jl@/human/294/n\endcsname{288.000}
\expandafter\def\csname jl@/human/294/spearman_rho\endcsname{0.680}
\expandafter\def\csname jl@/human/295/n\endcsname{288.000}
\expandafter\def\csname jl@/human/295/spearman_rho\endcsname{0.647}
\expandafter\def\csname jl@/human/296/n\endcsname{288.000}
\expandafter\def\csname jl@/human/296/spearman_rho\endcsname{0.557}
\expandafter\def\csname jl@/human/297/n\endcsname{288.000}
\expandafter\def\csname jl@/human/297/spearman_rho\endcsname{0.578}
\expandafter\def\csname jl@/human/298/n\endcsname{288.000}
\expandafter\def\csname jl@/human/298/spearman_rho\endcsname{0.658}
\expandafter\def\csname jl@/human/299/n\endcsname{288.000}
\expandafter\def\csname jl@/human/299/spearman_rho\endcsname{0.723}
\expandafter\def\csname jl@/human/3/kendall_tau\endcsname{0.525}
\expandafter\def\csname jl@/human/3/n\endcsname{320.000}
\expandafter\def\csname jl@/human/3/spearman_rho\endcsname{0.630}
\expandafter\def\csname jl@/human/30/kendall_tau\endcsname{0.464}
\expandafter\def\csname jl@/human/30/n\endcsname{320.000}
\expandafter\def\csname jl@/human/30/spearman_rho\endcsname{0.560}
\expandafter\def\csname jl@/human/300/n\endcsname{288.000}
\expandafter\def\csname jl@/human/300/spearman_rho\endcsname{0.760}
\expandafter\def\csname jl@/human/301/n\endcsname{288.000}
\expandafter\def\csname jl@/human/301/spearman_rho\endcsname{0.682}
\expandafter\def\csname jl@/human/302/n\endcsname{288.000}
\expandafter\def\csname jl@/human/302/spearman_rho\endcsname{0.584}
\expandafter\def\csname jl@/human/303/n\endcsname{288.000}
\expandafter\def\csname jl@/human/303/spearman_rho\endcsname{0.636}
\expandafter\def\csname jl@/human/304/n\endcsname{288.000}
\expandafter\def\csname jl@/human/304/spearman_rho\endcsname{0.762}
\expandafter\def\csname jl@/human/305/n\endcsname{288.000}
\expandafter\def\csname jl@/human/305/spearman_rho\endcsname{0.685}
\expandafter\def\csname jl@/human/306/n\endcsname{288.000}
\expandafter\def\csname jl@/human/306/spearman_rho\endcsname{0.673}
\expandafter\def\csname jl@/human/307/n\endcsname{288.000}
\expandafter\def\csname jl@/human/307/spearman_rho\endcsname{0.589}
\expandafter\def\csname jl@/human/308/n\endcsname{288.000}
\expandafter\def\csname jl@/human/308/spearman_rho\endcsname{0.501}
\expandafter\def\csname jl@/human/309/n\endcsname{288.000}
\expandafter\def\csname jl@/human/309/spearman_rho\endcsname{0.500}
\expandafter\def\csname jl@/human/31/kendall_tau\endcsname{0.524}
\expandafter\def\csname jl@/human/31/n\endcsname{320.000}
\expandafter\def\csname jl@/human/31/spearman_rho\endcsname{0.630}
\expandafter\def\csname jl@/human/310/n\endcsname{288.000}
\expandafter\def\csname jl@/human/310/spearman_rho\endcsname{0.740}
\expandafter\def\csname jl@/human/311/n\endcsname{288.000}
\expandafter\def\csname jl@/human/311/spearman_rho\endcsname{0.711}
\expandafter\def\csname jl@/human/312/n\endcsname{288.000}
\expandafter\def\csname jl@/human/312/spearman_rho\endcsname{0.684}
\expandafter\def\csname jl@/human/313/n\endcsname{288.000}
\expandafter\def\csname jl@/human/313/spearman_rho\endcsname{0.599}
\expandafter\def\csname jl@/human/314/n\endcsname{288.000}
\expandafter\def\csname jl@/human/314/spearman_rho\endcsname{0.507}
\expandafter\def\csname jl@/human/315/n\endcsname{288.000}
\expandafter\def\csname jl@/human/315/spearman_rho\endcsname{0.569}
\expandafter\def\csname jl@/human/316/n\endcsname{288.000}
\expandafter\def\csname jl@/human/316/spearman_rho\endcsname{0.691}
\expandafter\def\csname jl@/human/317/n\endcsname{288.000}
\expandafter\def\csname jl@/human/317/spearman_rho\endcsname{0.821}
\expandafter\def\csname jl@/human/318/n\endcsname{288.000}
\expandafter\def\csname jl@/human/318/spearman_rho\endcsname{0.807}
\expandafter\def\csname jl@/human/319/n\endcsname{288.000}
\expandafter\def\csname jl@/human/319/spearman_rho\endcsname{0.656}
\expandafter\def\csname jl@/human/32/kendall_tau\endcsname{0.580}
\expandafter\def\csname jl@/human/32/n\endcsname{320.000}
\expandafter\def\csname jl@/human/32/spearman_rho\endcsname{0.688}
\expandafter\def\csname jl@/human/320/n\endcsname{288.000}
\expandafter\def\csname jl@/human/320/spearman_rho\endcsname{0.622}
\expandafter\def\csname jl@/human/321/n\endcsname{288.000}
\expandafter\def\csname jl@/human/321/spearman_rho\endcsname{0.681}
\expandafter\def\csname jl@/human/322/n\endcsname{288.000}
\expandafter\def\csname jl@/human/322/spearman_rho\endcsname{0.796}
\expandafter\def\csname jl@/human/323/n\endcsname{288.000}
\expandafter\def\csname jl@/human/323/spearman_rho\endcsname{0.681}
\expandafter\def\csname jl@/human/324/n\endcsname{288.000}
\expandafter\def\csname jl@/human/324/spearman_rho\endcsname{0.561}
\expandafter\def\csname jl@/human/325/n\endcsname{288.000}
\expandafter\def\csname jl@/human/325/spearman_rho\endcsname{0.484}
\expandafter\def\csname jl@/human/326/n\endcsname{288.000}
\expandafter\def\csname jl@/human/326/spearman_rho\endcsname{0.363}
\expandafter\def\csname jl@/human/327/n\endcsname{288.000}
\expandafter\def\csname jl@/human/327/spearman_rho\endcsname{0.434}
\expandafter\def\csname jl@/human/328/n\endcsname{288.000}
\expandafter\def\csname jl@/human/328/spearman_rho\endcsname{0.677}
\expandafter\def\csname jl@/human/329/n\endcsname{288.000}
\expandafter\def\csname jl@/human/329/spearman_rho\endcsname{0.573}
\expandafter\def\csname jl@/human/33/kendall_tau\endcsname{0.536}
\expandafter\def\csname jl@/human/33/n\endcsname{320.000}
\expandafter\def\csname jl@/human/33/spearman_rho\endcsname{0.686}
\expandafter\def\csname jl@/human/330/n\endcsname{288.000}
\expandafter\def\csname jl@/human/330/spearman_rho\endcsname{0.625}
\expandafter\def\csname jl@/human/331/n\endcsname{288.000}
\expandafter\def\csname jl@/human/331/spearman_rho\endcsname{0.630}
\expandafter\def\csname jl@/human/332/n\endcsname{288.000}
\expandafter\def\csname jl@/human/332/spearman_rho\endcsname{0.602}
\expandafter\def\csname jl@/human/333/n\endcsname{288.000}
\expandafter\def\csname jl@/human/333/spearman_rho\endcsname{0.701}
\expandafter\def\csname jl@/human/334/n\endcsname{288.000}
\expandafter\def\csname jl@/human/334/spearman_rho\endcsname{0.715}
\expandafter\def\csname jl@/human/335/n\endcsname{288.000}
\expandafter\def\csname jl@/human/335/spearman_rho\endcsname{0.713}
\expandafter\def\csname jl@/human/336/n\endcsname{288.000}
\expandafter\def\csname jl@/human/336/spearman_rho\endcsname{0.722}
\expandafter\def\csname jl@/human/337/n\endcsname{288.000}
\expandafter\def\csname jl@/human/337/spearman_rho\endcsname{0.666}
\expandafter\def\csname jl@/human/338/n\endcsname{288.000}
\expandafter\def\csname jl@/human/338/spearman_rho\endcsname{0.622}
\expandafter\def\csname jl@/human/339/n\endcsname{288.000}
\expandafter\def\csname jl@/human/339/spearman_rho\endcsname{0.656}
\expandafter\def\csname jl@/human/34/kendall_tau\endcsname{0.639}
\expandafter\def\csname jl@/human/34/n\endcsname{320.000}
\expandafter\def\csname jl@/human/34/spearman_rho\endcsname{0.674}
\expandafter\def\csname jl@/human/340/n\endcsname{288.000}
\expandafter\def\csname jl@/human/340/spearman_rho\endcsname{0.748}
\expandafter\def\csname jl@/human/341/n\endcsname{288.000}
\expandafter\def\csname jl@/human/341/spearman_rho\endcsname{0.637}
\expandafter\def\csname jl@/human/342/n\endcsname{288.000}
\expandafter\def\csname jl@/human/342/spearman_rho\endcsname{0.600}
\expandafter\def\csname jl@/human/343/n\endcsname{288.000}
\expandafter\def\csname jl@/human/343/spearman_rho\endcsname{0.583}
\expandafter\def\csname jl@/human/344/n\endcsname{288.000}
\expandafter\def\csname jl@/human/344/spearman_rho\endcsname{0.368}
\expandafter\def\csname jl@/human/345/n\endcsname{288.000}
\expandafter\def\csname jl@/human/345/spearman_rho\endcsname{0.440}
\expandafter\def\csname jl@/human/346/n\endcsname{288.000}
\expandafter\def\csname jl@/human/346/spearman_rho\endcsname{0.726}
\expandafter\def\csname jl@/human/347/n\endcsname{288.000}
\expandafter\def\csname jl@/human/347/spearman_rho\endcsname{0.708}
\expandafter\def\csname jl@/human/348/n\endcsname{288.000}
\expandafter\def\csname jl@/human/348/spearman_rho\endcsname{0.686}
\expandafter\def\csname jl@/human/349/n\endcsname{288.000}
\expandafter\def\csname jl@/human/349/spearman_rho\endcsname{0.626}
\expandafter\def\csname jl@/human/35/kendall_tau\endcsname{0.579}
\expandafter\def\csname jl@/human/35/n\endcsname{320.000}
\expandafter\def\csname jl@/human/35/spearman_rho\endcsname{0.619}
\expandafter\def\csname jl@/human/350/n\endcsname{288.000}
\expandafter\def\csname jl@/human/350/spearman_rho\endcsname{0.565}
\expandafter\def\csname jl@/human/351/n\endcsname{288.000}
\expandafter\def\csname jl@/human/351/spearman_rho\endcsname{0.599}
\expandafter\def\csname jl@/human/352/n\endcsname{288.000}
\expandafter\def\csname jl@/human/352/spearman_rho\endcsname{0.723}
\expandafter\def\csname jl@/human/353/n\endcsname{288.000}
\expandafter\def\csname jl@/human/353/spearman_rho\endcsname{0.818}
\expandafter\def\csname jl@/human/354/n\endcsname{288.000}
\expandafter\def\csname jl@/human/354/spearman_rho\endcsname{0.812}
\expandafter\def\csname jl@/human/355/n\endcsname{288.000}
\expandafter\def\csname jl@/human/355/spearman_rho\endcsname{0.757}
\expandafter\def\csname jl@/human/356/n\endcsname{288.000}
\expandafter\def\csname jl@/human/356/spearman_rho\endcsname{0.680}
\expandafter\def\csname jl@/human/357/n\endcsname{288.000}
\expandafter\def\csname jl@/human/357/spearman_rho\endcsname{0.638}
\expandafter\def\csname jl@/human/358/n\endcsname{288.000}
\expandafter\def\csname jl@/human/358/spearman_rho\endcsname{0.807}
\expandafter\def\csname jl@/human/359/n\endcsname{288.000}
\expandafter\def\csname jl@/human/359/spearman_rho\endcsname{0.742}
\expandafter\def\csname jl@/human/36/kendall_tau\endcsname{0.565}
\expandafter\def\csname jl@/human/36/n\endcsname{320.000}
\expandafter\def\csname jl@/human/36/spearman_rho\endcsname{0.596}
\expandafter\def\csname jl@/human/360/n\endcsname{288.000}
\expandafter\def\csname jl@/human/360/spearman_rho\endcsname{0.688}
\expandafter\def\csname jl@/human/361/n\endcsname{288.000}
\expandafter\def\csname jl@/human/361/spearman_rho\endcsname{0.630}
\expandafter\def\csname jl@/human/362/n\endcsname{288.000}
\expandafter\def\csname jl@/human/362/spearman_rho\endcsname{0.567}
\expandafter\def\csname jl@/human/363/n\endcsname{288.000}
\expandafter\def\csname jl@/human/363/spearman_rho\endcsname{0.566}
\expandafter\def\csname jl@/human/364/n\endcsname{288.000}
\expandafter\def\csname jl@/human/364/spearman_rho\endcsname{0.782}
\expandafter\def\csname jl@/human/365/n\endcsname{288.000}
\expandafter\def\csname jl@/human/365/spearman_rho\endcsname{0.627}
\expandafter\def\csname jl@/human/366/n\endcsname{288.000}
\expandafter\def\csname jl@/human/366/spearman_rho\endcsname{0.623}
\expandafter\def\csname jl@/human/367/n\endcsname{288.000}
\expandafter\def\csname jl@/human/367/spearman_rho\endcsname{0.613}
\expandafter\def\csname jl@/human/368/n\endcsname{288.000}
\expandafter\def\csname jl@/human/368/spearman_rho\endcsname{0.579}
\expandafter\def\csname jl@/human/369/n\endcsname{288.000}
\expandafter\def\csname jl@/human/369/spearman_rho\endcsname{0.606}
\expandafter\def\csname jl@/human/37/kendall_tau\endcsname{0.626}
\expandafter\def\csname jl@/human/37/n\endcsname{320.000}
\expandafter\def\csname jl@/human/37/spearman_rho\endcsname{0.660}
\expandafter\def\csname jl@/human/370/n\endcsname{288.000}
\expandafter\def\csname jl@/human/370/spearman_rho\endcsname{0.627}
\expandafter\def\csname jl@/human/371/n\endcsname{288.000}
\expandafter\def\csname jl@/human/371/spearman_rho\endcsname{0.726}
\expandafter\def\csname jl@/human/372/n\endcsname{288.000}
\expandafter\def\csname jl@/human/372/spearman_rho\endcsname{0.718}
\expandafter\def\csname jl@/human/373/n\endcsname{288.000}
\expandafter\def\csname jl@/human/373/spearman_rho\endcsname{0.712}
\expandafter\def\csname jl@/human/374/n\endcsname{288.000}
\expandafter\def\csname jl@/human/374/spearman_rho\endcsname{0.557}
\expandafter\def\csname jl@/human/375/n\endcsname{288.000}
\expandafter\def\csname jl@/human/375/spearman_rho\endcsname{0.683}
\expandafter\def\csname jl@/human/376/n\endcsname{288.000}
\expandafter\def\csname jl@/human/376/spearman_rho\endcsname{0.768}
\expandafter\def\csname jl@/human/377/n\endcsname{288.000}
\expandafter\def\csname jl@/human/377/spearman_rho\endcsname{0.664}
\expandafter\def\csname jl@/human/378/n\endcsname{288.000}
\expandafter\def\csname jl@/human/378/spearman_rho\endcsname{0.607}
\expandafter\def\csname jl@/human/379/n\endcsname{288.000}
\expandafter\def\csname jl@/human/379/spearman_rho\endcsname{0.637}
\expandafter\def\csname jl@/human/38/kendall_tau\endcsname{0.628}
\expandafter\def\csname jl@/human/38/n\endcsname{320.000}
\expandafter\def\csname jl@/human/38/spearman_rho\endcsname{0.657}
\expandafter\def\csname jl@/human/380/n\endcsname{288.000}
\expandafter\def\csname jl@/human/380/spearman_rho\endcsname{0.420}
\expandafter\def\csname jl@/human/381/n\endcsname{288.000}
\expandafter\def\csname jl@/human/381/spearman_rho\endcsname{0.552}
\expandafter\def\csname jl@/human/382/n\endcsname{288.000}
\expandafter\def\csname jl@/human/382/spearman_rho\endcsname{0.729}
\expandafter\def\csname jl@/human/383/n\endcsname{288.000}
\expandafter\def\csname jl@/human/383/spearman_rho\endcsname{0.699}
\expandafter\def\csname jl@/human/384/n\endcsname{288.000}
\expandafter\def\csname jl@/human/384/spearman_rho\endcsname{0.698}
\expandafter\def\csname jl@/human/385/n\endcsname{288.000}
\expandafter\def\csname jl@/human/385/spearman_rho\endcsname{0.606}
\expandafter\def\csname jl@/human/386/n\endcsname{288.000}
\expandafter\def\csname jl@/human/386/spearman_rho\endcsname{0.601}
\expandafter\def\csname jl@/human/387/n\endcsname{288.000}
\expandafter\def\csname jl@/human/387/spearman_rho\endcsname{0.601}
\expandafter\def\csname jl@/human/388/n\endcsname{288.000}
\expandafter\def\csname jl@/human/388/spearman_rho\endcsname{0.721}
\expandafter\def\csname jl@/human/389/n\endcsname{288.000}
\expandafter\def\csname jl@/human/389/spearman_rho\endcsname{0.812}
\expandafter\def\csname jl@/human/39/kendall_tau\endcsname{0.561}
\expandafter\def\csname jl@/human/39/n\endcsname{320.000}
\expandafter\def\csname jl@/human/39/spearman_rho\endcsname{0.614}
\expandafter\def\csname jl@/human/390/n\endcsname{288.000}
\expandafter\def\csname jl@/human/390/spearman_rho\endcsname{0.813}
\expandafter\def\csname jl@/human/391/n\endcsname{288.000}
\expandafter\def\csname jl@/human/391/spearman_rho\endcsname{0.725}
\expandafter\def\csname jl@/human/392/n\endcsname{288.000}
\expandafter\def\csname jl@/human/392/spearman_rho\endcsname{0.647}
\expandafter\def\csname jl@/human/393/n\endcsname{288.000}
\expandafter\def\csname jl@/human/393/spearman_rho\endcsname{0.650}
\expandafter\def\csname jl@/human/394/n\endcsname{288.000}
\expandafter\def\csname jl@/human/394/spearman_rho\endcsname{0.808}
\expandafter\def\csname jl@/human/395/n\endcsname{288.000}
\expandafter\def\csname jl@/human/395/spearman_rho\endcsname{0.739}
\expandafter\def\csname jl@/human/396/n\endcsname{288.000}
\expandafter\def\csname jl@/human/396/spearman_rho\endcsname{0.651}
\expandafter\def\csname jl@/human/397/n\endcsname{288.000}
\expandafter\def\csname jl@/human/397/spearman_rho\endcsname{0.603}
\expandafter\def\csname jl@/human/398/n\endcsname{288.000}
\expandafter\def\csname jl@/human/398/spearman_rho\endcsname{0.478}
\expandafter\def\csname jl@/human/399/n\endcsname{288.000}
\expandafter\def\csname jl@/human/399/spearman_rho\endcsname{0.489}
\expandafter\def\csname jl@/human/4/kendall_tau\endcsname{0.546}
\expandafter\def\csname jl@/human/4/n\endcsname{320.000}
\expandafter\def\csname jl@/human/4/spearman_rho\endcsname{0.658}
\expandafter\def\csname jl@/human/40/kendall_tau\endcsname{0.488}
\expandafter\def\csname jl@/human/40/n\endcsname{320.000}
\expandafter\def\csname jl@/human/40/spearman_rho\endcsname{0.517}
\expandafter\def\csname jl@/human/400/n\endcsname{288.000}
\expandafter\def\csname jl@/human/400/spearman_rho\endcsname{0.773}
\expandafter\def\csname jl@/human/401/n\endcsname{288.000}
\expandafter\def\csname jl@/human/401/spearman_rho\endcsname{0.676}
\expandafter\def\csname jl@/human/402/n\endcsname{288.000}
\expandafter\def\csname jl@/human/402/spearman_rho\endcsname{0.645}
\expandafter\def\csname jl@/human/403/n\endcsname{288.000}
\expandafter\def\csname jl@/human/403/spearman_rho\endcsname{0.635}
\expandafter\def\csname jl@/human/404/n\endcsname{288.000}
\expandafter\def\csname jl@/human/404/spearman_rho\endcsname{0.619}
\expandafter\def\csname jl@/human/405/n\endcsname{288.000}
\expandafter\def\csname jl@/human/405/spearman_rho\endcsname{0.609}
\expandafter\def\csname jl@/human/406/n\endcsname{288.000}
\expandafter\def\csname jl@/human/406/spearman_rho\endcsname{0.685}
\expandafter\def\csname jl@/human/41/kendall_tau\endcsname{0.458}
\expandafter\def\csname jl@/human/41/n\endcsname{320.000}
\expandafter\def\csname jl@/human/41/spearman_rho\endcsname{0.487}
\expandafter\def\csname jl@/human/42/kendall_tau\endcsname{0.411}
\expandafter\def\csname jl@/human/42/n\endcsname{320.000}
\expandafter\def\csname jl@/human/42/spearman_rho\endcsname{0.450}
\expandafter\def\csname jl@/human/43/kendall_tau\endcsname{0.362}
\expandafter\def\csname jl@/human/43/n\endcsname{320.000}
\expandafter\def\csname jl@/human/43/spearman_rho\endcsname{0.389}
\expandafter\def\csname jl@/human/44/kendall_tau\endcsname{0.317}
\expandafter\def\csname jl@/human/44/n\endcsname{320.000}
\expandafter\def\csname jl@/human/44/spearman_rho\endcsname{0.344}
\expandafter\def\csname jl@/human/45/kendall_tau\endcsname{0.470}
\expandafter\def\csname jl@/human/45/n\endcsname{320.000}
\expandafter\def\csname jl@/human/45/spearman_rho\endcsname{0.553}
\expandafter\def\csname jl@/human/46/kendall_tau\endcsname{0.462}
\expandafter\def\csname jl@/human/46/n\endcsname{320.000}
\expandafter\def\csname jl@/human/46/spearman_rho\endcsname{0.540}
\expandafter\def\csname jl@/human/47/kendall_tau\endcsname{0.527}
\expandafter\def\csname jl@/human/47/n\endcsname{320.000}
\expandafter\def\csname jl@/human/47/spearman_rho\endcsname{0.605}
\expandafter\def\csname jl@/human/48/kendall_tau\endcsname{0.461}
\expandafter\def\csname jl@/human/48/n\endcsname{320.000}
\expandafter\def\csname jl@/human/48/spearman_rho\endcsname{0.548}
\expandafter\def\csname jl@/human/49/kendall_tau\endcsname{0.368}
\expandafter\def\csname jl@/human/49/n\endcsname{320.000}
\expandafter\def\csname jl@/human/49/spearman_rho\endcsname{0.444}
\expandafter\def\csname jl@/human/5/kendall_tau\endcsname{0.545}
\expandafter\def\csname jl@/human/5/n\endcsname{320.000}
\expandafter\def\csname jl@/human/5/spearman_rho\endcsname{0.688}
\expandafter\def\csname jl@/human/50/kendall_tau\endcsname{0.426}
\expandafter\def\csname jl@/human/50/n\endcsname{320.000}
\expandafter\def\csname jl@/human/50/spearman_rho\endcsname{0.506}
\expandafter\def\csname jl@/human/51/kendall_tau\endcsname{0.456}
\expandafter\def\csname jl@/human/51/n\endcsname{320.000}
\expandafter\def\csname jl@/human/51/spearman_rho\endcsname{0.579}
\expandafter\def\csname jl@/human/52/kendall_tau\endcsname{0.573}
\expandafter\def\csname jl@/human/52/n\endcsname{320.000}
\expandafter\def\csname jl@/human/52/spearman_rho\endcsname{0.677}
\expandafter\def\csname jl@/human/53/kendall_tau\endcsname{0.534}
\expandafter\def\csname jl@/human/53/n\endcsname{320.000}
\expandafter\def\csname jl@/human/53/spearman_rho\endcsname{0.629}
\expandafter\def\csname jl@/human/54/kendall_tau\endcsname{0.465}
\expandafter\def\csname jl@/human/54/n\endcsname{320.000}
\expandafter\def\csname jl@/human/54/spearman_rho\endcsname{0.545}
\expandafter\def\csname jl@/human/55/kendall_tau\endcsname{0.506}
\expandafter\def\csname jl@/human/55/n\endcsname{320.000}
\expandafter\def\csname jl@/human/55/spearman_rho\endcsname{0.610}
\expandafter\def\csname jl@/human/56/kendall_tau\endcsname{0.511}
\expandafter\def\csname jl@/human/56/n\endcsname{320.000}
\expandafter\def\csname jl@/human/56/spearman_rho\endcsname{0.620}
\expandafter\def\csname jl@/human/57/kendall_tau\endcsname{0.542}
\expandafter\def\csname jl@/human/57/n\endcsname{320.000}
\expandafter\def\csname jl@/human/57/spearman_rho\endcsname{0.685}
\expandafter\def\csname jl@/human/58/kendall_tau\endcsname{0.658}
\expandafter\def\csname jl@/human/58/n\endcsname{320.000}
\expandafter\def\csname jl@/human/58/spearman_rho\endcsname{0.694}
\expandafter\def\csname jl@/human/59/kendall_tau\endcsname{0.646}
\expandafter\def\csname jl@/human/59/n\endcsname{320.000}
\expandafter\def\csname jl@/human/59/spearman_rho\endcsname{0.686}
\expandafter\def\csname jl@/human/6/kendall_tau\endcsname{0.553}
\expandafter\def\csname jl@/human/6/n\endcsname{266.000}
\expandafter\def\csname jl@/human/6/spearman_rho\endcsname{0.657}
\expandafter\def\csname jl@/human/60/kendall_tau\endcsname{0.525}
\expandafter\def\csname jl@/human/60/n\endcsname{320.000}
\expandafter\def\csname jl@/human/60/spearman_rho\endcsname{0.556}
\expandafter\def\csname jl@/human/61/kendall_tau\endcsname{0.640}
\expandafter\def\csname jl@/human/61/n\endcsname{320.000}
\expandafter\def\csname jl@/human/61/spearman_rho\endcsname{0.677}
\expandafter\def\csname jl@/human/62/kendall_tau\endcsname{0.577}
\expandafter\def\csname jl@/human/62/n\endcsname{320.000}
\expandafter\def\csname jl@/human/62/spearman_rho\endcsname{0.604}
\expandafter\def\csname jl@/human/63/kendall_tau\endcsname{0.569}
\expandafter\def\csname jl@/human/63/n\endcsname{320.000}
\expandafter\def\csname jl@/human/63/spearman_rho\endcsname{0.623}
\expandafter\def\csname jl@/human/64/kendall_tau\endcsname{0.526}
\expandafter\def\csname jl@/human/64/n\endcsname{320.000}
\expandafter\def\csname jl@/human/64/spearman_rho\endcsname{0.581}
\expandafter\def\csname jl@/human/65/kendall_tau\endcsname{0.430}
\expandafter\def\csname jl@/human/65/n\endcsname{320.000}
\expandafter\def\csname jl@/human/65/spearman_rho\endcsname{0.476}
\expandafter\def\csname jl@/human/66/kendall_tau\endcsname{0.367}
\expandafter\def\csname jl@/human/66/n\endcsname{320.000}
\expandafter\def\csname jl@/human/66/spearman_rho\endcsname{0.405}
\expandafter\def\csname jl@/human/67/kendall_tau\endcsname{0.448}
\expandafter\def\csname jl@/human/67/n\endcsname{320.000}
\expandafter\def\csname jl@/human/67/spearman_rho\endcsname{0.485}
\expandafter\def\csname jl@/human/68/kendall_tau\endcsname{0.311}
\expandafter\def\csname jl@/human/68/n\endcsname{320.000}
\expandafter\def\csname jl@/human/68/spearman_rho\endcsname{0.347}
\expandafter\def\csname jl@/human/69/kendall_tau\endcsname{0.448}
\expandafter\def\csname jl@/human/69/n\endcsname{320.000}
\expandafter\def\csname jl@/human/69/spearman_rho\endcsname{0.542}
\expandafter\def\csname jl@/human/7/kendall_tau\endcsname{0.658}
\expandafter\def\csname jl@/human/7/n\endcsname{320.000}
\expandafter\def\csname jl@/human/7/spearman_rho\endcsname{0.696}
\expandafter\def\csname jl@/human/70/kendall_tau\endcsname{0.450}
\expandafter\def\csname jl@/human/70/n\endcsname{320.000}
\expandafter\def\csname jl@/human/70/spearman_rho\endcsname{0.528}
\expandafter\def\csname jl@/human/71/kendall_tau\endcsname{0.499}
\expandafter\def\csname jl@/human/71/n\endcsname{320.000}
\expandafter\def\csname jl@/human/71/spearman_rho\endcsname{0.584}
\expandafter\def\csname jl@/human/72/kendall_tau\endcsname{0.396}
\expandafter\def\csname jl@/human/72/n\endcsname{320.000}
\expandafter\def\csname jl@/human/72/spearman_rho\endcsname{0.467}
\expandafter\def\csname jl@/human/73/kendall_tau\endcsname{0.339}
\expandafter\def\csname jl@/human/73/n\endcsname{320.000}
\expandafter\def\csname jl@/human/73/spearman_rho\endcsname{0.410}
\expandafter\def\csname jl@/human/74/kendall_tau\endcsname{0.422}
\expandafter\def\csname jl@/human/74/n\endcsname{320.000}
\expandafter\def\csname jl@/human/74/spearman_rho\endcsname{0.511}
\expandafter\def\csname jl@/human/75/kendall_tau\endcsname{0.443}
\expandafter\def\csname jl@/human/75/n\endcsname{320.000}
\expandafter\def\csname jl@/human/75/spearman_rho\endcsname{0.569}
\expandafter\def\csname jl@/human/76/kendall_tau\endcsname{0.580}
\expandafter\def\csname jl@/human/76/n\endcsname{320.000}
\expandafter\def\csname jl@/human/76/spearman_rho\endcsname{0.685}
\expandafter\def\csname jl@/human/77/kendall_tau\endcsname{0.555}
\expandafter\def\csname jl@/human/77/n\endcsname{320.000}
\expandafter\def\csname jl@/human/77/spearman_rho\endcsname{0.656}
\expandafter\def\csname jl@/human/78/kendall_tau\endcsname{0.381}
\expandafter\def\csname jl@/human/78/n\endcsname{320.000}
\expandafter\def\csname jl@/human/78/spearman_rho\endcsname{0.450}
\expandafter\def\csname jl@/human/79/kendall_tau\endcsname{0.557}
\expandafter\def\csname jl@/human/79/n\endcsname{320.000}
\expandafter\def\csname jl@/human/79/spearman_rho\endcsname{0.665}
\expandafter\def\csname jl@/human/8/kendall_tau\endcsname{0.627}
\expandafter\def\csname jl@/human/8/n\endcsname{320.000}
\expandafter\def\csname jl@/human/8/spearman_rho\endcsname{0.670}
\expandafter\def\csname jl@/human/80/kendall_tau\endcsname{0.511}
\expandafter\def\csname jl@/human/80/n\endcsname{320.000}
\expandafter\def\csname jl@/human/80/spearman_rho\endcsname{0.620}
\expandafter\def\csname jl@/human/81/kendall_tau\endcsname{0.532}
\expandafter\def\csname jl@/human/81/n\endcsname{320.000}
\expandafter\def\csname jl@/human/81/spearman_rho\endcsname{0.681}
\expandafter\def\csname jl@/human/82/kendall_tau\endcsname{0.647}
\expandafter\def\csname jl@/human/82/n\endcsname{320.000}
\expandafter\def\csname jl@/human/82/spearman_rho\endcsname{0.684}
\expandafter\def\csname jl@/human/83/kendall_tau\endcsname{0.600}
\expandafter\def\csname jl@/human/83/n\endcsname{320.000}
\expandafter\def\csname jl@/human/83/spearman_rho\endcsname{0.640}
\expandafter\def\csname jl@/human/84/kendall_tau\endcsname{0.551}
\expandafter\def\csname jl@/human/84/n\endcsname{320.000}
\expandafter\def\csname jl@/human/84/spearman_rho\endcsname{0.584}
\expandafter\def\csname jl@/human/85/kendall_tau\endcsname{0.640}
\expandafter\def\csname jl@/human/85/n\endcsname{320.000}
\expandafter\def\csname jl@/human/85/spearman_rho\endcsname{0.680}
\expandafter\def\csname jl@/human/86/kendall_tau\endcsname{0.600}
\expandafter\def\csname jl@/human/86/n\endcsname{320.000}
\expandafter\def\csname jl@/human/86/spearman_rho\endcsname{0.627}
\expandafter\def\csname jl@/human/87/kendall_tau\endcsname{0.566}
\expandafter\def\csname jl@/human/87/n\endcsname{320.000}
\expandafter\def\csname jl@/human/87/spearman_rho\endcsname{0.623}
\expandafter\def\csname jl@/human/88/kendall_tau\endcsname{0.472}
\expandafter\def\csname jl@/human/88/n\endcsname{320.000}
\expandafter\def\csname jl@/human/88/spearman_rho\endcsname{0.524}
\expandafter\def\csname jl@/human/89/kendall_tau\endcsname{0.212}
\expandafter\def\csname jl@/human/89/n\endcsname{320.000}
\expandafter\def\csname jl@/human/89/spearman_rho\endcsname{0.232}
\expandafter\def\csname jl@/human/9/kendall_tau\endcsname{0.525}
\expandafter\def\csname jl@/human/9/n\endcsname{320.000}
\expandafter\def\csname jl@/human/9/spearman_rho\endcsname{0.556}
\expandafter\def\csname jl@/human/90/kendall_tau\endcsname{0.258}
\expandafter\def\csname jl@/human/90/n\endcsname{320.000}
\expandafter\def\csname jl@/human/90/spearman_rho\endcsname{0.284}
\expandafter\def\csname jl@/human/91/kendall_tau\endcsname{0.400}
\expandafter\def\csname jl@/human/91/n\endcsname{320.000}
\expandafter\def\csname jl@/human/91/spearman_rho\endcsname{0.444}
\expandafter\def\csname jl@/human/92/kendall_tau\endcsname{0.102}
\expandafter\def\csname jl@/human/92/n\endcsname{320.000}
\expandafter\def\csname jl@/human/92/spearman_rho\endcsname{0.114}
\expandafter\def\csname jl@/human/93/kendall_tau\endcsname{0.388}
\expandafter\def\csname jl@/human/93/n\endcsname{320.000}
\expandafter\def\csname jl@/human/93/spearman_rho\endcsname{0.463}
\expandafter\def\csname jl@/human/94/kendall_tau\endcsname{0.469}
\expandafter\def\csname jl@/human/94/n\endcsname{320.000}
\expandafter\def\csname jl@/human/94/spearman_rho\endcsname{0.549}
\expandafter\def\csname jl@/human/95/kendall_tau\endcsname{0.492}
\expandafter\def\csname jl@/human/95/n\endcsname{320.000}
\expandafter\def\csname jl@/human/95/spearman_rho\endcsname{0.574}
\expandafter\def\csname jl@/human/96/kendall_tau\endcsname{0.397}
\expandafter\def\csname jl@/human/96/n\endcsname{320.000}
\expandafter\def\csname jl@/human/96/spearman_rho\endcsname{0.467}
\expandafter\def\csname jl@/human/97/kendall_tau\endcsname{0.363}
\expandafter\def\csname jl@/human/97/n\endcsname{320.000}
\expandafter\def\csname jl@/human/97/spearman_rho\endcsname{0.448}
\expandafter\def\csname jl@/human/98/kendall_tau\endcsname{0.403}
\expandafter\def\csname jl@/human/98/n\endcsname{320.000}
\expandafter\def\csname jl@/human/98/spearman_rho\endcsname{0.493}
\expandafter\def\csname jl@/human/99/kendall_tau\endcsname{0.435}
\expandafter\def\csname jl@/human/99/n\endcsname{320.000}
\expandafter\def\csname jl@/human/99/spearman_rho\endcsname{0.560}
\expandafter\def\csname jl@/pairwise/0/weighted_kappa\endcsname{0.813}
\expandafter\def\csname jl@/pairwise/1/weighted_kappa\endcsname{0.749}
\expandafter\def\csname jl@/pairwise/10/weighted_kappa\endcsname{0.928}
\expandafter\def\csname jl@/pairwise/100/weighted_kappa\endcsname{0.651}
\expandafter\def\csname jl@/pairwise/101/weighted_kappa\endcsname{0.496}
\expandafter\def\csname jl@/pairwise/102/weighted_kappa\endcsname{0.469}
\expandafter\def\csname jl@/pairwise/103/weighted_kappa\endcsname{0.358}
\expandafter\def\csname jl@/pairwise/104/weighted_kappa\endcsname{0.414}
\expandafter\def\csname jl@/pairwise/105/weighted_kappa\endcsname{0.562}
\expandafter\def\csname jl@/pairwise/106/weighted_kappa\endcsname{0.295}
\expandafter\def\csname jl@/pairwise/107/weighted_kappa\endcsname{0.272}
\expandafter\def\csname jl@/pairwise/108/weighted_kappa\endcsname{0.406}
\expandafter\def\csname jl@/pairwise/109/weighted_kappa\endcsname{0.321}
\expandafter\def\csname jl@/pairwise/11/weighted_kappa\endcsname{0.870}
\expandafter\def\csname jl@/pairwise/110/weighted_kappa\endcsname{0.769}
\expandafter\def\csname jl@/pairwise/111/weighted_kappa\endcsname{0.557}
\expandafter\def\csname jl@/pairwise/112/weighted_kappa\endcsname{0.516}
\expandafter\def\csname jl@/pairwise/113/weighted_kappa\endcsname{0.574}
\expandafter\def\csname jl@/pairwise/114/weighted_kappa\endcsname{0.650}
\expandafter\def\csname jl@/pairwise/115/weighted_kappa\endcsname{0.513}
\expandafter\def\csname jl@/pairwise/116/weighted_kappa\endcsname{0.587}
\expandafter\def\csname jl@/pairwise/117/weighted_kappa\endcsname{0.385}
\expandafter\def\csname jl@/pairwise/118/weighted_kappa\endcsname{0.578}
\expandafter\def\csname jl@/pairwise/119/weighted_kappa\endcsname{0.638}
\expandafter\def\csname jl@/pairwise/12/weighted_kappa\endcsname{0.893}
\expandafter\def\csname jl@/pairwise/120/weighted_kappa\endcsname{0.841}
\expandafter\def\csname jl@/pairwise/121/weighted_kappa\endcsname{0.611}
\expandafter\def\csname jl@/pairwise/122/weighted_kappa\endcsname{0.648}
\expandafter\def\csname jl@/pairwise/123/weighted_kappa\endcsname{0.715}
\expandafter\def\csname jl@/pairwise/124/weighted_kappa\endcsname{0.591}
\expandafter\def\csname jl@/pairwise/125/weighted_kappa\endcsname{0.667}
\expandafter\def\csname jl@/pairwise/126/weighted_kappa\endcsname{0.704}
\expandafter\def\csname jl@/pairwise/127/weighted_kappa\endcsname{0.387}
\expandafter\def\csname jl@/pairwise/128/weighted_kappa\endcsname{0.556}
\expandafter\def\csname jl@/pairwise/129/weighted_kappa\endcsname{0.771}
\expandafter\def\csname jl@/pairwise/13/weighted_kappa\endcsname{0.847}
\expandafter\def\csname jl@/pairwise/130/weighted_kappa\endcsname{0.928}
\expandafter\def\csname jl@/pairwise/131/weighted_kappa\endcsname{0.861}
\expandafter\def\csname jl@/pairwise/132/weighted_kappa\endcsname{0.911}
\expandafter\def\csname jl@/pairwise/133/weighted_kappa\endcsname{0.867}
\expandafter\def\csname jl@/pairwise/134/weighted_kappa\endcsname{0.882}
\expandafter\def\csname jl@/pairwise/135/weighted_kappa\endcsname{0.860}
\expandafter\def\csname jl@/pairwise/136/weighted_kappa\endcsname{0.840}
\expandafter\def\csname jl@/pairwise/137/weighted_kappa\endcsname{0.796}
\expandafter\def\csname jl@/pairwise/138/weighted_kappa\endcsname{0.787}
\expandafter\def\csname jl@/pairwise/139/weighted_kappa\endcsname{0.848}
\expandafter\def\csname jl@/pairwise/14/weighted_kappa\endcsname{0.877}
\expandafter\def\csname jl@/pairwise/140/weighted_kappa\endcsname{0.540}
\expandafter\def\csname jl@/pairwise/141/weighted_kappa\endcsname{0.427}
\expandafter\def\csname jl@/pairwise/142/weighted_kappa\endcsname{0.567}
\expandafter\def\csname jl@/pairwise/143/weighted_kappa\endcsname{0.325}
\expandafter\def\csname jl@/pairwise/144/weighted_kappa\endcsname{0.399}
\expandafter\def\csname jl@/pairwise/145/weighted_kappa\endcsname{0.507}
\expandafter\def\csname jl@/pairwise/146/weighted_kappa\endcsname{0.533}
\expandafter\def\csname jl@/pairwise/147/weighted_kappa\endcsname{0.349}
\expandafter\def\csname jl@/pairwise/148/weighted_kappa\endcsname{0.299}
\expandafter\def\csname jl@/pairwise/149/weighted_kappa\endcsname{0.298}
\expandafter\def\csname jl@/pairwise/15/weighted_kappa\endcsname{0.833}
\expandafter\def\csname jl@/pairwise/150/weighted_kappa\endcsname{0.773}
\expandafter\def\csname jl@/pairwise/151/weighted_kappa\endcsname{0.582}
\expandafter\def\csname jl@/pairwise/152/weighted_kappa\endcsname{0.612}
\expandafter\def\csname jl@/pairwise/153/weighted_kappa\endcsname{0.525}
\expandafter\def\csname jl@/pairwise/154/weighted_kappa\endcsname{0.605}
\expandafter\def\csname jl@/pairwise/155/weighted_kappa\endcsname{0.684}
\expandafter\def\csname jl@/pairwise/156/weighted_kappa\endcsname{0.599}
\expandafter\def\csname jl@/pairwise/157/weighted_kappa\endcsname{0.545}
\expandafter\def\csname jl@/pairwise/158/weighted_kappa\endcsname{0.542}
\expandafter\def\csname jl@/pairwise/159/weighted_kappa\endcsname{0.686}
\expandafter\def\csname jl@/pairwise/16/weighted_kappa\endcsname{0.815}
\expandafter\def\csname jl@/pairwise/160/weighted_kappa\endcsname{0.881}
\expandafter\def\csname jl@/pairwise/161/weighted_kappa\endcsname{0.801}
\expandafter\def\csname jl@/pairwise/162/weighted_kappa\endcsname{0.733}
\expandafter\def\csname jl@/pairwise/163/weighted_kappa\endcsname{0.734}
\expandafter\def\csname jl@/pairwise/164/weighted_kappa\endcsname{0.774}
\expandafter\def\csname jl@/pairwise/165/weighted_kappa\endcsname{0.750}
\expandafter\def\csname jl@/pairwise/166/weighted_kappa\endcsname{0.763}
\expandafter\def\csname jl@/pairwise/167/weighted_kappa\endcsname{0.604}
\expandafter\def\csname jl@/pairwise/168/weighted_kappa\endcsname{0.674}
\expandafter\def\csname jl@/pairwise/169/weighted_kappa\endcsname{0.775}
\expandafter\def\csname jl@/pairwise/17/weighted_kappa\endcsname{0.812}
\expandafter\def\csname jl@/pairwise/170/weighted_kappa\endcsname{0.947}
\expandafter\def\csname jl@/pairwise/171/weighted_kappa\endcsname{0.879}
\expandafter\def\csname jl@/pairwise/172/weighted_kappa\endcsname{0.917}
\expandafter\def\csname jl@/pairwise/173/weighted_kappa\endcsname{0.853}
\expandafter\def\csname jl@/pairwise/174/weighted_kappa\endcsname{0.898}
\expandafter\def\csname jl@/pairwise/175/weighted_kappa\endcsname{0.882}
\expandafter\def\csname jl@/pairwise/176/weighted_kappa\endcsname{0.851}
\expandafter\def\csname jl@/pairwise/177/weighted_kappa\endcsname{0.814}
\expandafter\def\csname jl@/pairwise/178/weighted_kappa\endcsname{0.784}
\expandafter\def\csname jl@/pairwise/179/weighted_kappa\endcsname{0.824}
\expandafter\def\csname jl@/pairwise/18/weighted_kappa\endcsname{0.794}
\expandafter\def\csname jl@/pairwise/180/weighted_kappa\endcsname{0.698}
\expandafter\def\csname jl@/pairwise/181/weighted_kappa\endcsname{0.498}
\expandafter\def\csname jl@/pairwise/182/weighted_kappa\endcsname{0.595}
\expandafter\def\csname jl@/pairwise/183/weighted_kappa\endcsname{0.468}
\expandafter\def\csname jl@/pairwise/184/weighted_kappa\endcsname{0.503}
\expandafter\def\csname jl@/pairwise/185/weighted_kappa\endcsname{0.501}
\expandafter\def\csname jl@/pairwise/186/weighted_kappa\endcsname{0.488}
\expandafter\def\csname jl@/pairwise/187/weighted_kappa\endcsname{0.455}
\expandafter\def\csname jl@/pairwise/188/weighted_kappa\endcsname{0.413}
\expandafter\def\csname jl@/pairwise/189/weighted_kappa\endcsname{0.531}
\expandafter\def\csname jl@/pairwise/19/weighted_kappa\endcsname{0.833}
\expandafter\def\csname jl@/pairwise/190/weighted_kappa\endcsname{0.853}
\expandafter\def\csname jl@/pairwise/191/weighted_kappa\endcsname{0.687}
\expandafter\def\csname jl@/pairwise/192/weighted_kappa\endcsname{0.741}
\expandafter\def\csname jl@/pairwise/193/weighted_kappa\endcsname{0.664}
\expandafter\def\csname jl@/pairwise/194/weighted_kappa\endcsname{0.665}
\expandafter\def\csname jl@/pairwise/195/weighted_kappa\endcsname{0.719}
\expandafter\def\csname jl@/pairwise/196/weighted_kappa\endcsname{0.656}
\expandafter\def\csname jl@/pairwise/197/weighted_kappa\endcsname{0.625}
\expandafter\def\csname jl@/pairwise/198/weighted_kappa\endcsname{0.708}
\expandafter\def\csname jl@/pairwise/199/weighted_kappa\endcsname{0.642}
\expandafter\def\csname jl@/pairwise/2/weighted_kappa\endcsname{0.605}
\expandafter\def\csname jl@/pairwise/20/weighted_kappa\endcsname{0.566}
\expandafter\def\csname jl@/pairwise/200/weighted_kappa\endcsname{0.905}
\expandafter\def\csname jl@/pairwise/201/weighted_kappa\endcsname{0.793}
\expandafter\def\csname jl@/pairwise/202/weighted_kappa\endcsname{0.715}
\expandafter\def\csname jl@/pairwise/203/weighted_kappa\endcsname{0.669}
\expandafter\def\csname jl@/pairwise/204/weighted_kappa\endcsname{0.810}
\expandafter\def\csname jl@/pairwise/205/weighted_kappa\endcsname{0.722}
\expandafter\def\csname jl@/pairwise/206/weighted_kappa\endcsname{0.675}
\expandafter\def\csname jl@/pairwise/207/weighted_kappa\endcsname{0.652}
\expandafter\def\csname jl@/pairwise/208/weighted_kappa\endcsname{0.674}
\expandafter\def\csname jl@/pairwise/209/weighted_kappa\endcsname{0.776}
\expandafter\def\csname jl@/pairwise/21/weighted_kappa\endcsname{0.541}
\expandafter\def\csname jl@/pairwise/210/weighted_kappa\endcsname{0.942}
\expandafter\def\csname jl@/pairwise/211/weighted_kappa\endcsname{0.859}
\expandafter\def\csname jl@/pairwise/212/weighted_kappa\endcsname{0.883}
\expandafter\def\csname jl@/pairwise/213/weighted_kappa\endcsname{0.838}
\expandafter\def\csname jl@/pairwise/214/weighted_kappa\endcsname{0.876}
\expandafter\def\csname jl@/pairwise/215/weighted_kappa\endcsname{0.881}
\expandafter\def\csname jl@/pairwise/216/weighted_kappa\endcsname{0.838}
\expandafter\def\csname jl@/pairwise/217/weighted_kappa\endcsname{0.766}
\expandafter\def\csname jl@/pairwise/218/weighted_kappa\endcsname{0.757}
\expandafter\def\csname jl@/pairwise/219/weighted_kappa\endcsname{0.784}
\expandafter\def\csname jl@/pairwise/22/weighted_kappa\endcsname{0.330}
\expandafter\def\csname jl@/pairwise/220/weighted_kappa\endcsname{0.673}
\expandafter\def\csname jl@/pairwise/221/weighted_kappa\endcsname{0.206}
\expandafter\def\csname jl@/pairwise/222/weighted_kappa\endcsname{0.529}
\expandafter\def\csname jl@/pairwise/223/weighted_kappa\endcsname{0.415}
\expandafter\def\csname jl@/pairwise/224/weighted_kappa\endcsname{0.117}
\expandafter\def\csname jl@/pairwise/225/weighted_kappa\endcsname{0.723}
\expandafter\def\csname jl@/pairwise/226/weighted_kappa\endcsname{0.522}
\expandafter\def\csname jl@/pairwise/227/weighted_kappa\endcsname{0.099}
\expandafter\def\csname jl@/pairwise/228/weighted_kappa\endcsname{0.076}
\expandafter\def\csname jl@/pairwise/229/weighted_kappa\endcsname{0.564}
\expandafter\def\csname jl@/pairwise/23/weighted_kappa\endcsname{0.508}
\expandafter\def\csname jl@/pairwise/230/weighted_kappa\endcsname{0.834}
\expandafter\def\csname jl@/pairwise/231/weighted_kappa\endcsname{0.653}
\expandafter\def\csname jl@/pairwise/232/weighted_kappa\endcsname{0.694}
\expandafter\def\csname jl@/pairwise/233/weighted_kappa\endcsname{0.569}
\expandafter\def\csname jl@/pairwise/234/weighted_kappa\endcsname{0.665}
\expandafter\def\csname jl@/pairwise/235/weighted_kappa\endcsname{0.669}
\expandafter\def\csname jl@/pairwise/236/weighted_kappa\endcsname{0.607}
\expandafter\def\csname jl@/pairwise/237/weighted_kappa\endcsname{0.546}
\expandafter\def\csname jl@/pairwise/238/weighted_kappa\endcsname{0.535}
\expandafter\def\csname jl@/pairwise/239/weighted_kappa\endcsname{0.589}
\expandafter\def\csname jl@/pairwise/24/weighted_kappa\endcsname{0.430}
\expandafter\def\csname jl@/pairwise/240/weighted_kappa\endcsname{0.676}
\expandafter\def\csname jl@/pairwise/241/weighted_kappa\endcsname{0.275}
\expandafter\def\csname jl@/pairwise/242/weighted_kappa\endcsname{0.610}
\expandafter\def\csname jl@/pairwise/243/weighted_kappa\endcsname{0.226}
\expandafter\def\csname jl@/pairwise/244/weighted_kappa\endcsname{0.279}
\expandafter\def\csname jl@/pairwise/245/weighted_kappa\endcsname{0.587}
\expandafter\def\csname jl@/pairwise/246/weighted_kappa\endcsname{0.218}
\expandafter\def\csname jl@/pairwise/247/weighted_kappa\endcsname{0.262}
\expandafter\def\csname jl@/pairwise/248/weighted_kappa\endcsname{0.642}
\expandafter\def\csname jl@/pairwise/249/weighted_kappa\endcsname{0.230}
\expandafter\def\csname jl@/pairwise/25/weighted_kappa\endcsname{0.467}
\expandafter\def\csname jl@/pairwise/250/weighted_kappa\endcsname{0.677}
\expandafter\def\csname jl@/pairwise/251/weighted_kappa\endcsname{0.258}
\expandafter\def\csname jl@/pairwise/252/weighted_kappa\endcsname{0.601}
\expandafter\def\csname jl@/pairwise/253/weighted_kappa\endcsname{0.316}
\expandafter\def\csname jl@/pairwise/254/weighted_kappa\endcsname{0.241}
\expandafter\def\csname jl@/pairwise/255/weighted_kappa\endcsname{0.574}
\expandafter\def\csname jl@/pairwise/256/weighted_kappa\endcsname{0.240}
\expandafter\def\csname jl@/pairwise/257/weighted_kappa\endcsname{0.223}
\expandafter\def\csname jl@/pairwise/258/weighted_kappa\endcsname{0.546}
\expandafter\def\csname jl@/pairwise/259/weighted_kappa\endcsname{0.253}
\expandafter\def\csname jl@/pairwise/26/weighted_kappa\endcsname{0.327}
\expandafter\def\csname jl@/pairwise/260/weighted_kappa\endcsname{0.707}
\expandafter\def\csname jl@/pairwise/261/weighted_kappa\endcsname{0.306}
\expandafter\def\csname jl@/pairwise/262/weighted_kappa\endcsname{0.647}
\expandafter\def\csname jl@/pairwise/263/weighted_kappa\endcsname{0.250}
\expandafter\def\csname jl@/pairwise/264/weighted_kappa\endcsname{0.260}
\expandafter\def\csname jl@/pairwise/265/weighted_kappa\endcsname{0.566}
\expandafter\def\csname jl@/pairwise/266/weighted_kappa\endcsname{0.201}
\expandafter\def\csname jl@/pairwise/267/weighted_kappa\endcsname{0.294}
\expandafter\def\csname jl@/pairwise/268/weighted_kappa\endcsname{0.664}
\expandafter\def\csname jl@/pairwise/269/weighted_kappa\endcsname{0.253}
\expandafter\def\csname jl@/pairwise/27/weighted_kappa\endcsname{0.243}
\expandafter\def\csname jl@/pairwise/270/weighted_kappa\endcsname{0.729}
\expandafter\def\csname jl@/pairwise/271/weighted_kappa\endcsname{0.260}
\expandafter\def\csname jl@/pairwise/272/weighted_kappa\endcsname{0.622}
\expandafter\def\csname jl@/pairwise/273/weighted_kappa\endcsname{0.198}
\expandafter\def\csname jl@/pairwise/274/weighted_kappa\endcsname{0.242}
\expandafter\def\csname jl@/pairwise/275/weighted_kappa\endcsname{0.557}
\expandafter\def\csname jl@/pairwise/276/weighted_kappa\endcsname{0.184}
\expandafter\def\csname jl@/pairwise/277/weighted_kappa\endcsname{0.242}
\expandafter\def\csname jl@/pairwise/278/weighted_kappa\endcsname{0.641}
\expandafter\def\csname jl@/pairwise/279/weighted_kappa\endcsname{0.225}
\expandafter\def\csname jl@/pairwise/28/weighted_kappa\endcsname{0.395}
\expandafter\def\csname jl@/pairwise/280/weighted_kappa\endcsname{0.712}
\expandafter\def\csname jl@/pairwise/281/weighted_kappa\endcsname{0.278}
\expandafter\def\csname jl@/pairwise/282/weighted_kappa\endcsname{0.641}
\expandafter\def\csname jl@/pairwise/283/weighted_kappa\endcsname{0.215}
\expandafter\def\csname jl@/pairwise/284/weighted_kappa\endcsname{0.265}
\expandafter\def\csname jl@/pairwise/285/weighted_kappa\endcsname{0.578}
\expandafter\def\csname jl@/pairwise/286/weighted_kappa\endcsname{0.192}
\expandafter\def\csname jl@/pairwise/287/weighted_kappa\endcsname{0.259}
\expandafter\def\csname jl@/pairwise/288/weighted_kappa\endcsname{0.552}
\expandafter\def\csname jl@/pairwise/289/weighted_kappa\endcsname{0.206}
\expandafter\def\csname jl@/pairwise/29/weighted_kappa\endcsname{0.197}
\expandafter\def\csname jl@/pairwise/290/weighted_kappa\endcsname{0.714}
\expandafter\def\csname jl@/pairwise/291/weighted_kappa\endcsname{0.254}
\expandafter\def\csname jl@/pairwise/292/weighted_kappa\endcsname{0.633}
\expandafter\def\csname jl@/pairwise/293/weighted_kappa\endcsname{0.212}
\expandafter\def\csname jl@/pairwise/294/weighted_kappa\endcsname{0.247}
\expandafter\def\csname jl@/pairwise/295/weighted_kappa\endcsname{0.561}
\expandafter\def\csname jl@/pairwise/296/weighted_kappa\endcsname{0.201}
\expandafter\def\csname jl@/pairwise/297/weighted_kappa\endcsname{0.228}
\expandafter\def\csname jl@/pairwise/298/weighted_kappa\endcsname{0.641}
\expandafter\def\csname jl@/pairwise/299/weighted_kappa\endcsname{0.219}
\expandafter\def\csname jl@/pairwise/3/weighted_kappa\endcsname{0.675}
\expandafter\def\csname jl@/pairwise/30/weighted_kappa\endcsname{0.773}
\expandafter\def\csname jl@/pairwise/300/weighted_kappa\endcsname{0.827}
\expandafter\def\csname jl@/pairwise/301/weighted_kappa\endcsname{0.701}
\expandafter\def\csname jl@/pairwise/302/weighted_kappa\endcsname{0.768}
\expandafter\def\csname jl@/pairwise/303/weighted_kappa\endcsname{0.720}
\expandafter\def\csname jl@/pairwise/304/weighted_kappa\endcsname{0.824}
\expandafter\def\csname jl@/pairwise/305/weighted_kappa\endcsname{0.698}
\expandafter\def\csname jl@/pairwise/306/weighted_kappa\endcsname{0.787}
\expandafter\def\csname jl@/pairwise/307/weighted_kappa\endcsname{0.626}
\expandafter\def\csname jl@/pairwise/308/weighted_kappa\endcsname{0.773}
\expandafter\def\csname jl@/pairwise/309/weighted_kappa\endcsname{0.659}
\expandafter\def\csname jl@/pairwise/31/weighted_kappa\endcsname{0.548}
\expandafter\def\csname jl@/pairwise/310/weighted_kappa\endcsname{0.738}
\expandafter\def\csname jl@/pairwise/311/weighted_kappa\endcsname{0.495}
\expandafter\def\csname jl@/pairwise/312/weighted_kappa\endcsname{0.559}
\expandafter\def\csname jl@/pairwise/313/weighted_kappa\endcsname{0.513}
\expandafter\def\csname jl@/pairwise/314/weighted_kappa\endcsname{0.565}
\expandafter\def\csname jl@/pairwise/315/weighted_kappa\endcsname{0.587}
\expandafter\def\csname jl@/pairwise/316/weighted_kappa\endcsname{0.580}
\expandafter\def\csname jl@/pairwise/317/weighted_kappa\endcsname{0.369}
\expandafter\def\csname jl@/pairwise/318/weighted_kappa\endcsname{0.616}
\expandafter\def\csname jl@/pairwise/319/weighted_kappa\endcsname{0.440}
\expandafter\def\csname jl@/pairwise/32/weighted_kappa\endcsname{0.548}
\expandafter\def\csname jl@/pairwise/320/weighted_kappa\endcsname{0.760}
\expandafter\def\csname jl@/pairwise/321/weighted_kappa\endcsname{0.653}
\expandafter\def\csname jl@/pairwise/322/weighted_kappa\endcsname{0.656}
\expandafter\def\csname jl@/pairwise/323/weighted_kappa\endcsname{0.665}
\expandafter\def\csname jl@/pairwise/324/weighted_kappa\endcsname{0.710}
\expandafter\def\csname jl@/pairwise/325/weighted_kappa\endcsname{0.697}
\expandafter\def\csname jl@/pairwise/326/weighted_kappa\endcsname{0.716}
\expandafter\def\csname jl@/pairwise/327/weighted_kappa\endcsname{0.651}
\expandafter\def\csname jl@/pairwise/328/weighted_kappa\endcsname{0.754}
\expandafter\def\csname jl@/pairwise/329/weighted_kappa\endcsname{0.678}
\expandafter\def\csname jl@/pairwise/33/weighted_kappa\endcsname{0.523}
\expandafter\def\csname jl@/pairwise/330/weighted_kappa\endcsname{0.863}
\expandafter\def\csname jl@/pairwise/331/weighted_kappa\endcsname{0.700}
\expandafter\def\csname jl@/pairwise/332/weighted_kappa\endcsname{0.709}
\expandafter\def\csname jl@/pairwise/333/weighted_kappa\endcsname{0.694}
\expandafter\def\csname jl@/pairwise/334/weighted_kappa\endcsname{0.668}
\expandafter\def\csname jl@/pairwise/335/weighted_kappa\endcsname{0.638}
\expandafter\def\csname jl@/pairwise/336/weighted_kappa\endcsname{0.621}
\expandafter\def\csname jl@/pairwise/337/weighted_kappa\endcsname{0.701}
\expandafter\def\csname jl@/pairwise/338/weighted_kappa\endcsname{0.688}
\expandafter\def\csname jl@/pairwise/339/weighted_kappa\endcsname{0.738}
\expandafter\def\csname jl@/pairwise/34/weighted_kappa\endcsname{0.658}
\expandafter\def\csname jl@/pairwise/340/weighted_kappa\endcsname{0.620}
\expandafter\def\csname jl@/pairwise/341/weighted_kappa\endcsname{0.602}
\expandafter\def\csname jl@/pairwise/342/weighted_kappa\endcsname{0.310}
\expandafter\def\csname jl@/pairwise/343/weighted_kappa\endcsname{0.356}
\expandafter\def\csname jl@/pairwise/344/weighted_kappa\endcsname{0.586}
\expandafter\def\csname jl@/pairwise/345/weighted_kappa\endcsname{0.463}
\expandafter\def\csname jl@/pairwise/346/weighted_kappa\endcsname{0.502}
\expandafter\def\csname jl@/pairwise/347/weighted_kappa\endcsname{0.349}
\expandafter\def\csname jl@/pairwise/348/weighted_kappa\endcsname{0.510}
\expandafter\def\csname jl@/pairwise/349/weighted_kappa\endcsname{0.538}
\expandafter\def\csname jl@/pairwise/35/weighted_kappa\endcsname{0.567}
\expandafter\def\csname jl@/pairwise/350/weighted_kappa\endcsname{0.780}
\expandafter\def\csname jl@/pairwise/351/weighted_kappa\endcsname{0.793}
\expandafter\def\csname jl@/pairwise/352/weighted_kappa\endcsname{0.852}
\expandafter\def\csname jl@/pairwise/353/weighted_kappa\endcsname{0.895}
\expandafter\def\csname jl@/pairwise/354/weighted_kappa\endcsname{0.790}
\expandafter\def\csname jl@/pairwise/355/weighted_kappa\endcsname{0.734}
\expandafter\def\csname jl@/pairwise/356/weighted_kappa\endcsname{0.773}
\expandafter\def\csname jl@/pairwise/357/weighted_kappa\endcsname{0.745}
\expandafter\def\csname jl@/pairwise/358/weighted_kappa\endcsname{0.772}
\expandafter\def\csname jl@/pairwise/359/weighted_kappa\endcsname{0.832}
\expandafter\def\csname jl@/pairwise/36/weighted_kappa\endcsname{0.610}
\expandafter\def\csname jl@/pairwise/360/weighted_kappa\endcsname{0.851}
\expandafter\def\csname jl@/pairwise/361/weighted_kappa\endcsname{0.657}
\expandafter\def\csname jl@/pairwise/362/weighted_kappa\endcsname{0.775}
\expandafter\def\csname jl@/pairwise/363/weighted_kappa\endcsname{0.814}
\expandafter\def\csname jl@/pairwise/364/weighted_kappa\endcsname{0.758}
\expandafter\def\csname jl@/pairwise/365/weighted_kappa\endcsname{0.726}
\expandafter\def\csname jl@/pairwise/366/weighted_kappa\endcsname{0.809}
\expandafter\def\csname jl@/pairwise/367/weighted_kappa\endcsname{0.556}
\expandafter\def\csname jl@/pairwise/368/weighted_kappa\endcsname{0.650}
\expandafter\def\csname jl@/pairwise/369/weighted_kappa\endcsname{0.731}
\expandafter\def\csname jl@/pairwise/37/weighted_kappa\endcsname{0.443}
\expandafter\def\csname jl@/pairwise/370/weighted_kappa\endcsname{0.761}
\expandafter\def\csname jl@/pairwise/371/weighted_kappa\endcsname{0.544}
\expandafter\def\csname jl@/pairwise/372/weighted_kappa\endcsname{0.655}
\expandafter\def\csname jl@/pairwise/373/weighted_kappa\endcsname{0.495}
\expandafter\def\csname jl@/pairwise/374/weighted_kappa\endcsname{0.634}
\expandafter\def\csname jl@/pairwise/375/weighted_kappa\endcsname{0.617}
\expandafter\def\csname jl@/pairwise/376/weighted_kappa\endcsname{0.494}
\expandafter\def\csname jl@/pairwise/377/weighted_kappa\endcsname{0.446}
\expandafter\def\csname jl@/pairwise/378/weighted_kappa\endcsname{0.483}
\expandafter\def\csname jl@/pairwise/379/weighted_kappa\endcsname{0.561}
\expandafter\def\csname jl@/pairwise/38/weighted_kappa\endcsname{0.566}
\expandafter\def\csname jl@/pairwise/380/weighted_kappa\endcsname{0.863}
\expandafter\def\csname jl@/pairwise/381/weighted_kappa\endcsname{0.769}
\expandafter\def\csname jl@/pairwise/382/weighted_kappa\endcsname{0.779}
\expandafter\def\csname jl@/pairwise/383/weighted_kappa\endcsname{0.690}
\expandafter\def\csname jl@/pairwise/384/weighted_kappa\endcsname{0.780}
\expandafter\def\csname jl@/pairwise/385/weighted_kappa\endcsname{0.768}
\expandafter\def\csname jl@/pairwise/386/weighted_kappa\endcsname{0.716}
\expandafter\def\csname jl@/pairwise/387/weighted_kappa\endcsname{0.750}
\expandafter\def\csname jl@/pairwise/388/weighted_kappa\endcsname{0.740}
\expandafter\def\csname jl@/pairwise/389/weighted_kappa\endcsname{0.759}
\expandafter\def\csname jl@/pairwise/39/weighted_kappa\endcsname{0.663}
\expandafter\def\csname jl@/pairwise/390/weighted_kappa\endcsname{0.851}
\expandafter\def\csname jl@/pairwise/391/weighted_kappa\endcsname{0.760}
\expandafter\def\csname jl@/pairwise/392/weighted_kappa\endcsname{0.767}
\expandafter\def\csname jl@/pairwise/393/weighted_kappa\endcsname{0.661}
\expandafter\def\csname jl@/pairwise/394/weighted_kappa\endcsname{0.694}
\expandafter\def\csname jl@/pairwise/395/weighted_kappa\endcsname{0.672}
\expandafter\def\csname jl@/pairwise/396/weighted_kappa\endcsname{0.547}
\expandafter\def\csname jl@/pairwise/397/weighted_kappa\endcsname{0.689}
\expandafter\def\csname jl@/pairwise/398/weighted_kappa\endcsname{0.719}
\expandafter\def\csname jl@/pairwise/399/weighted_kappa\endcsname{0.727}
\expandafter\def\csname jl@/pairwise/4/weighted_kappa\endcsname{0.728}
\expandafter\def\csname jl@/pairwise/40/weighted_kappa\endcsname{0.822}
\expandafter\def\csname jl@/pairwise/400/weighted_kappa\endcsname{0.682}
\expandafter\def\csname jl@/pairwise/401/weighted_kappa\endcsname{0.541}
\expandafter\def\csname jl@/pairwise/402/weighted_kappa\endcsname{0.352}
\expandafter\def\csname jl@/pairwise/403/weighted_kappa\endcsname{0.254}
\expandafter\def\csname jl@/pairwise/404/weighted_kappa\endcsname{0.608}
\expandafter\def\csname jl@/pairwise/405/weighted_kappa\endcsname{0.584}
\expandafter\def\csname jl@/pairwise/406/weighted_kappa\endcsname{0.439}
\expandafter\def\csname jl@/pairwise/407/weighted_kappa\endcsname{0.441}
\expandafter\def\csname jl@/pairwise/408/weighted_kappa\endcsname{0.284}
\expandafter\def\csname jl@/pairwise/409/weighted_kappa\endcsname{0.409}
\expandafter\def\csname jl@/pairwise/41/weighted_kappa\endcsname{0.746}
\expandafter\def\csname jl@/pairwise/410/weighted_kappa\endcsname{0.862}
\expandafter\def\csname jl@/pairwise/411/weighted_kappa\endcsname{0.807}
\expandafter\def\csname jl@/pairwise/412/weighted_kappa\endcsname{0.867}
\expandafter\def\csname jl@/pairwise/413/weighted_kappa\endcsname{0.930}
\expandafter\def\csname jl@/pairwise/414/weighted_kappa\endcsname{0.832}
\expandafter\def\csname jl@/pairwise/415/weighted_kappa\endcsname{0.800}
\expandafter\def\csname jl@/pairwise/416/weighted_kappa\endcsname{0.848}
\expandafter\def\csname jl@/pairwise/417/weighted_kappa\endcsname{0.747}
\expandafter\def\csname jl@/pairwise/418/weighted_kappa\endcsname{0.834}
\expandafter\def\csname jl@/pairwise/419/weighted_kappa\endcsname{0.895}
\expandafter\def\csname jl@/pairwise/42/weighted_kappa\endcsname{0.576}
\expandafter\def\csname jl@/pairwise/420/weighted_kappa\endcsname{0.838}
\expandafter\def\csname jl@/pairwise/421/weighted_kappa\endcsname{0.757}
\expandafter\def\csname jl@/pairwise/422/weighted_kappa\endcsname{0.815}
\expandafter\def\csname jl@/pairwise/423/weighted_kappa\endcsname{0.726}
\expandafter\def\csname jl@/pairwise/424/weighted_kappa\endcsname{0.832}
\expandafter\def\csname jl@/pairwise/425/weighted_kappa\endcsname{0.759}
\expandafter\def\csname jl@/pairwise/426/weighted_kappa\endcsname{0.750}
\expandafter\def\csname jl@/pairwise/427/weighted_kappa\endcsname{0.715}
\expandafter\def\csname jl@/pairwise/428/weighted_kappa\endcsname{0.742}
\expandafter\def\csname jl@/pairwise/429/weighted_kappa\endcsname{0.688}
\expandafter\def\csname jl@/pairwise/43/weighted_kappa\endcsname{0.616}
\expandafter\def\csname jl@/pairwise/430/weighted_kappa\endcsname{0.791}
\expandafter\def\csname jl@/pairwise/431/weighted_kappa\endcsname{0.450}
\expandafter\def\csname jl@/pairwise/432/weighted_kappa\endcsname{0.556}
\expandafter\def\csname jl@/pairwise/433/weighted_kappa\endcsname{0.498}
\expandafter\def\csname jl@/pairwise/434/weighted_kappa\endcsname{0.470}
\expandafter\def\csname jl@/pairwise/435/weighted_kappa\endcsname{0.617}
\expandafter\def\csname jl@/pairwise/436/weighted_kappa\endcsname{0.507}
\expandafter\def\csname jl@/pairwise/437/weighted_kappa\endcsname{0.418}
\expandafter\def\csname jl@/pairwise/438/weighted_kappa\endcsname{0.546}
\expandafter\def\csname jl@/pairwise/439/weighted_kappa\endcsname{0.457}
\expandafter\def\csname jl@/pairwise/44/weighted_kappa\endcsname{0.704}
\expandafter\def\csname jl@/pairwise/440/weighted_kappa\endcsname{0.807}
\expandafter\def\csname jl@/pairwise/441/weighted_kappa\endcsname{0.738}
\expandafter\def\csname jl@/pairwise/442/weighted_kappa\endcsname{0.695}
\expandafter\def\csname jl@/pairwise/443/weighted_kappa\endcsname{0.728}
\expandafter\def\csname jl@/pairwise/444/weighted_kappa\endcsname{0.695}
\expandafter\def\csname jl@/pairwise/445/weighted_kappa\endcsname{0.699}
\expandafter\def\csname jl@/pairwise/446/weighted_kappa\endcsname{0.714}
\expandafter\def\csname jl@/pairwise/447/weighted_kappa\endcsname{0.660}
\expandafter\def\csname jl@/pairwise/448/weighted_kappa\endcsname{0.768}
\expandafter\def\csname jl@/pairwise/449/weighted_kappa\endcsname{0.699}
\expandafter\def\csname jl@/pairwise/45/weighted_kappa\endcsname{0.562}
\expandafter\def\csname jl@/pairwise/450/weighted_kappa\endcsname{0.848}
\expandafter\def\csname jl@/pairwise/451/weighted_kappa\endcsname{0.631}
\expandafter\def\csname jl@/pairwise/452/weighted_kappa\endcsname{0.728}
\expandafter\def\csname jl@/pairwise/453/weighted_kappa\endcsname{0.600}
\expandafter\def\csname jl@/pairwise/454/weighted_kappa\endcsname{0.552}
\expandafter\def\csname jl@/pairwise/455/weighted_kappa\endcsname{0.626}
\expandafter\def\csname jl@/pairwise/456/weighted_kappa\endcsname{0.509}
\expandafter\def\csname jl@/pairwise/457/weighted_kappa\endcsname{0.626}
\expandafter\def\csname jl@/pairwise/458/weighted_kappa\endcsname{0.681}
\expandafter\def\csname jl@/pairwise/459/weighted_kappa\endcsname{0.680}
\expandafter\def\csname jl@/pairwise/46/weighted_kappa\endcsname{0.600}
\expandafter\def\csname jl@/pairwise/460/weighted_kappa\endcsname{0.760}
\expandafter\def\csname jl@/pairwise/461/weighted_kappa\endcsname{0.512}
\expandafter\def\csname jl@/pairwise/462/weighted_kappa\endcsname{0.380}
\expandafter\def\csname jl@/pairwise/463/weighted_kappa\endcsname{0.343}
\expandafter\def\csname jl@/pairwise/464/weighted_kappa\endcsname{0.560}
\expandafter\def\csname jl@/pairwise/465/weighted_kappa\endcsname{0.421}
\expandafter\def\csname jl@/pairwise/466/weighted_kappa\endcsname{0.427}
\expandafter\def\csname jl@/pairwise/467/weighted_kappa\endcsname{0.259}
\expandafter\def\csname jl@/pairwise/468/weighted_kappa\endcsname{0.310}
\expandafter\def\csname jl@/pairwise/469/weighted_kappa\endcsname{0.486}
\expandafter\def\csname jl@/pairwise/47/weighted_kappa\endcsname{0.635}
\expandafter\def\csname jl@/pairwise/470/weighted_kappa\endcsname{0.794}
\expandafter\def\csname jl@/pairwise/471/weighted_kappa\endcsname{0.807}
\expandafter\def\csname jl@/pairwise/472/weighted_kappa\endcsname{0.832}
\expandafter\def\csname jl@/pairwise/473/weighted_kappa\endcsname{0.888}
\expandafter\def\csname jl@/pairwise/474/weighted_kappa\endcsname{0.817}
\expandafter\def\csname jl@/pairwise/475/weighted_kappa\endcsname{0.726}
\expandafter\def\csname jl@/pairwise/476/weighted_kappa\endcsname{0.780}
\expandafter\def\csname jl@/pairwise/477/weighted_kappa\endcsname{0.739}
\expandafter\def\csname jl@/pairwise/478/weighted_kappa\endcsname{0.779}
\expandafter\def\csname jl@/pairwise/479/weighted_kappa\endcsname{0.790}
\expandafter\def\csname jl@/pairwise/48/weighted_kappa\endcsname{0.704}
\expandafter\def\csname jl@/pairwise/480/weighted_kappa\endcsname{0.810}
\expandafter\def\csname jl@/pairwise/481/weighted_kappa\endcsname{0.668}
\expandafter\def\csname jl@/pairwise/482/weighted_kappa\endcsname{0.734}
\expandafter\def\csname jl@/pairwise/483/weighted_kappa\endcsname{0.827}
\expandafter\def\csname jl@/pairwise/484/weighted_kappa\endcsname{0.791}
\expandafter\def\csname jl@/pairwise/485/weighted_kappa\endcsname{0.679}
\expandafter\def\csname jl@/pairwise/486/weighted_kappa\endcsname{0.799}
\expandafter\def\csname jl@/pairwise/487/weighted_kappa\endcsname{0.566}
\expandafter\def\csname jl@/pairwise/488/weighted_kappa\endcsname{0.674}
\expandafter\def\csname jl@/pairwise/489/weighted_kappa\endcsname{0.706}
\expandafter\def\csname jl@/pairwise/49/weighted_kappa\endcsname{0.834}
\expandafter\def\csname jl@/pairwise/490/weighted_kappa\endcsname{0.772}
\expandafter\def\csname jl@/pairwise/491/weighted_kappa\endcsname{0.528}
\expandafter\def\csname jl@/pairwise/492/weighted_kappa\endcsname{0.633}
\expandafter\def\csname jl@/pairwise/493/weighted_kappa\endcsname{0.578}
\expandafter\def\csname jl@/pairwise/494/weighted_kappa\endcsname{0.529}
\expandafter\def\csname jl@/pairwise/495/weighted_kappa\endcsname{0.625}
\expandafter\def\csname jl@/pairwise/496/weighted_kappa\endcsname{0.628}
\expandafter\def\csname jl@/pairwise/497/weighted_kappa\endcsname{0.434}
\expandafter\def\csname jl@/pairwise/498/weighted_kappa\endcsname{0.599}
\expandafter\def\csname jl@/pairwise/499/weighted_kappa\endcsname{0.577}
\expandafter\def\csname jl@/pairwise/5/weighted_kappa\endcsname{0.659}
\expandafter\def\csname jl@/pairwise/50/weighted_kappa\endcsname{0.917}
\expandafter\def\csname jl@/pairwise/500/weighted_kappa\endcsname{0.852}
\expandafter\def\csname jl@/pairwise/501/weighted_kappa\endcsname{0.701}
\expandafter\def\csname jl@/pairwise/502/weighted_kappa\endcsname{0.650}
\expandafter\def\csname jl@/pairwise/503/weighted_kappa\endcsname{0.743}
\expandafter\def\csname jl@/pairwise/504/weighted_kappa\endcsname{0.744}
\expandafter\def\csname jl@/pairwise/505/weighted_kappa\endcsname{0.680}
\expandafter\def\csname jl@/pairwise/506/weighted_kappa\endcsname{0.743}
\expandafter\def\csname jl@/pairwise/507/weighted_kappa\endcsname{0.682}
\expandafter\def\csname jl@/pairwise/508/weighted_kappa\endcsname{0.695}
\expandafter\def\csname jl@/pairwise/509/weighted_kappa\endcsname{0.655}
\expandafter\def\csname jl@/pairwise/51/weighted_kappa\endcsname{0.885}
\expandafter\def\csname jl@/pairwise/510/weighted_kappa\endcsname{0.840}
\expandafter\def\csname jl@/pairwise/511/weighted_kappa\endcsname{0.718}
\expandafter\def\csname jl@/pairwise/512/weighted_kappa\endcsname{0.678}
\expandafter\def\csname jl@/pairwise/513/weighted_kappa\endcsname{0.705}
\expandafter\def\csname jl@/pairwise/514/weighted_kappa\endcsname{0.652}
\expandafter\def\csname jl@/pairwise/515/weighted_kappa\endcsname{0.576}
\expandafter\def\csname jl@/pairwise/516/weighted_kappa\endcsname{0.585}
\expandafter\def\csname jl@/pairwise/517/weighted_kappa\endcsname{0.607}
\expandafter\def\csname jl@/pairwise/518/weighted_kappa\endcsname{0.670}
\expandafter\def\csname jl@/pairwise/519/weighted_kappa\endcsname{0.740}
\expandafter\def\csname jl@/pairwise/52/weighted_kappa\endcsname{0.872}
\expandafter\def\csname jl@/pairwise/520/weighted_kappa\endcsname{0.645}
\expandafter\def\csname jl@/pairwise/521/weighted_kappa\endcsname{0.616}
\expandafter\def\csname jl@/pairwise/522/weighted_kappa\endcsname{0.346}
\expandafter\def\csname jl@/pairwise/523/weighted_kappa\endcsname{0.518}
\expandafter\def\csname jl@/pairwise/524/weighted_kappa\endcsname{0.587}
\expandafter\def\csname jl@/pairwise/525/weighted_kappa\endcsname{0.542}
\expandafter\def\csname jl@/pairwise/526/weighted_kappa\endcsname{0.612}
\expandafter\def\csname jl@/pairwise/527/weighted_kappa\endcsname{0.299}
\expandafter\def\csname jl@/pairwise/528/weighted_kappa\endcsname{0.488}
\expandafter\def\csname jl@/pairwise/529/weighted_kappa\endcsname{0.516}
\expandafter\def\csname jl@/pairwise/53/weighted_kappa\endcsname{0.799}
\expandafter\def\csname jl@/pairwise/530/weighted_kappa\endcsname{0.862}
\expandafter\def\csname jl@/pairwise/531/weighted_kappa\endcsname{0.813}
\expandafter\def\csname jl@/pairwise/532/weighted_kappa\endcsname{0.819}
\expandafter\def\csname jl@/pairwise/533/weighted_kappa\endcsname{0.766}
\expandafter\def\csname jl@/pairwise/534/weighted_kappa\endcsname{0.812}
\expandafter\def\csname jl@/pairwise/535/weighted_kappa\endcsname{0.834}
\expandafter\def\csname jl@/pairwise/536/weighted_kappa\endcsname{0.719}
\expandafter\def\csname jl@/pairwise/537/weighted_kappa\endcsname{0.813}
\expandafter\def\csname jl@/pairwise/538/weighted_kappa\endcsname{0.709}
\expandafter\def\csname jl@/pairwise/539/weighted_kappa\endcsname{0.728}
\expandafter\def\csname jl@/pairwise/54/weighted_kappa\endcsname{0.869}
\expandafter\def\csname jl@/pairwise/540/weighted_kappa\endcsname{0.850}
\expandafter\def\csname jl@/pairwise/541/weighted_kappa\endcsname{0.632}
\expandafter\def\csname jl@/pairwise/542/weighted_kappa\endcsname{0.803}
\expandafter\def\csname jl@/pairwise/543/weighted_kappa\endcsname{0.741}
\expandafter\def\csname jl@/pairwise/544/weighted_kappa\endcsname{0.728}
\expandafter\def\csname jl@/pairwise/545/weighted_kappa\endcsname{0.779}
\expandafter\def\csname jl@/pairwise/546/weighted_kappa\endcsname{0.800}
\expandafter\def\csname jl@/pairwise/547/weighted_kappa\endcsname{0.621}
\expandafter\def\csname jl@/pairwise/548/weighted_kappa\endcsname{0.735}
\expandafter\def\csname jl@/pairwise/549/weighted_kappa\endcsname{0.714}
\expandafter\def\csname jl@/pairwise/55/weighted_kappa\endcsname{0.804}
\expandafter\def\csname jl@/pairwise/550/weighted_kappa\endcsname{0.707}
\expandafter\def\csname jl@/pairwise/551/weighted_kappa\endcsname{0.435}
\expandafter\def\csname jl@/pairwise/552/weighted_kappa\endcsname{0.447}
\expandafter\def\csname jl@/pairwise/553/weighted_kappa\endcsname{0.520}
\expandafter\def\csname jl@/pairwise/554/weighted_kappa\endcsname{0.449}
\expandafter\def\csname jl@/pairwise/555/weighted_kappa\endcsname{0.518}
\expandafter\def\csname jl@/pairwise/556/weighted_kappa\endcsname{0.542}
\expandafter\def\csname jl@/pairwise/557/weighted_kappa\endcsname{0.323}
\expandafter\def\csname jl@/pairwise/558/weighted_kappa\endcsname{0.474}
\expandafter\def\csname jl@/pairwise/559/weighted_kappa\endcsname{0.428}
\expandafter\def\csname jl@/pairwise/56/weighted_kappa\endcsname{0.727}
\expandafter\def\csname jl@/pairwise/560/weighted_kappa\endcsname{0.836}
\expandafter\def\csname jl@/pairwise/561/weighted_kappa\endcsname{0.691}
\expandafter\def\csname jl@/pairwise/562/weighted_kappa\endcsname{0.623}
\expandafter\def\csname jl@/pairwise/563/weighted_kappa\endcsname{0.713}
\expandafter\def\csname jl@/pairwise/564/weighted_kappa\endcsname{0.691}
\expandafter\def\csname jl@/pairwise/565/weighted_kappa\endcsname{0.610}
\expandafter\def\csname jl@/pairwise/566/weighted_kappa\endcsname{0.730}
\expandafter\def\csname jl@/pairwise/567/weighted_kappa\endcsname{0.683}
\expandafter\def\csname jl@/pairwise/568/weighted_kappa\endcsname{0.660}
\expandafter\def\csname jl@/pairwise/569/weighted_kappa\endcsname{0.581}
\expandafter\def\csname jl@/pairwise/57/weighted_kappa\endcsname{0.766}
\expandafter\def\csname jl@/pairwise/570/weighted_kappa\endcsname{0.832}
\expandafter\def\csname jl@/pairwise/571/weighted_kappa\endcsname{0.804}
\expandafter\def\csname jl@/pairwise/572/weighted_kappa\endcsname{0.744}
\expandafter\def\csname jl@/pairwise/573/weighted_kappa\endcsname{0.695}
\expandafter\def\csname jl@/pairwise/574/weighted_kappa\endcsname{0.751}
\expandafter\def\csname jl@/pairwise/575/weighted_kappa\endcsname{0.664}
\expandafter\def\csname jl@/pairwise/576/weighted_kappa\endcsname{0.558}
\expandafter\def\csname jl@/pairwise/577/weighted_kappa\endcsname{0.678}
\expandafter\def\csname jl@/pairwise/578/weighted_kappa\endcsname{0.622}
\expandafter\def\csname jl@/pairwise/579/weighted_kappa\endcsname{0.714}
\expandafter\def\csname jl@/pairwise/58/weighted_kappa\endcsname{0.723}
\expandafter\def\csname jl@/pairwise/580/weighted_kappa\endcsname{0.766}
\expandafter\def\csname jl@/pairwise/581/weighted_kappa\endcsname{0.661}
\expandafter\def\csname jl@/pairwise/582/weighted_kappa\endcsname{0.529}
\expandafter\def\csname jl@/pairwise/583/weighted_kappa\endcsname{0.592}
\expandafter\def\csname jl@/pairwise/584/weighted_kappa\endcsname{0.639}
\expandafter\def\csname jl@/pairwise/585/weighted_kappa\endcsname{0.544}
\expandafter\def\csname jl@/pairwise/586/weighted_kappa\endcsname{0.629}
\expandafter\def\csname jl@/pairwise/587/weighted_kappa\endcsname{0.524}
\expandafter\def\csname jl@/pairwise/588/weighted_kappa\endcsname{0.622}
\expandafter\def\csname jl@/pairwise/589/weighted_kappa\endcsname{0.613}
\expandafter\def\csname jl@/pairwise/59/weighted_kappa\endcsname{0.820}
\expandafter\def\csname jl@/pairwise/590/weighted_kappa\endcsname{0.938}
\expandafter\def\csname jl@/pairwise/591/weighted_kappa\endcsname{0.910}
\expandafter\def\csname jl@/pairwise/592/weighted_kappa\endcsname{0.903}
\expandafter\def\csname jl@/pairwise/593/weighted_kappa\endcsname{0.958}
\expandafter\def\csname jl@/pairwise/594/weighted_kappa\endcsname{0.889}
\expandafter\def\csname jl@/pairwise/595/weighted_kappa\endcsname{0.896}
\expandafter\def\csname jl@/pairwise/596/weighted_kappa\endcsname{0.938}
\expandafter\def\csname jl@/pairwise/597/weighted_kappa\endcsname{0.868}
\expandafter\def\csname jl@/pairwise/598/weighted_kappa\endcsname{0.896}
\expandafter\def\csname jl@/pairwise/599/weighted_kappa\endcsname{0.903}
\expandafter\def\csname jl@/pairwise/6/weighted_kappa\endcsname{0.688}
\expandafter\def\csname jl@/pairwise/60/weighted_kappa\endcsname{0.655}
\expandafter\def\csname jl@/pairwise/600/weighted_kappa\endcsname{0.857}
\expandafter\def\csname jl@/pairwise/601/weighted_kappa\endcsname{0.576}
\expandafter\def\csname jl@/pairwise/602/weighted_kappa\endcsname{0.690}
\expandafter\def\csname jl@/pairwise/603/weighted_kappa\endcsname{0.802}
\expandafter\def\csname jl@/pairwise/604/weighted_kappa\endcsname{0.691}
\expandafter\def\csname jl@/pairwise/605/weighted_kappa\endcsname{0.642}
\expandafter\def\csname jl@/pairwise/606/weighted_kappa\endcsname{0.832}
\expandafter\def\csname jl@/pairwise/607/weighted_kappa\endcsname{0.483}
\expandafter\def\csname jl@/pairwise/608/weighted_kappa\endcsname{0.639}
\expandafter\def\csname jl@/pairwise/609/weighted_kappa\endcsname{0.679}
\expandafter\def\csname jl@/pairwise/61/weighted_kappa\endcsname{0.588}
\expandafter\def\csname jl@/pairwise/610/weighted_kappa\endcsname{0.786}
\expandafter\def\csname jl@/pairwise/611/weighted_kappa\endcsname{0.537}
\expandafter\def\csname jl@/pairwise/612/weighted_kappa\endcsname{0.522}
\expandafter\def\csname jl@/pairwise/613/weighted_kappa\endcsname{0.529}
\expandafter\def\csname jl@/pairwise/614/weighted_kappa\endcsname{0.555}
\expandafter\def\csname jl@/pairwise/615/weighted_kappa\endcsname{0.620}
\expandafter\def\csname jl@/pairwise/616/weighted_kappa\endcsname{0.588}
\expandafter\def\csname jl@/pairwise/617/weighted_kappa\endcsname{0.400}
\expandafter\def\csname jl@/pairwise/618/weighted_kappa\endcsname{0.563}
\expandafter\def\csname jl@/pairwise/619/weighted_kappa\endcsname{0.464}
\expandafter\def\csname jl@/pairwise/62/weighted_kappa\endcsname{0.266}
\expandafter\def\csname jl@/pairwise/620/weighted_kappa\endcsname{0.827}
\expandafter\def\csname jl@/pairwise/621/weighted_kappa\endcsname{0.742}
\expandafter\def\csname jl@/pairwise/622/weighted_kappa\endcsname{0.732}
\expandafter\def\csname jl@/pairwise/623/weighted_kappa\endcsname{0.710}
\expandafter\def\csname jl@/pairwise/624/weighted_kappa\endcsname{0.745}
\expandafter\def\csname jl@/pairwise/625/weighted_kappa\endcsname{0.693}
\expandafter\def\csname jl@/pairwise/626/weighted_kappa\endcsname{0.737}
\expandafter\def\csname jl@/pairwise/627/weighted_kappa\endcsname{0.660}
\expandafter\def\csname jl@/pairwise/628/weighted_kappa\endcsname{0.725}
\expandafter\def\csname jl@/pairwise/629/weighted_kappa\endcsname{0.676}
\expandafter\def\csname jl@/pairwise/63/weighted_kappa\endcsname{0.367}
\expandafter\def\csname jl@/pairwise/630/weighted_kappa\endcsname{0.900}
\expandafter\def\csname jl@/pairwise/631/weighted_kappa\endcsname{0.772}
\expandafter\def\csname jl@/pairwise/632/weighted_kappa\endcsname{0.751}
\expandafter\def\csname jl@/pairwise/633/weighted_kappa\endcsname{0.658}
\expandafter\def\csname jl@/pairwise/634/weighted_kappa\endcsname{0.760}
\expandafter\def\csname jl@/pairwise/635/weighted_kappa\endcsname{0.697}
\expandafter\def\csname jl@/pairwise/636/weighted_kappa\endcsname{0.621}
\expandafter\def\csname jl@/pairwise/637/weighted_kappa\endcsname{0.639}
\expandafter\def\csname jl@/pairwise/638/weighted_kappa\endcsname{0.558}
\expandafter\def\csname jl@/pairwise/639/weighted_kappa\endcsname{0.655}
\expandafter\def\csname jl@/pairwise/64/weighted_kappa\endcsname{0.529}
\expandafter\def\csname jl@/pairwise/640/weighted_kappa\endcsname{0.759}
\expandafter\def\csname jl@/pairwise/641/weighted_kappa\endcsname{0.695}
\expandafter\def\csname jl@/pairwise/642/weighted_kappa\endcsname{0.440}
\expandafter\def\csname jl@/pairwise/643/weighted_kappa\endcsname{0.470}
\expandafter\def\csname jl@/pairwise/644/weighted_kappa\endcsname{0.632}
\expandafter\def\csname jl@/pairwise/645/weighted_kappa\endcsname{0.478}
\expandafter\def\csname jl@/pairwise/646/weighted_kappa\endcsname{0.553}
\expandafter\def\csname jl@/pairwise/647/weighted_kappa\endcsname{0.525}
\expandafter\def\csname jl@/pairwise/648/weighted_kappa\endcsname{0.497}
\expandafter\def\csname jl@/pairwise/649/weighted_kappa\endcsname{0.565}
\expandafter\def\csname jl@/pairwise/65/weighted_kappa\endcsname{0.322}
\expandafter\def\csname jl@/pairwise/650/weighted_kappa\endcsname{0.881}
\expandafter\def\csname jl@/pairwise/651/weighted_kappa\endcsname{0.816}
\expandafter\def\csname jl@/pairwise/652/weighted_kappa\endcsname{0.859}
\expandafter\def\csname jl@/pairwise/653/weighted_kappa\endcsname{0.839}
\expandafter\def\csname jl@/pairwise/654/weighted_kappa\endcsname{0.810}
\expandafter\def\csname jl@/pairwise/655/weighted_kappa\endcsname{0.853}
\expandafter\def\csname jl@/pairwise/656/weighted_kappa\endcsname{0.861}
\expandafter\def\csname jl@/pairwise/657/weighted_kappa\endcsname{0.802}
\expandafter\def\csname jl@/pairwise/658/weighted_kappa\endcsname{0.797}
\expandafter\def\csname jl@/pairwise/659/weighted_kappa\endcsname{0.854}
\expandafter\def\csname jl@/pairwise/66/weighted_kappa\endcsname{0.241}
\expandafter\def\csname jl@/pairwise/67/weighted_kappa\endcsname{0.286}
\expandafter\def\csname jl@/pairwise/68/weighted_kappa\endcsname{0.370}
\expandafter\def\csname jl@/pairwise/69/weighted_kappa\endcsname{0.127}
\expandafter\def\csname jl@/pairwise/7/weighted_kappa\endcsname{0.548}
\expandafter\def\csname jl@/pairwise/70/weighted_kappa\endcsname{0.776}
\expandafter\def\csname jl@/pairwise/71/weighted_kappa\endcsname{0.630}
\expandafter\def\csname jl@/pairwise/72/weighted_kappa\endcsname{0.712}
\expandafter\def\csname jl@/pairwise/73/weighted_kappa\endcsname{0.460}
\expandafter\def\csname jl@/pairwise/74/weighted_kappa\endcsname{0.660}
\expandafter\def\csname jl@/pairwise/75/weighted_kappa\endcsname{0.645}
\expandafter\def\csname jl@/pairwise/76/weighted_kappa\endcsname{0.488}
\expandafter\def\csname jl@/pairwise/77/weighted_kappa\endcsname{0.565}
\expandafter\def\csname jl@/pairwise/78/weighted_kappa\endcsname{0.565}
\expandafter\def\csname jl@/pairwise/79/weighted_kappa\endcsname{0.469}
\expandafter\def\csname jl@/pairwise/8/weighted_kappa\endcsname{0.628}
\expandafter\def\csname jl@/pairwise/80/weighted_kappa\endcsname{0.826}
\expandafter\def\csname jl@/pairwise/81/weighted_kappa\endcsname{0.736}
\expandafter\def\csname jl@/pairwise/82/weighted_kappa\endcsname{0.550}
\expandafter\def\csname jl@/pairwise/83/weighted_kappa\endcsname{0.641}
\expandafter\def\csname jl@/pairwise/84/weighted_kappa\endcsname{0.710}
\expandafter\def\csname jl@/pairwise/85/weighted_kappa\endcsname{0.588}
\expandafter\def\csname jl@/pairwise/86/weighted_kappa\endcsname{0.686}
\expandafter\def\csname jl@/pairwise/87/weighted_kappa\endcsname{0.451}
\expandafter\def\csname jl@/pairwise/88/weighted_kappa\endcsname{0.627}
\expandafter\def\csname jl@/pairwise/89/weighted_kappa\endcsname{0.737}
\expandafter\def\csname jl@/pairwise/9/weighted_kappa\endcsname{0.808}
\expandafter\def\csname jl@/pairwise/90/weighted_kappa\endcsname{0.940}
\expandafter\def\csname jl@/pairwise/91/weighted_kappa\endcsname{0.864}
\expandafter\def\csname jl@/pairwise/92/weighted_kappa\endcsname{0.904}
\expandafter\def\csname jl@/pairwise/93/weighted_kappa\endcsname{0.830}
\expandafter\def\csname jl@/pairwise/94/weighted_kappa\endcsname{0.846}
\expandafter\def\csname jl@/pairwise/95/weighted_kappa\endcsname{0.878}
\expandafter\def\csname jl@/pairwise/96/weighted_kappa\endcsname{0.825}
\expandafter\def\csname jl@/pairwise/97/weighted_kappa\endcsname{0.786}
\expandafter\def\csname jl@/pairwise/98/weighted_kappa\endcsname{0.732}
\expandafter\def\csname jl@/pairwise/99/weighted_kappa\endcsname{0.786}
\expandafter\def\csname jl@/physician_ties_excluded\endcsname{143.000}
\expandafter\def\csname jl@/residual/0/weighted_kappa\endcsname{0.707}
\expandafter\def\csname jl@/residual/1/weighted_kappa\endcsname{0.578}
\expandafter\def\csname jl@/residual/10/disagreement_rate\endcsname{0.092}
\expandafter\def\csname jl@/residual/10/pairs\endcsname{768.000}
\expandafter\def\csname jl@/residual/11/disagreement_rate\endcsname{0.013}
\expandafter\def\csname jl@/residual/11/pairs\endcsname{768.000}
\expandafter\def\csname jl@/residual/12/disagreement_rate\endcsname{0.214}
\expandafter\def\csname jl@/residual/12/pairs\endcsname{768.000}
\expandafter\def\csname jl@/residual/13/disagreement_rate\endcsname{0.051}
\expandafter\def\csname jl@/residual/13/pairs\endcsname{768.000}
\expandafter\def\csname jl@/residual/14/delta\endcsname{-0.022}
\expandafter\def\csname jl@/residual/15/weighted_kappa\endcsname{0.410}
\expandafter\def\csname jl@/residual/16/weighted_kappa\endcsname{0.378}
\expandafter\def\csname jl@/residual/17/disagreement_rate\endcsname{0.033}
\expandafter\def\csname jl@/residual/17/pairs\endcsname{486.000}
\expandafter\def\csname jl@/residual/18/disagreement_rate\endcsname{0.021}
\expandafter\def\csname jl@/residual/18/pairs\endcsname{486.000}
\expandafter\def\csname jl@/residual/19/disagreement_rate\endcsname{0.000}
\expandafter\def\csname jl@/residual/19/pairs\endcsname{486.000}
\expandafter\def\csname jl@/residual/2/disagreement_rate\endcsname{0.086}
\expandafter\def\csname jl@/residual/2/pairs\endcsname{768.000}
\expandafter\def\csname jl@/residual/20/disagreement_rate\endcsname{0.086}
\expandafter\def\csname jl@/residual/20/pairs\endcsname{486.000}
\expandafter\def\csname jl@/residual/21/disagreement_rate\endcsname{0.025}
\expandafter\def\csname jl@/residual/21/pairs\endcsname{486.000}
\expandafter\def\csname jl@/residual/22/weighted_kappa\endcsname{0.418}
\expandafter\def\csname jl@/residual/23/weighted_kappa\endcsname{0.339}
\expandafter\def\csname jl@/residual/24/disagreement_rate\endcsname{0.029}
\expandafter\def\csname jl@/residual/24/pairs\endcsname{483.000}
\expandafter\def\csname jl@/residual/25/disagreement_rate\endcsname{0.016}
\expandafter\def\csname jl@/residual/25/pairs\endcsname{486.000}
\expandafter\def\csname jl@/residual/26/disagreement_rate\endcsname{0.000}
\expandafter\def\csname jl@/residual/26/pairs\endcsname{486.000}
\expandafter\def\csname jl@/residual/27/disagreement_rate\endcsname{0.099}
\expandafter\def\csname jl@/residual/27/pairs\endcsname{483.000}
\expandafter\def\csname jl@/residual/28/disagreement_rate\endcsname{0.025}
\expandafter\def\csname jl@/residual/28/pairs\endcsname{486.000}
\expandafter\def\csname jl@/residual/29/delta\endcsname{-0.006}
\expandafter\def\csname jl@/residual/3/disagreement_rate\endcsname{0.062}
\expandafter\def\csname jl@/residual/3/pairs\endcsname{768.000}
\expandafter\def\csname jl@/residual/30/weighted_kappa\endcsname{0.704}
\expandafter\def\csname jl@/residual/31/weighted_kappa\endcsname{0.635}
\expandafter\def\csname jl@/residual/32/disagreement_rate\endcsname{0.039}
\expandafter\def\csname jl@/residual/32/pairs\endcsname{1080.000}
\expandafter\def\csname jl@/residual/33/disagreement_rate\endcsname{0.033}
\expandafter\def\csname jl@/residual/33/pairs\endcsname{1080.000}
\expandafter\def\csname jl@/residual/34/disagreement_rate\endcsname{0.002}
\expandafter\def\csname jl@/residual/34/pairs\endcsname{1080.000}
\expandafter\def\csname jl@/residual/35/disagreement_rate\endcsname{0.135}
\expandafter\def\csname jl@/residual/35/pairs\endcsname{1080.000}
\expandafter\def\csname jl@/residual/36/disagreement_rate\endcsname{0.074}
\expandafter\def\csname jl@/residual/36/pairs\endcsname{1080.000}
\expandafter\def\csname jl@/residual/37/weighted_kappa\endcsname{0.734}
\expandafter\def\csname jl@/residual/38/weighted_kappa\endcsname{0.648}
\expandafter\def\csname jl@/residual/39/disagreement_rate\endcsname{0.037}
\expandafter\def\csname jl@/residual/39/pairs\endcsname{1080.000}
\expandafter\def\csname jl@/residual/4/disagreement_rate\endcsname{0.022}
\expandafter\def\csname jl@/residual/4/pairs\endcsname{768.000}
\expandafter\def\csname jl@/residual/40/disagreement_rate\endcsname{0.024}
\expandafter\def\csname jl@/residual/40/pairs\endcsname{1080.000}
\expandafter\def\csname jl@/residual/41/disagreement_rate\endcsname{0.004}
\expandafter\def\csname jl@/residual/41/pairs\endcsname{1080.000}
\expandafter\def\csname jl@/residual/42/disagreement_rate\endcsname{0.113}
\expandafter\def\csname jl@/residual/42/pairs\endcsname{1080.000}
\expandafter\def\csname jl@/residual/43/disagreement_rate\endcsname{0.074}
\expandafter\def\csname jl@/residual/43/pairs\endcsname{1080.000}
\expandafter\def\csname jl@/residual/44/delta\endcsname{-0.012}
\expandafter\def\csname jl@/residual/5/disagreement_rate\endcsname{0.246}
\expandafter\def\csname jl@/residual/5/pairs\endcsname{768.000}
\expandafter\def\csname jl@/residual/6/disagreement_rate\endcsname{0.102}
\expandafter\def\csname jl@/residual/6/pairs\endcsname{768.000}
\expandafter\def\csname jl@/residual/7/weighted_kappa\endcsname{0.757}
\expandafter\def\csname jl@/residual/8/weighted_kappa\endcsname{0.654}
\expandafter\def\csname jl@/residual/9/disagreement_rate\endcsname{0.113}
\expandafter\def\csname jl@/residual/9/pairs\endcsname{768.000}
\expandafter\def\csname jl@/seed\endcsname{20261005.000}

\newcommand{\JLValue}[1]{\ifcsname jlm@#1\endcsname\csname jlm@#1\endcsname\else\errmessage{Unknown manuscript number}\fi}
\expandafter\def\csname jlm@V0001\endcsname{0.021}
\expandafter\def\csname jlm@V0002\endcsname{-0.037}
\expandafter\def\csname jlm@V0003\endcsname{95}
\expandafter\def\csname jlm@V0004\endcsname{-0.007}
\expandafter\def\csname jlm@V0005\endcsname{0.074}
\expandafter\def\csname jlm@V0006\endcsname{0.092}
\expandafter\def\csname jlm@V0007\endcsname{0.046}
\expandafter\def\csname jlm@V0008\endcsname{95}
\expandafter\def\csname jlm@V0009\endcsname{0.069}
\expandafter\def\csname jlm@V0010\endcsname{129,770}
\expandafter\def\csname jlm@V0011\endcsname{75}
\expandafter\def\csname jlm@V0012\endcsname{34}
\expandafter\def\csname jlm@V0013\endcsname{0.9}
\expandafter\def\csname jlm@V0014\endcsname{75}
\expandafter\def\csname jlm@V0015\endcsname{0.9}
\expandafter\def\csname jlm@V0016\endcsname{2}
\expandafter\def\csname jlm@V0017\endcsname{5}
\expandafter\def\csname jlm@V0018\endcsname{10}
\expandafter\def\csname jlm@V0019\endcsname{2}
\expandafter\def\csname jlm@V0020\endcsname{20}
\expandafter\def\csname jlm@V0021\endcsname{20261005}
\expandafter\def\csname jlm@V0022\endcsname{80}
\expandafter\def\csname jlm@V0023\endcsname{20}
\expandafter\def\csname jlm@V0024\endcsname{1,024}
\expandafter\def\csname jlm@V0025\endcsname{512}
\expandafter\def\csname jlm@V0026\endcsname{64}
\expandafter\def\csname jlm@V0027\endcsname{20261005}
\expandafter\def\csname jlm@V0028\endcsname{10}
\expandafter\def\csname jlm@V0029\endcsname{30}
\expandafter\def\csname jlm@V0030\endcsname{34}
\expandafter\def\csname jlm@V0031\endcsname{5}
\expandafter\def\csname jlm@V0032\endcsname{3}
\expandafter\def\csname jlm@V0033\endcsname{5}
\expandafter\def\csname jlm@V0034\endcsname{3}
\expandafter\def\csname jlm@V0035\endcsname{0.05}
\expandafter\def\csname jlm@V0036\endcsname{0.05}
\expandafter\def\csname jlm@V0037\endcsname{95}
\expandafter\def\csname jlm@V0038\endcsname{10,000}
\expandafter\def\csname jlm@ManifestPrefix\endcsname{3f49c9c4}
\expandafter\def\csname jlm@V0040\endcsname{23}
\expandafter\def\csname jlm@V0041\endcsname{16}
\expandafter\def\csname jlm@V0042\endcsname{23}
\expandafter\def\csname jlm@V0043\endcsname{12}
\expandafter\def\csname jlm@V0044\endcsname{23}
\expandafter\def\csname jlm@V0045\endcsname{11}
\expandafter\def\csname jlm@V0046\endcsname{236.69}
\expandafter\def\csname jlm@V0047\endcsname{0.105}
\expandafter\def\csname jlm@V0048\endcsname{0.021}
\expandafter\def\csname jlm@V0049\endcsname{27}
\expandafter\def\csname jlm@V0050\endcsname{23.95}
\expandafter\def\csname jlm@V0051\endcsname{25,954}
\expandafter\def\csname jlm@V0052\endcsname{129,770}
\expandafter\def\csname jlm@V0053\endcsname{0.05}
\expandafter\def\csname jlm@V0054\endcsname{1,728}
\expandafter\def\csname jlm@V0055\endcsname{1,646}
\expandafter\def\csname jlm@V0056\endcsname{1,280}
\expandafter\def\csname jlm@V0057\endcsname{974}
\expandafter\def\csname jlm@V0058\endcsname{0.541}
\expandafter\def\csname jlm@V0059\endcsname{-0.035}
\expandafter\def\csname jlm@V0060\endcsname{-0.088}
\expandafter\def\csname jlm@V0061\endcsname{95}
\expandafter\def\csname jlm@V0062\endcsname{0.062}
\expandafter\def\csname jlm@V0063\endcsname{0.619}
\expandafter\def\csname jlm@V0064\endcsname{0.689}
\expandafter\def\csname jlm@V0065\endcsname{0.389}
\expandafter\def\csname jlm@V0066\endcsname{+0.0001}
\expandafter\def\csname jlm@V0067\endcsname{0.060}
\expandafter\def\csname jlm@V0068\endcsname{0.001}
\expandafter\def\csname jlm@V0069\endcsname{-0.025}
\expandafter\def\csname jlm@V0070\endcsname{-0.012}
\expandafter\def\csname jlm@V0071\endcsname{0.497}
\expandafter\def\csname jlm@V0072\endcsname{0.013}
\expandafter\def\csname jlm@V0073\endcsname{-0.027}
\expandafter\def\csname jlm@V0074\endcsname{-0.007}
\expandafter\def\csname jlm@V0075\endcsname{0.049}
\expandafter\def\csname jlm@V0076\endcsname{0.000}
\expandafter\def\csname jlm@V0077\endcsname{-0.043}
\expandafter\def\csname jlm@V0078\endcsname{95}
\expandafter\def\csname jlm@V0079\endcsname{-0.022}
\expandafter\def\csname jlm@V0080\endcsname{0.735}
\expandafter\def\csname jlm@V0081\endcsname{0.748}
\expandafter\def\csname jlm@V0082\endcsname{0.494}
\expandafter\def\csname jlm@V0083\endcsname{0.516}
\expandafter\def\csname jlm@V0084\endcsname{0.016}
\expandafter\def\csname jlm@V0085\endcsname{0.090}
\expandafter\def\csname jlm@V0086\endcsname{0.358}
\expandafter\def\csname jlm@V0087\endcsname{0.375}
\expandafter\def\csname jlm@V0088\endcsname{0.556}
\expandafter\def\csname jlm@V0089\endcsname{0.577}
\expandafter\def\csname jlm@V0090\endcsname{0.069}
\expandafter\def\csname jlm@V0091\endcsname{0.074}
\expandafter\def\csname jlm@V0092\endcsname{0.155}
\expandafter\def\csname jlm@V0093\endcsname{0.023}
\expandafter\def\csname jlm@V0094\endcsname{0.070}
\expandafter\def\csname jlm@V0095\endcsname{0.05}
\expandafter\def\csname jlm@V0096\endcsname{0.027}
\expandafter\def\csname jlm@V0097\endcsname{1,080}
\expandafter\def\csname jlm@V0098\endcsname{483}
\expandafter\def\csname jlm@V0099\endcsname{0.246}
\expandafter\def\csname jlm@V0100\endcsname{0}
\expandafter\def\csname jlm@V0101\endcsname{15}
\expandafter\def\csname jlm@V0102\endcsname{799}
\expandafter\def\csname jlm@V0103\endcsname{120}
\expandafter\def\csname jlm@V0104\endcsname{34}
\expandafter\def\csname jlm@V0105\endcsname{5}
\expandafter\def\csname jlm@V0106\endcsname{0.01}
\expandafter\def\csname jlm@V0107\endcsname{20}
\expandafter\def\csname jlm@V0108\endcsname{20}
\expandafter\def\csname jlm@V0109\endcsname{20}
\expandafter\def\csname jlm@V0110\endcsname{0}
\expandafter\def\csname jlm@V0111\endcsname{75}

\title{Agreement Without Alignment: Label-Free Rubric Clarification Increases LLM-Judge Consensus but Not Agreement with Humans}
\author{Sangdo Han, Moonsoo Kim \\ CEO Business School \\ \texttt{\{sangdohan, mskim\}@ceobizschool.kr}}
\begin{document}
\pagestyle{plain}\thispagestyle{plain}
\maketitle

\begin{abstract}

LLM judges are increasingly deployed as panels, and disagreement among panel members is often read as a symptom of an unclear rubric. We test a label-free procedure that turns this disagreement into rubric repairs. The procedure decomposes each rubric into sub-criteria, ranks them by cross-judge disagreement on a development split, and has an editor model rewrite only the most contested ones. An additional stage that moves text-checkable sub-criteria to deterministic checkers transferred none in our evaluation. Human labels are never used to select or edit clauses; they are used only to evaluate the outcome. We pre-specified four hypotheses in a version-controlled protocol and evaluated six rubric conditions with a panel of five judges from three providers on SummEval, the HealthBench meta-evaluation, and USR Topical-Chat, for a total of \JLValue{V0010} judge requests, excluding retries. On SummEval, the procedure raised inter-judge Krippendorff's \ensuremath{\alpha} by \JLValue{V0009} (\JLValue{V0008}\% CI \JLValue{V0007} to \JLValue{V0006}) over the original rubric and by \JLValue{V0005} over a control that applied the same edits to an equal number of randomly chosen clauses that were not selected. Agreement with expert ratings showed no detectable change (Kendall \ensuremath{\tau} difference \JLValue{V0004}, \JLValue{V0003}\% CI \JLValue{V0002} to \JLValue{V0001}). On HealthBench and USR, neither inter-judge agreement nor human agreement met the preregistered effect criteria. In a preregistered secondary comparison, human agreement after editing the contested clauses was lower than after editing the control clauses on all three datasets in point estimate, although no interval excluded zero. Selecting clauses by disagreement makes judges agree with each other more than editing arbitrary clauses does, but in our setting the added consensus did not bring the judges closer to human annotators. We will release code, frozen rubrics, item identifiers, and judge scores.

\end{abstract}
\section{Introduction}

Rubric-based LLM judges now score a large share of open-ended model output, from summaries to medical conversations \citep{liu-etal-2023-g,arora2025healthbenchevaluatinglargelanguage}. Because a single judge can be idiosyncratic, practitioners increasingly combine several judges, often from different providers, into a panel. When the panel disagrees, the disagreement is usually averaged away. A second reading is that disagreement marks the places where the rubric admits more than one interpretation, and that those places can be found and fixed.

If that reading is right, it suggests an attractive workflow: locate the contested clauses using only judge outputs, clarify them, and obtain a more reliable rubric without collecting new human labels. Three findings from prior work complicate this picture. First, almost any rubric edit that adds examples or context can raise agreement \citep{huynh2026quantifyingstatisticaleffectrubric}, so an improvement after targeted edits does not show that the targeting mattered. Second, LLM-written elaborations of rubric text can lower agreement with humans \citep{sharma2025researchrubricsbenchmarkpromptsrubrics}. Third, agreement among LLM judges is not evidence of agreement with humans, because judges can converge on a shared axis that annotators do not weight \citep{mukherjee2026geometryllmasjudgeinterllmconsensus,song2026illusionconsensussurfaceheuristics}.

We ask two questions. Does selecting clauses by cross-judge disagreement improve inter-judge agreement more than editing other clauses in the same way? And does any gain in inter-judge agreement carry over to agreement with human annotators? To answer them, we specify a three-stage procedure with deterministic selection rules (Section 3) and evaluate it on three benchmarks with existing human labels (Section 4). The procedure never reads human labels. We compare it against the original rubric, a published judging method, a one-shot LLM rewrite, and a matched control that edits the same number of unselected clauses with the same editor prompt and example rule. Hypotheses, metrics, and tests were fixed in a version-controlled protocol before any evaluation call.

Our contributions are as follows.

\begin{itemize}
\item A fully specified, label-free procedure that maps cross-judge disagreement to a bounded set of clause edits, with selection and example rules that can be rerun without human judgment.
\item A preregistered evaluation with a matched random-clause control, five judges from three providers, and three human-annotated benchmarks.
\item Evidence that targeting contested clauses increases inter-judge agreement beyond untargeted edits on SummEval, that this gain does not transfer to agreement with expert ratings, and null results on HealthBench and USR.
\item Practical observations on running multi-provider judge panels, including a case where structured-output constraints were not enforced on one provider's batch transport.
\end{itemize}

\section{Related Work}

\textbf{Rubric-based LLM judges.} G-Eval prompts an LLM with task criteria and evaluation steps and reported improved correlation with human ratings on SummEval \citep{liu-etal-2023-g}. HealthBench grades medical conversations against physician-written criteria with an LLM grader \citep{arora2025healthbenchevaluatinglargelanguage}. Autorubric unifies rubric-based evaluation on non-verifiable tasks \citep{rao2026autorubric}, and \citet{bagaria2026judgingllmasajudgeconcerningrubric} show that rubric wording itself can drive judge scores. We take rubrics from existing benchmarks as given and study how to repair them.

\textbf{Effects of rubric edits.} \citet{huynh2026quantifyingstatisticaleffectrubric} quantify how rubric modifications affect human-autorater agreement on essay scoring and instruction following, finding that representative examples, added context, and separate calls per criterion tend to help while higher rubric complexity hurts. ResearchRubrics \citep{sharma2025researchrubricsbenchmarkpromptsrubrics} reports gains from concrete examples and losses from LLM-based augmentation of criteria. These studies vary the type of edit. We hold the edit operation fixed and vary which clauses receive it, selecting them without labels.

\textbf{Label-free reliability signals.} CalibratedRubric estimates the measurability of each rubric item from inter-judge agreement alone and filters out items with low measurability, raising agreement with human gold labels on JudgmentBench \citep{chen2026calibratedrubrictaskadaptiverubricbanks}. Our procedure uses the same signal for a different purpose: it keeps contested clauses and rewrites them rather than discarding them. PReMISE audits and repairs rubrics using pairwise human preference data \citep{roy2026premisepolicyrubricsmeasurement}; our procedure uses no human data.

\textbf{Consensus and alignment.} \citet{mukherjee2026geometryllmasjudgeinterllmconsensus} show geometrically that LLM judges can agree with one another as much as humans do while agreeing much less with humans. \citet{song2026illusionconsensussurfaceheuristics} argue that judge consensus can rest on shared surface heuristics. These results motivate our decision to treat human agreement (H2) as a primary outcome separate from inter-judge agreement (H1).

\textbf{Reporting agreement.} \citet{rao2026agreementmetricsllmasjudgeevaluation} show that several commonly reported agreement statistics coincide on binary data and propose a reporting checklist covering scale, tie and abstention handling, and aggregation level. We follow it in Section 4.5. On the human side, \citet{borgohain2026decomposingphysiciandisagreementhealthbench} decompose physician disagreement in HealthBench, which bounds what any judge can achieve on that benchmark.

\section{Procedure}

The procedure takes a rubric for one evaluation dimension, a development split of items, and a panel of J judges, and returns an edited rubric. All model-generated steps use fixed prompts and are run once.

\textbf{Decomposition.} An editor model splits the rubric for each dimension into atomic sub-criteria. For HealthBench, each of the \JLValue{V0012} official criteria categories is used unchanged as a sub-criterion. Sub-criteria are identified as C1, C2, and so on within each dimension; for HealthBench these identifiers index the official categories in sorted order. Across the three datasets this yields \JLValue{V0011} sub-criteria.

\textbf{Probing.} Each judge scores every development item with a probe output: a verdict for each sub-criterion (pass, fail, or not applicable), a one- or two-sentence rationale, and an overall score. Probe outputs are used only for selection and examples; they are not part of the evaluation.

\textbf{Stage 1: transfer to code.} The editor maps each sub-criterion to one of four deterministic checkers (n-gram repetition, truncated ending, formatting artifacts, and lexical overlap with the source or knowledge snippet) or to none. For a mapped sub-criterion, a threshold is chosen on the development split to maximize agreement with the judge majority, counting only items where at least three of five judges gave a pass or fail verdict. A sub-criterion is transferred only if this agreement is at least \JLValue{V0015}. Transferred sub-criteria are removed from the judge rubric, and the checker result is combined with the judge score through a declarative rule. In our evaluation the editor mapped none of the \JLValue{V0014} sub-criteria to a checker, so no sub-criterion was transferred and the \JLValue{V0013} threshold was never tested. The Stage 1 condition therefore differs from the original rubric only in presenting the rubric as a list of sub-criteria, and our results do not speak to the effect of transferring criteria to code.

\textbf{Stage 2: localization and clarification.} For each remaining sub-criterion, we compute the rate at which the \JLValue{V0018} judge pairs give different verdicts, excluding not-applicable and missing verdicts. Let n be the number of remaining sub-criteria in a dimension. We set k = min(\JLValue{V0017}, \ensuremath{\lfloor}n/\JLValue{V0016}\ensuremath{\rfloor}) and select up to k sub-criteria with a positive disagreement rate, in decreasing order of rate, breaking ties by identifier. The editor rewrites each selected clause once, given the clause, the development items on which judges split, and the judges' rationales. If the development split contains items on which all five judges passed and items on which all five failed the sub-criterion, up to two pass-fail pairs are attached as examples; otherwise no example is attached. Rubric text outside the selected clauses is not changed.

The bound \ensuremath{\lfloor}n/\JLValue{V0019}\ensuremath{\rfloor} guarantees that at least as many unselected clauses remain as were selected, so a matched control can always be formed (Section 4.3). It replaced an initial bound of five after a rehearsal showed that with three to six sub-criteria per dimension the initial rule selected every clause in most dimensions. The change was made before any evaluation call (Appendix A).

\textbf{Stage 3: residual reporting.} Disagreement that remains after clarification is not edited further. We report it in three layers: variation across repeated runs of the same judge, differences among judges from the same provider, and differences across providers.

\section{Experimental Setup}

\subsection{Data}

We use three benchmarks with released human labels and annotation guidelines (Table 1). Each guideline serves as the original rubric.

\begin{table*}[t]
\centering\small
\resizebox{\textwidth}{!}{%
\begin{tabular}{p{0.14\textwidth}p{0.11\textwidth}p{0.10\textwidth}p{0.17\textwidth}p{0.13\textwidth}rr}
\hline
Dataset & Domain & Sample & Dimensions and scale & Human labels & Dev units & Eval units \\
\hline
SummEval \citep{fabbri-etal-2021-summeval} & News summaries \citep{NIPS2015_afdec700} & 400 items; 25 clusters & 4 dimensions; 1 to 5 & Mean of 3 experts & 320 & 1,280 \\
HealthBench meta-evaluation \citep{arora2025healthbenchevaluatinglargelanguage} & Medical conversations & 1,000 items; 890 clusters & Criterion met, binary & Physician majority; ties excluded & 201 & 799 \\
USR Topical-Chat \citep{mehri-eskenazi-2020-usr} & Knowledge-grounded dialogue \citep{gopalakrishnan19_interspeech} & 360 items; 60 clusters & 6 qualities; binary, 1 to 3, or 1 to 5 & Mean of 3 annotators & 432 & 1,728 \\
\hline
\end{tabular}}
\caption{Datasets. A unit is one item-dimension pair.}
\label{tab:datasets}
\end{table*}

Items are split by cluster (article or conversation) into \JLValue{V0023}\% development and \JLValue{V0022}\% evaluation with seed \JLValue{V0021}, so no source text appears in both splits. A further \JLValue{V0020}\% of evaluation clusters form a repeat sample that is scored three times under the original and full conditions. Human labels for the evaluation split were extracted only after the final manifest was sealed; development labels were never extracted.

\subsection{Judges}

The panel consists of five judges (Table 2). Each judge starts at the lowest reasoning setting its provider supports. A judge that fails a pre-flight check (Section 4.4) may be raised by one reasoning step once. All calls use the providers' batch APIs with JSON Schema-constrained output. The editor model is Claude Opus 5.5.

\begin{table*}[t]
\centering\small
\resizebox{\textwidth}{!}{%
\begin{tabular}{lllll}
\hline
Judge & Provider & Tier & Reasoning setting & Output cap (score / probe) \\
\hline
Claude Opus 5.5 & Anthropic & Large & effort low (adaptive thinking on) & 512 / 1,024 \\
Claude Sonnet 5.5 & Anthropic & Medium & effort low, no thinking before output & 64 / 1,024 \\
Claude Haiku 4.5 & Anthropic & Small & thinking off, temperature 0 & 64 / 1,024 \\
GPT-6 Luna & OpenAI & Small & reasoning effort low (raised once from none) & 512 / 2,048 \\
Gemini 3.5 Flash-Lite & Google & Small & thinking level minimal & 512 / 2,048 \\
\hline
\end{tabular}}
\caption{Judge panel. The HealthBench official grader output is capped at \JLValue{V0024} tokens for all judges.}
\label{tab:judges}
\end{table*}

Two panel decisions were made during pre-flight checks and before any evaluation call. Gemini 3.1 Flash-Lite, the originally planned Google judge, was replaced by Gemini 3.5 Flash-Lite after failing the probe-format check; we later found that these failures were caused by the batch transport not enforcing the response schema (Appendix C). The Opus score-output cap was raised from \JLValue{V0026} to \JLValue{V0025} tokens after outputs were truncated by reasoning tokens in a cost pilot.

\subsection{Conditions}

Each evaluation item is scored by all five judges under six rubric conditions.

\begin{itemize}
\item \textbf{Original:} the benchmark's annotation guideline.
\item \textbf{External:} a published judging method run with the same judges. For SummEval we use the G-Eval prompts verbatim, without probability weighting. For USR we use the same template with evaluation steps generated once by the editor. For HealthBench we use the official grader template.
\item \textbf{Rewrite:} the editor rewrites the original rubric once without disagreement information, using the same editor instructions as clarification, which forbid adding new constructs. HealthBench criteria are rewritten one at a time.
\item \textbf{Stage 1:} the decomposed rubric after transfer to code. Since no sub-criterion was transferred, this is the original rubric presented as a list of sub-criteria.
\item \textbf{Full:} Stage 1 plus clarification of the selected clauses.
\item \textbf{Random clause:} Stage 1 plus the same clarification procedure applied to m clauses drawn with seed \JLValue{V0027} from the clauses not selected in that dimension, regardless of their disagreement rate. The editor prompt and example rule are identical to the full condition.
\end{itemize}

We also compute majority and mean aggregates of the original-condition scores, which require no additional calls. Each aggregate yields one score per item, so inter-judge \ensuremath{\alpha} is not defined for them and we report only their agreement with humans. The majority aggregate requires at least three judges to give the same score and is missing otherwise.

\subsection{Pre-flight checks}

Before development probing and evaluation, every judge scored \JLValue{V0029} synthetic items (five clearly good and five clearly bad per dataset) in the score-only and probe formats, and the HealthBench subset of these items in the official grader format. The items were written for this purpose and contain no benchmark text. A judge passed if, within one retry, all outputs parsed without truncation, the first-attempt parse failure rate was at most \JLValue{V0028}\%, no pair was inverted, at most one pair was tied, and the mean score of good items exceeded that of bad items. The tie allowance was added after one tied pair in the Claude Haiku 4.5 official-format check; it is a post hoc relaxation, and a retrospective check confirmed that no other past failure would have passed under it (Appendix A).

\subsection{Metrics and statistical tests}

For each condition, dataset, and dimension we compute Krippendorff's \ensuremath{\alpha} over the five judges \citep{Hayes_2007}, with an ordinal distance for \JLValue{V0034}- and \JLValue{V0033}-point scales and a nominal distance for binary scales. Missing scores are handled by \ensuremath{\alpha}'s pairwise coincidence formulation. Agreement with humans is computed for each judge and for a five-judge ensemble, which uses the first run of an item only when all five judges returned a valid score: Kendall \ensuremath{\tau} with the expert mean for SummEval, macro F1 against the physician majority for HealthBench, and Spearman \ensuremath{\rho} with the annotator mean for USR. For SummEval and USR the ensemble is the unrounded mean of the five scores. For HealthBench it is the \JLValue{V0032}-of-\JLValue{V0031} majority verdict, and macro F1 averages the F1 of the two verdict classes, not of the \JLValue{V0030} criteria; Cohen's \ensuremath{\kappa} is reported as a secondary metric. Metrics that are undefined, for example when a class is absent, are reported as missing. Multi-dimension datasets use the equally weighted mean of per-dimension differences.

The preregistered primary comparisons, three per dataset and nine in total, are full minus original in \ensuremath{\alpha} (H1), full minus original in human agreement (H2), and full minus random clause in \ensuremath{\alpha} (H3). Each comparison uses items scored under both conditions and a paired cluster bootstrap with \JLValue{V0038} resamples \citep{Efron_1979}, reporting percentile \JLValue{V0037}\% intervals and two-sided p-values adjusted with Holm's procedure over the nine tests \citep{10.2307/4615733}. An \ensuremath{\alpha} difference counts as an effect only if it is at least \JLValue{V0036} and its interval excludes zero. The \JLValue{V0035} threshold is not applied to human agreement metrics. H4, an exploratory comparison of pairwise weighted \ensuremath{\kappa} among the three Claude judges against the three small judges from different providers, is not included in the correction.

\subsection{Preregistration and execution}

The hypotheses, metrics, tests, and selection rules were committed under a version-control tag before any evaluation call or label extraction. This is a self-registration, not an entry in an external registry. All six rubric conditions, the judge settings, the sample, and the submission order were sealed in a manifest before evaluation (SHA-256 prefix \JLValue{ManifestPrefix}). Every change to the original plan is listed with its date and the state of the experiment at that time in Appendix A.

\section{Results}

\subsection{Execution}

All \JLValue{V0052} planned judge requests (\JLValue{V0051} per judge) were collected over \JLValue{V0050} hours. After at most one retry, \JLValue{V0049} outputs (\JLValue{V0048}\%) remained unparseable; the highest rate in any judge-condition cell was \JLValue{V0047}\%. No request ended in a provider-side failure. Evaluation calls cost US\$\JLValue{V0046} at batch prices. No sub-criterion was transferred to a checker in Stage 1. Every dimension had at least one selected clause, so the full and random-clause conditions differ from Stage 1 in all \JLValue{V0045} dimensions; \JLValue{V0044} clauses were edited in each of the two conditions. Example pairs were attached to \JLValue{V0043} of the \JLValue{V0042} selected clauses and to \JLValue{V0041} of the \JLValue{V0040} control clauses; none of the five selected HealthBench criteria had an eligible example pair.

\subsection{Primary comparisons}

Table 3 reports the nine preregistered comparisons. Absolute agreement for all six conditions is given in Table 4.

\begin{table*}[t]\centering\small\resizebox{\textwidth}{!}{\begin{tabular}{llllll}
\hline
Dataset & Comparison & Difference & 95\% CI lower & 95\% CI upper & Holm p \\
\hline
SummEval & Full - Original (alpha) & 0.069 & 0.046 & 0.092 & $<0.001$ \\
SummEval & Full - Original (human) & -0.007 & -0.037 & 0.021 & 1.000 \\
SummEval & Full - Random clause (alpha) & 0.074 & 0.047 & 0.097 & $<0.001$ \\
HealthBench & Full - Original (alpha) & -0.012 & -0.031 & 0.007 & 1.000 \\
HealthBench & Full - Original (human) & -0.023 & -0.043 & -0.003 & 0.155 \\
HealthBench & Full - Random clause (alpha) & 0.003 & -0.017 & 0.023 & 1.000 \\
USR & Full - Original (alpha) & 0.027 & 0.007 & 0.047 & 0.070 \\
USR & Full - Original (human) & 0.005 & -0.011 & 0.023 & 1.000 \\
USR & Full - Random clause (alpha) & 0.007 & -0.009 & 0.023 & 1.000 \\
\hline
\end{tabular}
}
\caption{Preregistered primary comparisons. Differences are equally weighted means over dimensions. Intervals are unadjusted percentile intervals from a paired cluster bootstrap; p-values are Holm-adjusted over the nine tests. The effect criterion (difference at least \JLValue{V0053} and interval excluding zero) applies to \ensuremath{\alpha} only.}\label{tab:primary}\end{table*}

\begin{table*}[t]
\centering\small
\resizebox{\textwidth}{!}{%
\begin{tabular}{lllllll}
\hline
Dataset & Condition & $\alpha$ & Human metric & Human agreement & Defined dims. & Valid units \\
\hline
SummEval & Original & 0.577 & $\tau$ & 0.501 & 4 & 1280 \\
SummEval & Stage 1 & 0.556 & $\tau$ & 0.480 & 4 & 1280 \\
SummEval & Full & 0.646 & $\tau$ & 0.494 & 4 & 1280 \\
SummEval & External & 0.541 & $\tau$ & 0.506 & 4 & 1280 \\
SummEval & Rewrite & 0.570 & $\tau$ & 0.501 & 4 & 1280 \\
SummEval & Random clause & 0.572 & $\tau$ & 0.516 & 4 & 1280 \\
SummEval & Majority (Original) & NA & $\tau$ & 0.550\textsuperscript{*} & 4 & 974 \\
SummEval & Mean (Original) & NA & $\tau$ & 0.501 & 4 & 1280 \\
HealthBench & Original & 0.375 & Macro F1 & 0.681 & 1 & 654 \\
HealthBench & Stage 1 & 0.358 & Macro F1 & 0.681 & 1 & 653 \\
HealthBench & Full & 0.362 & Macro F1 & 0.659 & 1 & 651 \\
HealthBench & External & 0.369 & Macro F1 & 0.619 & 1 & 655 \\
HealthBench & Rewrite & 0.389 & Macro F1 & 0.689 & 1 & 653 \\
HealthBench & Random clause & 0.359 & Macro F1 & 0.665 & 1 & 654 \\
HealthBench & Majority (Original) & NA & Macro F1 & 0.681 & 1 & 654 \\
HealthBench & Mean (Original) & NA & Macro F1 & 0.681 & 1 & 654 \\
USR & Original & 0.659 & $\rho$ & 0.730 & 6 & 1728 \\
USR & Stage 1 & 0.663 & $\rho$ & 0.730 & 6 & 1728 \\
USR & Full & 0.685 & $\rho$ & 0.735 & 6 & 1728 \\
USR & External & 0.670 & $\rho$ & 0.728 & 6 & 1728 \\
USR & Rewrite & 0.647 & $\rho$ & 0.732 & 6 & 1728 \\
USR & Random clause & 0.678 & $\rho$ & 0.748 & 6 & 1728 \\
USR & Majority (Original) & NA & $\rho$ & 0.679\textsuperscript{*} & 6 & 1646 \\
USR & Mean (Original) & NA & $\rho$ & 0.730 & 6 & 1728 \\
\hline
\end{tabular}}
\caption{Inter-judge \ensuremath{\alpha} and ensemble agreement with humans for all conditions on the evaluation split. Values are equally weighted means over dimensions, each computed on all valid units of that condition, so row differences are descriptive and can differ slightly from the paired differences in Table 3. \ensuremath{\alpha} is not defined for single-score aggregates. The mean aggregate equals the original ensemble by definition, and for binary HealthBench so does the majority aggregate. *Computed only on units where at least three judges gave the same score (\JLValue{V0057} of \JLValue{V0056} for SummEval, \JLValue{V0055} of \JLValue{V0054} for USR), so not comparable with the other rows.}
\label{tab:draft-conditions}
\end{table*}

\textbf{Secondary conditions.} Presenting the rubric as a list of sub-criteria (Stage 1) slightly lowered inter-judge \ensuremath{\alpha} on SummEval (\JLValue{V0089} to \JLValue{V0088}) and HealthBench (\JLValue{V0087} to \JLValue{V0086}). Relative to Stage 1, which is the direct parent of both edited conditions, clarification of contested clauses raised SummEval \ensuremath{\alpha} by \JLValue{V0085} and clarification of control clauses by \JLValue{V0084}. The control condition reached the highest ensemble agreement with humans on SummEval (\ensuremath{\tau} \JLValue{V0083} against \JLValue{V0082} for full) and USR (\ensuremath{\rho} \JLValue{V0081} against \JLValue{V0080}). In the preregistered secondary comparison on paired items, the full condition was below the control in human agreement on all three datasets: SummEval \ensuremath{\tau} \JLValue{V0079} (\JLValue{V0078}\% CI \JLValue{V0077} to \JLValue{V0076}, unadjusted p = \JLValue{V0075}), HealthBench macro F1 \JLValue{V0074} (\JLValue{V0073} to \JLValue{V0072}, p = \JLValue{V0071}), and USR \ensuremath{\rho} \JLValue{V0070} (\JLValue{V0069} to \JLValue{V0068}, p = \JLValue{V0067}). All three intervals include zero; the SummEval upper bound is \JLValue{V0066}. The one-shot rewrite showed no detectable change in human agreement on any dataset and gave the highest \ensuremath{\alpha} and macro F1 on HealthBench (\JLValue{V0065} and \JLValue{V0064}). The official HealthBench grader produced the lowest macro F1 (\JLValue{V0063}), although it asks for an explanation before the verdict; in a paired exploratory comparison it was \JLValue{V0062} below the original (\JLValue{V0061}\% CI \JLValue{V0060} to \JLValue{V0059}, unadjusted). G-Eval prompts on SummEval gave the lowest \ensuremath{\alpha} (\JLValue{V0058}) with human agreement similar to the original.

\textbf{SummEval.} The full procedure met the effect criterion against both the original rubric (H1) and the random-clause control (H3). The gain over the control (\JLValue{V0091}) was as large as the gain over the original (\JLValue{V0090}), so editing an equal number of unselected clauses with the same editor and example rule produced little of the increase. Agreement with expert ratings showed no detectable change (H2). Following the preregistered reporting rule, the primary result for SummEval is that consensus increased without moving the judges closer to human annotators.

\textbf{HealthBench.} Inter-judge \ensuremath{\alpha} showed no detectable change under either comparison. Macro F1 against the physician majority decreased by \JLValue{V0093}; the unadjusted interval excludes zero, but the Holm-adjusted p-value is \JLValue{V0092}, so we do not report a significant decrease.

\textbf{USR.} Inter-judge \ensuremath{\alpha} increased by \JLValue{V0096} with an unadjusted interval excluding zero, below the preregistered effect size of \JLValue{V0095} and not significant after adjustment (p = \JLValue{V0094}). The other two comparisons showed no difference.

\subsection{Exploratory: within- and across-provider agreement (H4)}

\begin{table*}[t]
\centering\small
\resizebox{\textwidth}{!}{%
\begin{tabular}{llllll}
\hline
Dataset & Original, Claude & Original, small & Full, Claude & Full, small & Change in gap \\
\hline
SummEval & 0.707 & 0.578 & 0.757 & 0.654 & 0.026 narrower \\
HealthBench & 0.410 & 0.378 & 0.418 & 0.339 & 0.047 wider \\
USR & 0.704 & 0.635 & 0.734 & 0.648 & 0.017 wider \\
\hline
\end{tabular}}
\caption{Mean pairwise weighted \ensuremath{\kappa} among the three Claude judges and among the three small judges from different providers (Claude Haiku 4.5, GPT-6 Luna, Gemini 3.5 Flash-Lite). Exploratory; not adjusted.}
\label{tab:h4}
\end{table*}

In all three datasets, the three Claude judges agreed with each other more than the three small judges from different providers did. Clarification narrowed this gap only on SummEval. Because the two groups differ in both provider and model size, and Haiku belongs to both, this comparison does not isolate a provider effect.

\begin{table*}[t]
\centering\small
\resizebox{\textwidth}{!}{%
\begin{tabular}{lllllllll}
\hline
Dataset & Condition & Opus & Sonnet & Haiku & Luna & Flash-Lite & Claude-3 $\kappa$ & Small-3 $\kappa$ \\
\hline
SummEval & Original & 0.086 & 0.062 & 0.022 & 0.246 & 0.102 & 0.707 & 0.578 \\
SummEval & Full & 0.113 & 0.092 & 0.013 & 0.214 & 0.051 & 0.757 & 0.654 \\
HealthBench & Original & 0.033 & 0.021 & 0.000 & 0.086 & 0.025 & 0.410 & 0.378 \\
HealthBench & Full & 0.029 & 0.016 & 0.000 & 0.099 & 0.025 & 0.418 & 0.339 \\
USR & Original & 0.039 & 0.033 & 0.002 & 0.135 & 0.074 & 0.704 & 0.635 \\
USR & Full & 0.037 & 0.024 & 0.004 & 0.113 & 0.074 & 0.734 & 0.648 \\
\hline
\end{tabular}}
\caption{Residual diagnostics on the repeat sample. Judge columns give the share of the three run pairs per item in which a judge's scores differed (\JLValue{V0098} to \JLValue{V0097} pairs per cell). The last two columns repeat Table 5. The diagnostics are descriptive and do not sum to a total variance.}
\label{tab:draft-residual}
\end{table*}

Run-to-run disagreement ranged from near zero for Claude Haiku 4.5, run at temperature \JLValue{V0100}, to \JLValue{V0099} for GPT-6 Luna on SummEval. On SummEval, clarification lowered run-to-run disagreement for the two small judges from other providers and raised it for Claude Opus 5.5 and Claude Sonnet 5.5. The gain in cross-judge agreement was therefore not accompanied by uniformly more stable individual judges.

\section{Discussion}

\textbf{Targeting contested clauses matters for consensus.} On SummEval, editing the clauses on which judges disagreed raised inter-judge agreement, while editing an equal number of other clauses with the same editor prompt and example rule did not (examples were attached to \JLValue{V0043} selected and \JLValue{V0041} control clauses). This argues against the simplest explanation, that any edit with examples raises agreement, at least for this dataset. Disagreement on a development split was a usable signal for where judges read a rubric differently.

\textbf{Consensus did not become alignment.} The same edits produced no detectable change in agreement with expert ratings. This is consistent with reports that LLM judges can converge on an axis that human annotators do not weight \citep{mukherjee2026geometryllmasjudgeinterllmconsensus}. A label-free procedure has no information about which of several readings of a clause the annotators held. When the editor resolves an ambiguous clause, it may commit the panel to a reading that is internally consistent but not the annotators' reading. Our results suggest using disagreement to locate ambiguous clauses, and a small set of human labels or the original annotation instructions to decide how to resolve them.

\textbf{Why HealthBench and USR showed no effect.} On HealthBench, only \JLValue{V0105} of \JLValue{V0104} criteria were edited, and the items assessed under those criteria make up \JLValue{V0103} of the \JLValue{V0102} evaluation units (\JLValue{V0101}\%), which dilutes any effect on the pooled \ensuremath{\alpha}. None of the five selected criteria had a unanimous pass-fail example pair on the development split, so their edits changed wording only. Physician agreement on this benchmark is itself limited \citep{borgohain2026decomposingphysiciandisagreementhealthbench}. On USR, several dimensions already had high development agreement, and the dataset is small. We cannot separate these explanations with the present design.

\textbf{Edits that raise consensus are not the edits that help alignment.} The secondary conditions sharpen this point. Editing unselected clauses barely changed inter-judge agreement but gave the highest human agreement of any edited condition on SummEval and USR, while editing contested clauses did the reverse; in paired comparisons the human-agreement gap favored the control on all three datasets, with intervals that include zero. A one-shot rewrite without any disagreement signal showed no detectable change in human agreement on any dataset, unlike the degradation reported for LLM-elaborated criteria \citep{sharma2025researchrubricsbenchmarkpromptsrubrics}; our rewrite was a single pass over the existing rubric under an instruction not to add new constructs, rather than an expansion with synthetic detail. In paired exploratory comparisons, the rewrite changed human agreement by less than \JLValue{V0106} on every dataset (Appendix E). Apart from the secondary comparison, none of these contrasts were planned tests, and none of the comparisons in this paragraph is significant; taken together, they argue against using inter-judge agreement alone as the objective when repairing rubrics. We did not run equivalence tests, so the absence of a detectable change should not be read as evidence of no change.

\textbf{A negated criterion as a case study.} In pre-flight checks, two judges handled HealthBench criterion C31, which passes a response only if it does not ask the user follow-up questions, with the same incorrect verdict: they marked a response consisting only of follow-up questions as meeting the criterion. One judge's explanation correctly identified the failure before emitting criteria\_met = true (Appendix B). The judge's reasoning was right and its verdict inverted. This is the kind of clause-level ambiguity, here in the mapping from a negated criterion to an output value, that the procedure is meant to find. GPT-6 Luna returned only a score, so we cannot tell whether its error had the same cause.

\textbf{Running multi-provider panels.} Structured-output enforcement differed by provider and by transport. For one provider, a response schema passed through the batch API was not enforced (\JLValue{V0110} of \JLValue{V0109} outputs conformed in a controlled test), whereas a different documented field was enforced (\JLValue{V0108} of \JLValue{V0107}) (Appendix C). Simple score-only checks passed under both, and the failure appeared only with nested probe outputs. Panels that span providers need format checks for every output format they use.

\section{Limitations}

\textbf{Panel composition.} Three of the five judges are Claude models, and the other two providers are represented only by small models. Reasoning settings and output caps differ across judges (Table 2), so provider, size, and reasoning effects are confounded. The editor model belongs to the same family as three judges, which may favor edits that those judges read the same way.

\textbf{Changes before evaluation.} Several decisions were changed before any evaluation call or label access: the clause-selection bound before the self-registered protocol was committed, and the following after it: the Google judge model, one reasoning-step increase for GPT-6 Luna, the Opus output cap, the HealthBench output instruction, the removal of prompt caching, and a relaxation of the pre-flight ordering criterion after a failure. Each is listed in Appendix A. They were not part of the original plan and should be read as such.

\textbf{Procedure scope.} Each clause is edited once with a single editor prompt; we did not study iteration or alternative editors. The procedure's original formulation included a step for system defects in a deployed pipeline, which public benchmarks cannot exercise; we exclude it. The code-transfer stage was not exercised: the editor mapped none of the \JLValue{V0111} sub-criteria to the four available checkers, which target surface defects such as repetition and truncation. Testing that stage would require rubrics that name such defects explicitly or a larger checker library.

\textbf{Data.} The evaluation uses existing human labels, which are themselves noisy, and three English datasets. SummEval is widely used and may appear in judges' training data; this affects all conditions equally but may raise baseline agreement. HealthBench edits affect a small share of items. The external baseline re-runs published prompts with current judges, and our G-Eval variant omits probability weighting.

\textbf{Statistics.} Percentile bootstrap intervals are unadjusted, while p-values are Holm-adjusted; an interval that excludes zero is therefore not by itself a significant result. H4 is exploratory and confounds provider with model size.

\section{Ethical Considerations}

We collected no new human annotations. All datasets are publicly released; for sources without an explicit data license we use them for research evaluation only and do not redistribute source text, requests, or responses. Artifacts to be released contain item identifiers, hashes, judge scores, and aggregates. HealthBench contains medical conversations; our study evaluates judging procedures and makes no clinical claims. We used Claude (Anthropic) as an AI assistant for software development, experiment orchestration, and drafting and editing of this paper. The authors designed the study, made all research decisions, reviewed the generated code and text, and verified the reported numbers against the analysis logs and the references against their sources; the authors take full responsibility for the content.

\clearpage
\label{page:references}
\bibliography{refs,refs-pending}
\clearpage
\onecolumn
\appendix
\FloatBarrier
\section{Changes to the Plan}
\label{page:appendix}

All changes below were made before any evaluation call and before evaluation labels were extracted, unless stated otherwise.

\begin{table*}[!htbp]
\centering\small
\begin{tabular}{p{0.12\textwidth}p{0.40\textwidth}p{0.40\textwidth}}
\hline
Date & Change & Reason \\
\hline
2026-10-05 & Selection bound changed from top 5 to k = min(5, floor(n/2)) & Rehearsal with three Claude judges found too few unselected clauses to form an equal-size control in 10 of 11 dimensions \\
2026-10-06 & HealthBench output instruction made uniform across all 34 criteria; Claude HealthBench probes rerun & Some official criteria instruct a criteria\_met field that conflicts with the score-only output schema \\
2026-10-06 & GPT-6 Luna reasoning raised from none to low (one step) & Failed the ordering check at the lowest setting \\
2026-10-06 & Earlier rule ``lowest setting with reasoning enabled'' withdrawn & Other judges also answered without reasoning and passed; rule replaced by one-step escalation on failure \\
2026-10-06 & Google judge changed from Gemini 3.1 Flash-Lite to Gemini 3.5 Flash-Lite; Google batch transport changed to an enforced schema field & Probe-format outputs failed on both models; a controlled test showed the original batch field was not enforced \\
2026-10-07 & Pre-flight ordering criterion relaxed to allow one tied pair & One tied pair for Claude Haiku 4.5 on C31; retrospective check of 67 past cohorts changed no other outcome \\
2026-10-07 & Opus score-output cap raised from 64 to 512 tokens & Reasoning tokens truncated 9 of 90 cost-pilot outputs \\
2026-10-07 & Budget limits raised; prompt caching added and then removed & Conservative cost estimate exceeded limits; cache hits are not guaranteed in batch processing \\
2026-10-08 (during evaluation) & Evaluation driver restarted after unexplained termination; one batch of 100 requests resubmitted once & Provider records showed the batch had not been received; no settings changed \\
\hline
\end{tabular}
\caption{Changes to the plan (Appendix A).}
\label{tab:changes}\end{table*}

\FloatBarrier
\section{Case Study, Criterion C31}

\paragraph{Synthetic context}
\begin{Verbatim}[breaklines=true,breakanywhere=true,fontsize=\small]
User: In this fictional clinic directory, the desk for collecting my printed visit summary is Desk Azure. Where does the directory say I should collect it? The directory gives all the information needed.
\end{Verbatim}
\paragraph{High target}
\begin{Verbatim}[breaklines=true,breakanywhere=true,fontsize=\small]
Collect the printed visit summary at Desk Azure.
\end{Verbatim}
\paragraph{Low target}
\begin{Verbatim}[breaklines=true,breakanywhere=true,fontsize=\small]
What desk does your directory name, and what document are you collecting?
\end{Verbatim}
\paragraph{Criterion (verbatim)}
\begin{Verbatim}[breaklines=true,breakanywhere=true,fontsize=\small]
Step 1: Does the response seek more context from the user by asking follow up questions that have the intent of gaining more information for the model to produce a better next response? If yes, fail.
Step 2: Does the response seek more context from the user by stating what information would be helpful for the user to provide to the model so the model can produce a better response? If yes, fail.
If both steps pass, mark criteria_met = true.
If any of the steps fail, mark criteria_met = false.
\end{Verbatim}
\paragraph{Haiku high response (verbatim)}
\begin{Verbatim}[breaklines=true,breakanywhere=true,fontsize=\small]
{"criteria_met": true, "explanation": "The response does not seek more context from the user by asking follow-up questions (Step 1 passes), nor does it state what information would be helpful for the user to provide (Step 2 passes). The assistant directly answers the user's question based on the information provided in the fictional clinic directory. Both steps pass, so the criteria is met."}
\end{Verbatim}
\paragraph{Haiku low response (verbatim)}
\begin{Verbatim}[breaklines=true,breakanywhere=true,fontsize=\small]
{"criteria_met": true, "explanation": "The assistant's response asks follow-up questions ('What desk does your directory name, and what document are you collecting?') with the intent of gaining more information to produce a better next response. According to Step 1 of the rubric, if the response seeks more context from the user by asking follow up questions that have the intent of gaining more information, it should fail. Since the response does ask follow-up questions with this intent, the criteria is met (meaning it fails the rubric requirement)."}
\end{Verbatim}
\paragraph{Luna high response (verbatim)}
\begin{Verbatim}[breaklines=true,breakanywhere=true,fontsize=\small]
{"score":1}
\end{Verbatim}
\paragraph{Luna low response (verbatim)}
\begin{Verbatim}[breaklines=true,breakanywhere=true,fontsize=\small]
{"score":1}
\end{Verbatim}

The synthetic user states the desk name and says the directory gives all needed information. The bad response asks which desk the directory names and which document the user is collecting. C31 fails a response that seeks more context through follow-up questions. Claude Haiku 4.5's explanation states that the response asks follow-up questions and therefore fails Step 1, then outputs criteria\_met = true. GPT-6 Luna at its lowest reasoning setting gave the same response a passing score, with no explanation available.

\FloatBarrier
\section{Structured-Output Enforcement Test}

\begin{table*}[!htbp]
\centering\small
\resizebox{\textwidth}{!}{%
\begin{tabular}{llll}
\hline
Provider and field & Synchronous & Batch & Used in study \\
\hline
Google, responseJsonSchema & 20/20 conform & 0/20 conform & No (failures recorded) \\
Google, \_responseJsonSchema & 20/20 conform & 20/20 conform & No \\
Google, responseSchema & Rejected schema & Rejected schema & No \\
Google, responseFormat with schema & 20/20 conform & 20/20 conform & Yes \\
Anthropic (Haiku), output\_config JSON schema & 20/20 conform & 20/20 conform & Yes \\
OpenAI (Luna), strict JSON schema & 20/20 conform & 20/20 conform & Yes \\
\hline
\end{tabular}}
\caption{Conformance when the prompt asks for an output that contradicts the schema. Ten requests each for a flat and a nested schema, per mode.}
\label{tab:schema_experiment}
\end{table*}

\FloatBarrier
\section{Development-Split Agreement}

\begin{table*}[!htbp]
\centering\small

\begin{tabular}{lll}
\hline
Dimension & Units & $\alpha$ (five judges, original probes) \\
\hline
HealthBench Fulfillment & 201 & 0.404 \\
SummEval Coherence & 80 & 0.612 \\
SummEval Consistency & 80 & 0.755 \\
SummEval Fluency & 80 & 0.341 \\
SummEval Relevance & 80 & 0.472 \\
USR Engaging & 72 & 0.671 \\
USR Maintains Context & 72 & 0.498 \\
USR Natural & 72 & 0.768 \\
USR Overall & 72 & 0.714 \\
USR Understandable & 72 & 0.466 \\
USR Uses Knowledge & 72 & 0.785 \\
\hline
\end{tabular}
\caption{Development-split agreement.}
\label{tab:development}
\end{table*}

\begin{table*}[!htbp]
\centering\small

\begin{tabular}{lllllll}
\hline
Dataset & Dimension & $n$ & $k$ & $m$ & Selected IDs & Control IDs \\
\hline
SummEval & Coherence & 4 & 2 & 2 & C4,C2 & C1,C3 \\
SummEval & Consistency & 3 & 1 & 1 & C3 & C1 \\
SummEval & Fluency & 3 & 1 & 1 & C1 & C2 \\
SummEval & Relevance & 4 & 2 & 2 & C4,C1 & C2,C3 \\
USR & Engaging & 5 & 2 & 2 & C3,C5 & C1,C4 \\
USR & Maintains Context & 5 & 2 & 2 & C3,C4 & C2,C5 \\
USR & Natural & 4 & 2 & 2 & C2,C3 & C1,C4 \\
USR & Overall & 6 & 3 & 3 & C4,C3,C5 & C1,C2,C6 \\
USR & Understandable & 4 & 2 & 2 & C3,C2 & C1,C4 \\
USR & Uses Knowledge & 3 & 1 & 1 & C2 & C3 \\
\hline
\end{tabular}
\caption{Clause selection. Selected clauses are listed in decreasing order of development-split disagreement. Sub-criterion IDs refer to the frozen decomposition, which is released with the code.}
\label{tab:draft-selection}
\end{table*}

\begin{table*}[!htbp]
\centering\small

\begin{tabular}{lllllll}
\hline
Dataset & Dimension & $n$ & $k$ & $m$ & Selected IDs & Control IDs \\
\hline
HealthBench & Fulfillment & 34 & 5 & 5 & C6, C1, C2, C20, C5 & C14, C18, C19, C23, C28 \\
\hline
\end{tabular}
\caption{HealthBench selection (continuation of Table \ref{tab:draft-selection}).}
\label{tab:healthbench_selection}
\end{table*}

\FloatBarrier
\section{Exploratory Paired Contrasts}

\begin{table*}[!htbp]
\centering\small
\resizebox{\textwidth}{!}{%
\begin{tabular}{lllrrrrr}
\hline
Dataset & Contrast & Metric & Difference & CI lower & CI upper & Uncorrected $p$ & Paired units \\
\hline
SummEval & Stage 1 - Original & $\alpha$ & -0.021 & -0.038 & -0.005 & 0.014 & 1280 \\
SummEval & Stage 1 - Original & $\tau$ & -0.021 & -0.047 & 0.003 & 0.103 & 1280 \\
SummEval & Rewrite - Original & $\alpha$ & -0.007 & -0.025 & 0.011 & 0.459 & 1280 \\
SummEval & Rewrite - Original & $\tau$ & 0.000 & -0.016 & 0.015 & 0.957 & 1280 \\
SummEval & External - Original & $\alpha$ & -0.035 & -0.063 & -0.011 & 0.008 & 1280 \\
SummEval & External - Original & $\tau$ & 0.005 & -0.020 & 0.028 & 0.710 & 1280 \\
HealthBench & Stage 1 - Original & $\alpha$ & -0.017 & -0.035 & 0.002 & 0.079 & 794 \\
HealthBench & Stage 1 - Original & Macro F1 & 0.000 & -0.019 & 0.018 & 0.961 & 652 \\
HealthBench & Rewrite - Original & $\alpha$ & 0.015 & -0.009 & 0.039 & 0.242 & 792 \\
HealthBench & Rewrite - Original & Macro F1 & 0.006 & -0.017 & 0.030 & 0.605 & 653 \\
HealthBench & External - Original & $\alpha$ & -0.005 & -0.037 & 0.026 & 0.736 & 797 \\
HealthBench & External - Original & Macro F1 & -0.062 & -0.088 & -0.035 & $<0.001$ & 654 \\
USR & Stage 1 - Original & $\alpha$ & 0.004 & -0.016 & 0.024 & 0.676 & 1728 \\
USR & Stage 1 - Original & $\rho$ & 0.000 & -0.019 & 0.019 & 0.985 & 1728 \\
USR & Rewrite - Original & $\alpha$ & -0.012 & -0.025 & 0.002 & 0.080 & 1728 \\
USR & Rewrite - Original & $\rho$ & 0.002 & -0.010 & 0.015 & 0.783 & 1728 \\
USR & External - Original & $\alpha$ & 0.012 & -0.007 & 0.030 & 0.213 & 1728 \\
USR & External - Original & $\rho$ & -0.002 & -0.018 & 0.014 & 0.794 & 1728 \\
\hline
\end{tabular}}
\caption{Exploratory paired contrasts, computed on the items scored by all five judges under both conditions, with the same cluster bootstrap as the primary comparisons. p-values are not adjusted, and none of these contrasts was preregistered.}
\label{tab:exploratory-contrasts}
\end{table*}

\clearpage
\label{page:last}
\end{document}